%% file: main.tex
\documentclass{article} 
\usepackage{iclr2026_conference,times}

\input{math_commands.tex}

\usepackage{url}
\usepackage{wrapfig}
\usepackage{booktabs}    
\usepackage{multirow}    
\usepackage{makecell}    
\usepackage{subfig}
\usepackage{graphicx}
\usepackage[table]{xcolor}
\def\rebuttal{\textcolor{black}}
\definecolor{c6}{HTML}{95baa6}
\definecolor{c7}{HTML}{4b8dbc}

\definecolor{c5}{HTML}{C00000}
\newcommand{\red}[1]{\textcolor{c5}{#1}}

\definecolor{mycolor}{HTML}{a7caea} 
 
\definecolor{mygreen}{HTML}{95BAA6}
\definecolor{myblue}{HTML}{4b8dbc}
\definecolor{myyellow}{HTML}{E2CD89}

\definecolor{linkcolor}{HTML}{ED1C24}

\definecolor{mydarkgreen}{rgb}{0.02,0.6,0.02}

\definecolor{iccvblue}{RGB}{0,102,204} 
\usepackage[pagebackref,colorlinks,citecolor=iccvblue,linkcolor=linkcolor]{hyperref}
\newcommand{\cmark}{\ding{51}}

\usepackage{algorithm}
\usepackage{algpseudocode}
\usepackage{amsmath}
\usepackage{amssymb}
\usepackage{mathtools}
\usepackage{amsthm}
\usepackage{makecell}
\usepackage{enumitem}
\usepackage{pifont}
\usepackage{marvosym}

\theoremstyle{plain}
\newtheorem{theorem}{Theorem}[section]
\newtheorem{proposition}[theorem]{Proposition}

\theoremstyle{definition}

\theoremstyle{remark}

\title{QuantSparse: Comprehensively Compressing Video Diffusion Transformer with Model Quantization and Attention Sparsification}

\author{Weilun Feng$^{1,2}$,\enspace Chuanguang Yang$^{1}$\thanks{Corresponding authors: Chuanguang Yang, yangchuanguang@ict.ac.cn; Zhulin An, anzhulin@ict.ac.cn},\enspace Haotong Qin$^{3}$,\enspace Mingqiang Wu$^{1,2}$,\enspace Yuqi Li$^{4}$,\\
\textbf{Xiangqi Li$^{1,2}$,\enspace Zhulin An$^{1}$\footnotemark[1],\enspace Libo Huang$^{1}$, Yulun Zhang$^{5}$,\enspace Michele Magno$^{3}$,\enspace Yongjun Xu$^{1}$}  \\
\textsuperscript{1}State Key Laboratory of AI Safety, Institute of Computing Technology, Chinese Academy of Sciences\\
\textsuperscript{2}University of Chinese Academy of Sciences\quad
\textsuperscript{3}ETH Z\"{u}rich \\
\textsuperscript{4}City College of New York, City University of New York, USA \quad
\textsuperscript{5}Shanghai Jiao Tong University \\
\texttt{\small\{fengweilun24s,yangchuanguang,lixiangqi24s,anzhulin,xyj\}@ict.ac.cn}\\
\texttt{\small\{haotong.qin,michele.magno\}@pbl.ee.ethz.ch,}\ 
\texttt{\small wumingqiang25@mails.ucas.ac.cn,}\\
\texttt{\small\{yuqili010602,www.huanglibo,yulun100\}@gmail.com}
}

\iclrfinalcopy 
\begin{document}

\maketitle

\begin{figure}[h]
  \centering
  \includegraphics[width=1.0\linewidth]{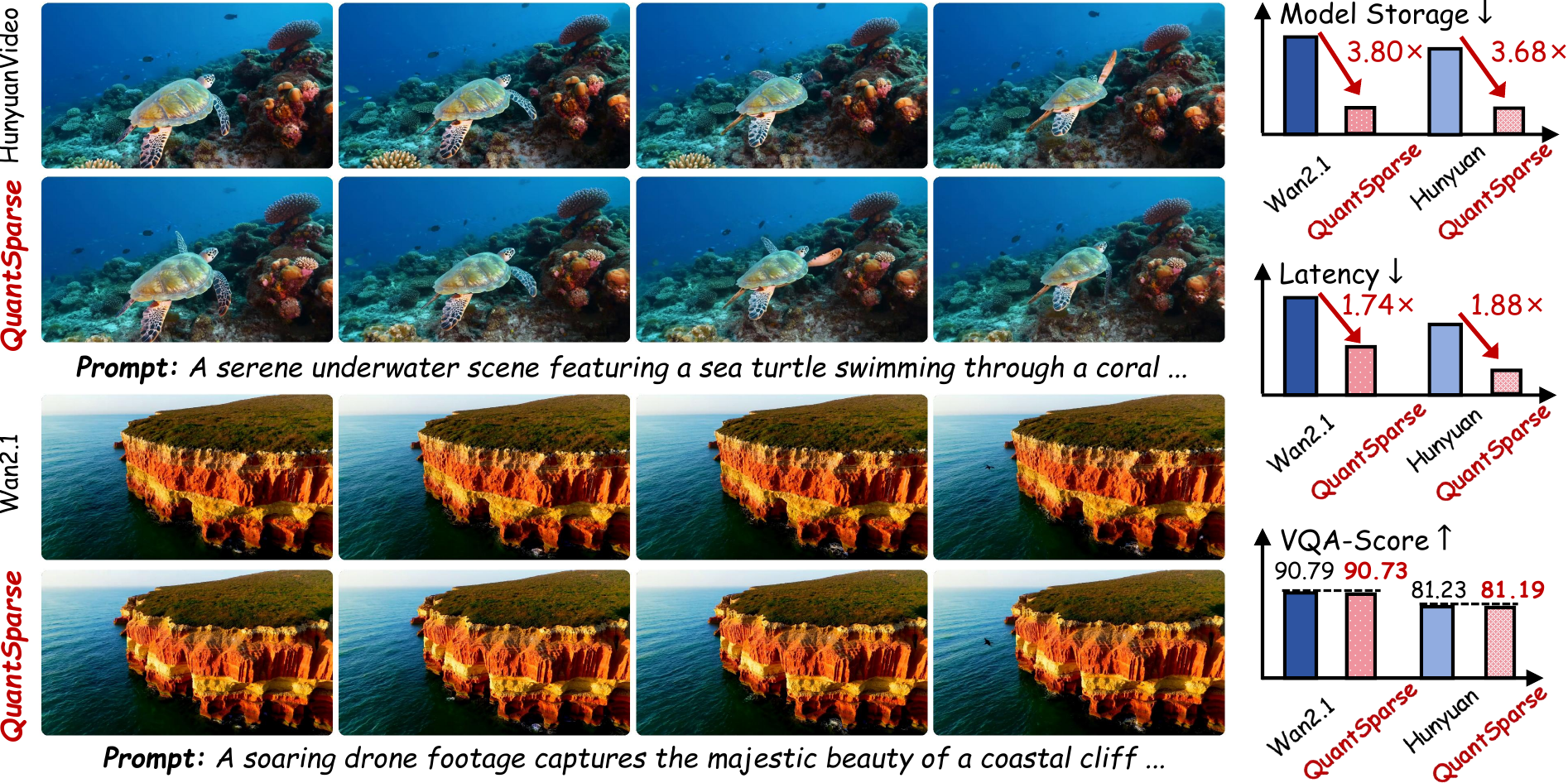}
  \caption{\textbf{QuantSparse} effectively quantizes Wan2.1-14B~\citep{wan2025wan} and HunyuanVideo~\citep{kong2024hunyuanvideo} to W4A8 with 15\% attention density without compromising visual quality.}
  \label{fig:teaser}
\end{figure}

\input{sec/0_abs}

\input{sec/1_intro}

\input{sec/2_related}

\input{sec/3_methods}

\input{sec/4_experiments}

\input{sec/5_conclusion}

\section{Acknowledgements}
This work is supported by the National Natural Science Foundation of China under Grant Number 62406312 and 62476264, the Beijing Natural Science Foundation under Grant Number 4244098, the Science Foundation of the Chinese Academy of Sciences, and the Swiss National Science Foundation (SNSF) project 200021E\_219943 Neuromorphic Attention Models for Event Data (NAMED).

\section{Ethics statement}
This research strictly adheres to the ICLR Code of Ethics with no ethics-related risks: it uses public open-source video-generation models (Wan2.1~\citep{wan2025wan} and HunyuanVideo~\citep{kong2024hunyuanvideo}) and focuses on algorithmic innovation for inference acceleration and compression, without involving scenarios endangering public safety, infringing privacy, or producing discrimination.

\section{Reproducibility statement}
To ensure reproducibility, experimental configurations, method details, and evaluation metrics are thoroughly described in Sec.~\ref{sec:main_detail} and Appendix Sec.~\ref{sec:more_expe_detail}. Experimental results of comparative methods are sourced from public literature, and our experiments strictly follow the same configurations as baseline methods for fair comparison. The key codes and the presented video source files are also attached in the supplementary materials. For the theorem used in the paper, we also provided a detailed proof in Appendix Sec.~\ref{sec:proof}.

\bibliography{iclr2026_conference}
\bibliographystyle{iclr2026_conference}

\newpage
\appendix

\input{sec/6_appendix}

\end{document}

%% file: math_commands.tex
\usepackage{amsmath,amsfonts,bm}

\def\eqref#1{equation~\ref{#1}}

\def\1{\bm{1}}

\DeclareMathAlphabet{\mathsfit}{\encodingdefault}{\sfdefault}{m}{sl}
\SetMathAlphabet{\mathsfit}{bold}{\encodingdefault}{\sfdefault}{bx}{n}



%% file: sec/0_abs.tex

\begin{abstract}


Diffusion transformers exhibit remarkable video generation capability, yet their prohibitive computational and memory costs hinder practical deployment. Model quantization and attention sparsification are two promising directions for compression, but each alone suffers severe performance degradation under aggressive compression. Combining them promises compounded efficiency gains, but naive integration is ineffective. The sparsity-induced information loss exacerbates quantization noise, leading to amplified attention shifts. To address this, we propose \textbf{QuantSparse}, a unified framework that integrates model quantization with attention sparsification. Specifically, we introduce \textit{Multi-Scale Salient Attention Distillation}, which leverages both global structural guidance and local salient supervision to mitigate quantization-induced bias. In addition, we develop \textit{Second-Order Sparse Attention Reparameterization}, which exploits the temporal stability of second-order residuals to efficiently recover information lost under sparsity. Experiments on HunyuanVideo-13B demonstrate that QuantSparse achieves 20.88 PSNR, substantially outperforming the state-of-the-art quantization baseline Q-VDiT (16.85 PSNR), while simultaneously delivering a \textbf{3.68$\times$} reduction in storage and \textbf{1.88$\times$} acceleration in end-to-end inference.
Our code will be released in \url{https://github.com/wlfeng0509/QuantSparse}.

\end{abstract}

%% file: sec/1_intro.tex
\section{Introduction}


Recently, Diffusion Transformer (DiT)~\citep{peebles2023dit} has attracted significant attention due to its outstanding capability in visual generation, particularly in video generation~\citep{liu2024sora, sun2024sorasurvey, open-sora, an2025TheInnovationInformatics}. Despite the remarkable progress, state-of-the-art models such as Wan2.1-14B~\citep{wan2025wan} still demand extraordinary computational resources: generating a single high-resolution video clip can consume more than \textbf{20GB} of GPU memory and take nearly \textbf{one hour} of inference time. Such prohibitive memory and latency requirements fundamentally limit the deployment of diffusion-based video generation models in real-world applications, especially under resource-constrained scenarios.

Model quantization~\citep{jacob2018quantizationandtrain, gholami2022quantizationsurvey,ma2023ompq,ma2023solving} and attention sparsification~\citep{xi2025svg, yuan2024ditfastattn} have emerged as two promising directions for compression and acceleration. Quantization reduces memory footprint and computation by representing weights and activations in compact integer formats, while attention sparsification prunes redundant computations by removing negligible attention scores. However, pushing either technique to the extreme inevitably causes severe degradation. For instance, binary quantization~\citep{zheng2024binarydm, zheng2024bidm} collapses representational capacity, while aggressive sparsification~\citep{xi2025svg, zhang2025jenga} discards crucial context information.

Since quantization and sparsification are fundamentally orthogonal, a natural idea is to combine them for compounded efficiency gains while maintaining complementary benefits. Ideally, such integration could approach a Pareto frontier between performance and efficiency. Yet, our empirical analysis shows that \textbf{naïvely combining quantization and sparsification leads to severe performance degradation}. We attribute this to an \emph{amplified attention shift}: while sparsification removes low-magnitude attention weights, quantization introduces systematic perturbations to the remaining attention products. These two effects reinforce each other, producing compounded distortions in attention distributions and severely impairing fine-grained dependency modeling in video generation.

\begin{figure}[t]
    \centering
    \includegraphics[width=1.0\linewidth]{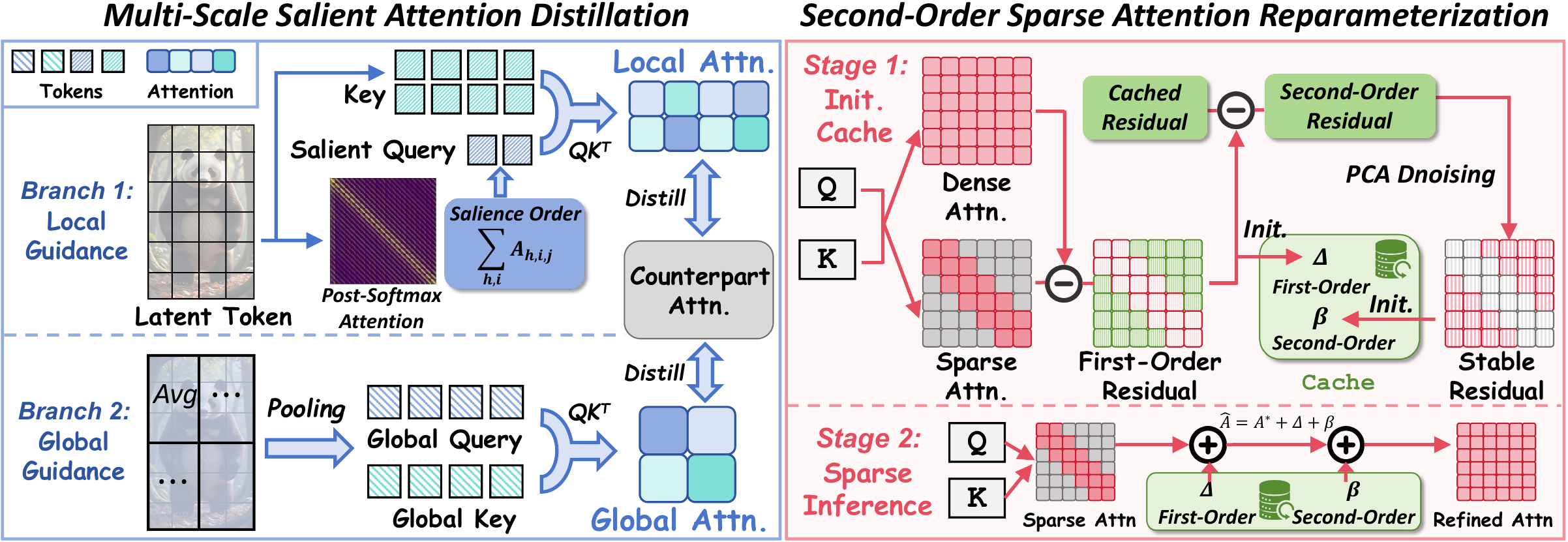}
    \caption{\textbf{Overview of proposed QuantSparse.} \rebuttal{\textbf{Left:} During calibration, we apply two parallel attention distillation branch for efficient and robust attention alignment. \textbf{Right:} During inference, we apply an accurate attention approximation using temporal stable second-order residual.}}
\label{fig:overview}
\vspace*{-0.2in}
\end{figure}

To overcome this challenge, we propose \textbf{QuantSparse}, a unified compression framework that synergistically integrates model quantization and attention sparsification as shown in Fig.~\ref{fig:overview}. QuantSparse introduces two novel techniques. First, \textit{Multi-Scale Salient Attention Distillation (MSAD).} We design a memory-efficient distillation scheme that balances global and local supervision. Specifically, we employ \emph{global guidance} by distilling attention patterns on downsampled token sequences to capture coarse structural topology, while \emph{local guidance} focuses high-resolution supervision on a small set of salient tokens that dominate the attention distribution. Second, \textit{Second-Order Sparse Attention Reparameterization (SSAR).} We exploit the temporal stability of \textit{second-order residuals} to recover information lost due to sparsity. Furthermore, we introduce singular value decomposition (SVD) projection onto dominant principal components, enabling a lightweight yet accurate correction mechanism that restores fine-grained attention outputs at negligible computational overhead.

Our contributions can be summarized as follows:
\begin{enumerate}
\item We provide formal analysis of the \emph{amplified attention shift} problem, showing that naive integration of quantization and sparsification severely damages video generation quality.
\item We propose \textbf{QuantSparse}, a unified compression framework that seamlessly combines model quantization and attention sparsification, breaking the traditional trade-off between efficiency and performance.
\item We introduce two key techniques: \textit{Multi-Scale Salient Attention Distillation} for robust attention alignment and \textit{Second-Order Sparse Attention Reparameterization} for temporally stable correction for efficient yet accurate approximation of full-attention outputs.
\item Extensive experiments on large-scale video generation models ranging from 1.3B to 14B parameters demonstrate that QuantSparse achieves superior efficiency–quality trade-offs,  outperforming both quantization-only and sparsification-only baselines, while preserving state-of-the-art performance.
\end{enumerate}

%% file: sec/2_related.tex
\section{Related Works}

\subsection{Sparse Attention in Diffusion Models}
Sparse attention has been extensively explored in transformer-based models to accelerate attention computation~\citep{lu2025moba, yuan2025native, lou2024sparser, gao2024seerattention, zhang2025spargeattn}.  
In large language models, common designs include sliding-window~\citep{xiao2024infllm, xiao2024duoattention, zhang2023h2o} and sink-based patterns~\citep{fu2406moa, xiao2023efficient}.  
For diffusion-based visual generation, spatial window masks~\citep{yuan2024ditfastattn, zhang2025fasttile, ren2025grouping} and spatial-temporal masks~\citep{xi2025svg} have been proposed.  
Other approaches dynamically generate masks via sampling~\citep{zhang2025spargeattn} or low-resolution attention~\citep{zhang2025jenga}, though at higher computational cost.  
However, these works mainly focus on preserving the original attention pattern, while the adaptation to other acceleration techniques that alter attention distributions, such as quantization, remains underexplored.

\subsection{Quantization in Diffusion Models}
Quantization~\citep{gholami2022quantizationsurvey, chitty2023transformerquantsurvey,pilipovic2018cnnquantsurvey,ma2024affinequant,ma2024outlier,feng2025s2qvdit,ding2024reg,feng2025mpqdmv2} reduces model precision to improve efficiency and has been applied to diffusion-based visual generation~\citep{shang2023ptq4dm, li2024qdm, he2024ptqd, huang2024tfmq, he2023efficientdm, feng2025mpqdm, wu2024ptq4dit, zheng2024bidm, zheng2024binarydm, li2024svdqunat}.  
For video generation, some works target the attention module~\citep{zhang2024sageattention, zhang2024sageattention2, zhang2025sageattention3}, but often keep linear operations in high precision.  
Other methods focus on quantizing linear layers: Q-DiT~\citep{chen2024qdit} uses automatic granularity allocation; ViDiT-Q~\citep{zhao2024vidit} adopts a static–dynamic strategy; Q-VDiT~\citep{feng2025qvdit} introduces temporal distillation.  
These methods primarily pursue acceleration via quantization, without exploring its synergy with sparse attention.  
In this work, we integrate the two orthogonal compression techniques to enhance the efficiency and practicality of video generation models.

%% file: sec/3_methods.tex
\section{Methods}

\subsection{Preliminary}

\subsubsection{Post-Training Quantization (PTQ)}  

Model Quantization~\citep{gholami2022quantizationsurvey, chitty2023transformerquantsurvey} reduces weights/activations from floating-point (FP32) to low-bit integers (e.g., INT8). Given an floating-point tensor $ \mathbf{X} \in \mathbb{R}^{d} $ with dimension $d$, quantization maps $\mathbf{X}$ to a discrete representation $\mathbf{X}_\mathbf{Q} \in \{0, 1, \dots, 2^b - 1\}^{d}$ as:

\begin{equation}
\mathbf{X}_{Q} = \text{clip}\left( \left\lfloor \frac{\mathbf{X}}{s} \right\rceil + z, 0, 2^b - 1 \right),\quad
Q(\mathbf{X}) = s \cdot (\mathbf{X}_Q - z),
\end{equation}

with scale $s$, zero-point $z$, and bit-width $b$, $Q(\mathbf{X})$ denotes the de-quantized value. Post-training Quantization (PTQ)~\citep{wei202ptqsurvey, wu2024ptq4dit} calibrates $(s,z)$ on a small dataset by minimizing reconstruction error:

\begin{equation}
\mathcal{L}_{\text{quant}} = \min_{s, z} \sum_{\mathbf{X}_i \in \mathcal{D}_{\text{cal}}} \left\| \mathbf{X}_i - Q(\mathbf{X}_{Q_i}; s, z) \right\|_2^2.
\end{equation}

Notably, PTQ avoids retraining the model weights, thus being computationally efficient.

\subsubsection{Sparse Attention}

Sparse attention~\citep{zhang2025spargeattn, xi2025svg, yuan2024ditfastattn} prunes token pairs via a mask $\mathbf{M} \in \{0, 1\}^{L \times L}$, reducing complexity from $\mathcal{O}(L^2)$ to near-linear ($L$ is the sequence length).  
Given $\mathbf{X} \in \mathbb{R}^{L \times d_{\text{in}}}$ and query, key, value projection matrices $\mathbf{W}_q, \mathbf{W}_k, \mathbf{W}_v \in \mathbb{R}^{d_{\text{out}} \times d_{\text{in}}}$, sparse attention computes:
\begin{equation}
\begin{gathered}
\mathbf{Q} = \mathbf{X} \mathbf{W}_q^\top,\quad \mathbf{K} = \mathbf{X} \mathbf{W}_k^\top,\quad \mathbf{V} = \mathbf{X} \mathbf{W}_v^\top, \\
\text{SparseAttention}(\mathbf{Q}, \mathbf{K}, \mathbf{V}; \mathbf{M}) = \text{softmax}\left( \frac{\mathbf{Q}\mathbf{K}^\top}{\sqrt{d_k}} \odot \mathbf{M} \right)\mathbf{V},
\end{gathered}
\end{equation}
where $\odot$ denotes element-wise multiplication.

\begin{figure}[h]
    \centering
    \subfloat[][Token saliency distribution.]{
        \includegraphics[width=0.295\linewidth]{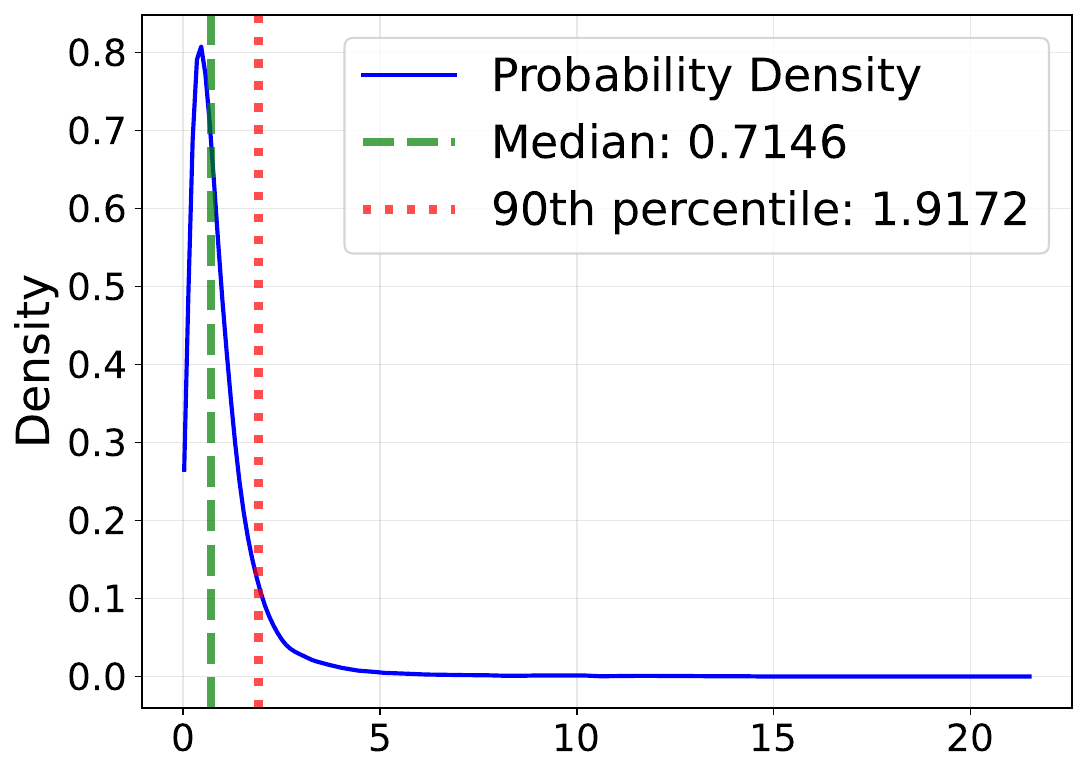}
        \label{fig:token_saliency}
    }
    \subfloat[][w/o distillation.]{
        \includegraphics[width=0.185\linewidth]{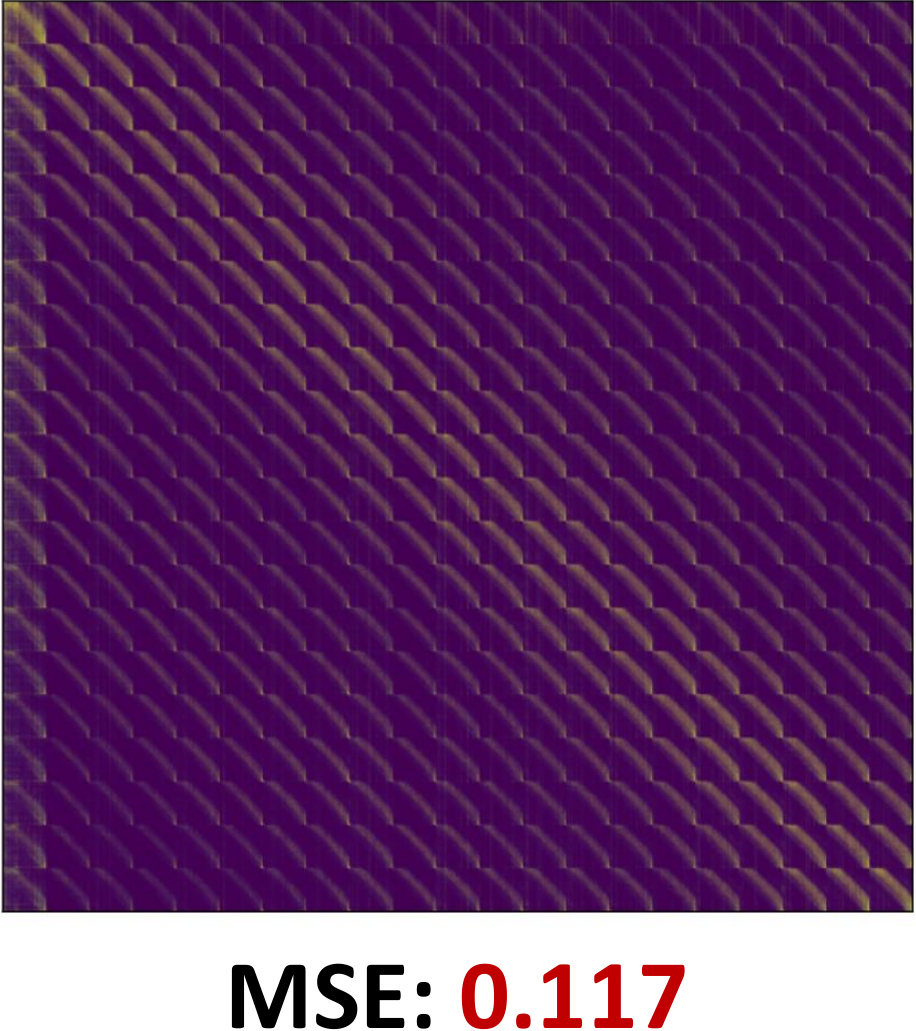}
         \label{fig:attn_wo_distill}
    }
    \subfloat[][w distillation.]{
        \includegraphics[width=0.185\linewidth]{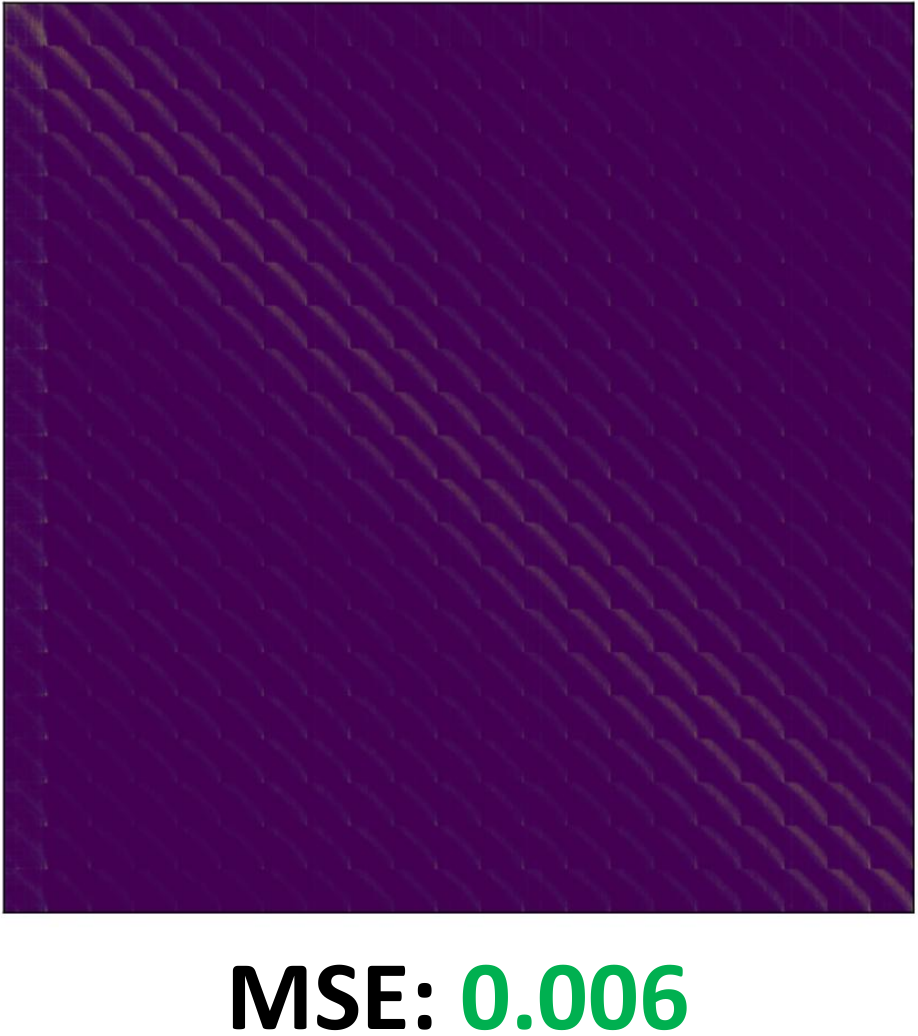}
         \label{fig:attn_w_distill}
    }
    \subfloat[][Memory consumption.]{
        \includegraphics[width=0.275\linewidth]{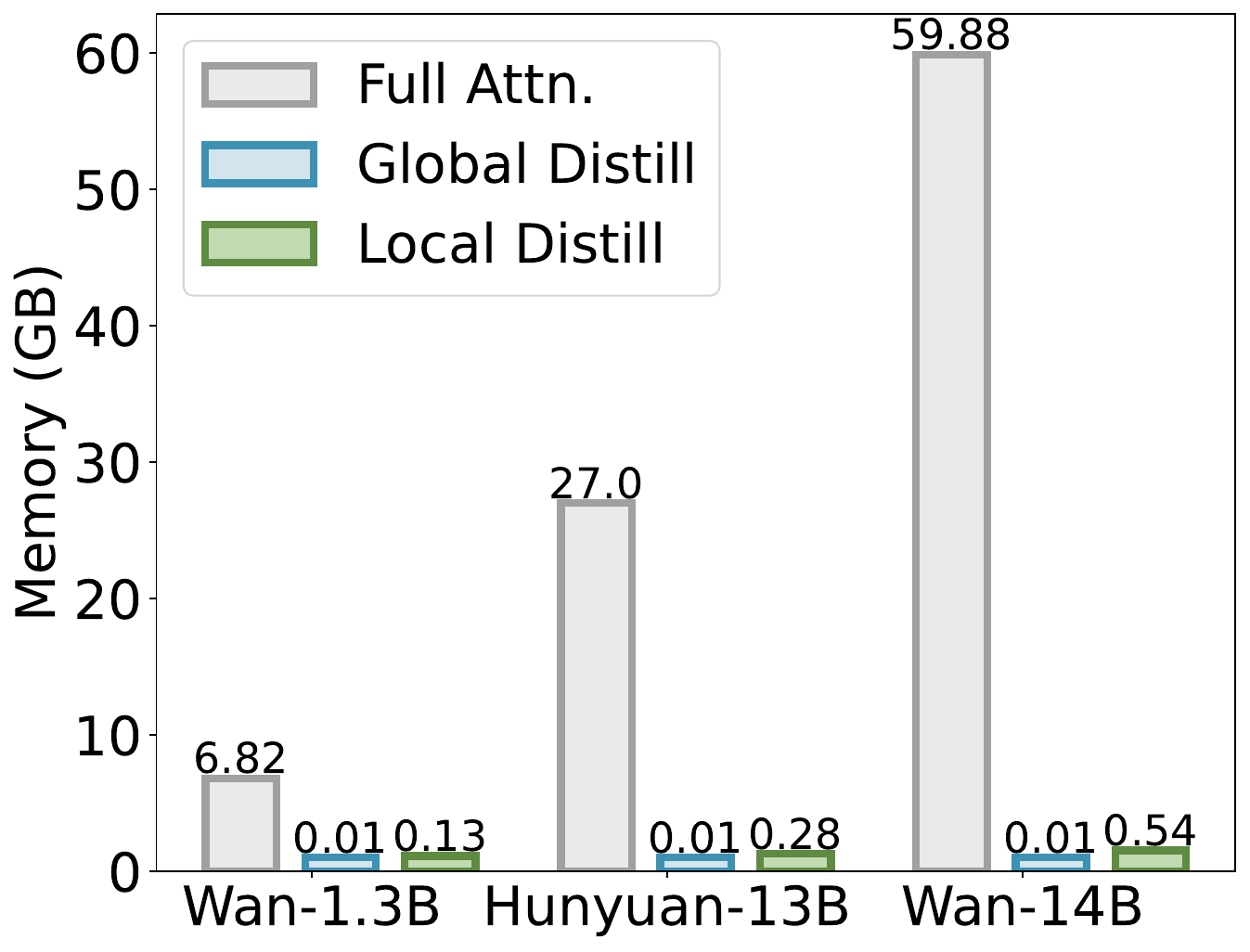}
         \label{fig:attn_distill_memory}
    }
    \caption{\textbf{The motivation and effect of Multi-Scale Salient Attention Distillation.} (a): Token saliency distribution of Wan2.1-1.3B~\citep{wan2025wan} \textit{block19 head1}. Only less than 10\% tokens are salient. (b)(c): Visualization of attention difference between quantized model and FP model. (d): Memory consumption of different attention distillation.}
\label{fig:attn_distill}
\end{figure}

\subsection{Multi-Scale Salient Attention Distillation}

The combination PTQ and sparse attention offers a promising route toward efficient video generation. However, naively integrating these techniques results in severe performance degradation.

\begin{proposition}
Quantization injects noise \( \epsilon \) into the QK dot product \( \mathbf{Q}\mathbf{K}^\top \), yielding a systematic bias \( \delta \):
\begin{equation}
\begin{gathered}
\hat{\mathbf{Q}}=Q(\mathbf{X})Q(\mathbf{W}_q)^{\top},~
\hat{\mathbf{K}}=Q(\mathbf{X})Q(\mathbf{W}_k)^{\top}, \\
\hat{\mathbf{Q}}\hat{\mathbf{K}}^\top = \mathbf{Q}\mathbf{K}^\top + \epsilon, \quad \text{where} \quad \|\epsilon\|_F \leq \delta.
\end{gathered}
\label{eq:quant_error}
\end{equation}
The parallel error caused by quantization and sparse attention further leads to a compounded shift:
\begin{equation}
    \Delta_{\emph{total}} = \Delta_{\emph{sparse}} + \Delta_{\emph{quant}} + \mathcal{O}(\|\epsilon\|_F \cdot \|\mathbf{M}\|_0).
\end{equation}
\label{prop:attn_shift}
\end{proposition}

\vspace{-0.2in}

Proposition~\ref{prop:attn_shift} indicates that the joint of quantization and sparse attention introduces an \textit{amplified attention shift} (see Fig.~\ref{fig:attn_wo_distill}), resulting in notable attention degradation. A straightforward mitigation strategy is to perform attention distillation during PTQ. However, for large-scale video generation models (e.g., with \( L > 10^4 \) for HunyuanVideo~\citep{kong2024hunyuanvideo}), storing the full attention matrices is prohibitively expensive as shown in Fig.~\ref{fig:attn_distill_memory}, incurring \( \mathcal{O}(L^2) \) memory and compute overhead. 

To address this, we propose \textit{Multi-Scale Salient Attention Distillation} (MSAD), a memory-efficient framework that distills attention across multiple resolutions, preserving both global structure and local saliency without excessive resource consumption. MSAD employs two complementary guidance mechanisms: \textit{global guidance} for high-level structural supervision, and \textit{local guidance} for fine-grained detail preservation.

\textbf{Global Guidance.} Our approach exploits the intrinsic \textit{locality} of video data: patially adjacent tokens exhibit high similarity due to temporal smoothness and spatial continuity~\citep{ren2025grouping, xi2025svg, yuan2024ditfastattn,feng2024rdd}. To efficiently capture global attention patterns, we downsample \( \mathbf{Q} \) and \( \mathbf{K} \) via average pooling with stride \( s \), producing low-resolution features \( \tilde{\mathbf{Q}}, \tilde{\mathbf{K}} \in \mathbb{R}^{\tilde{L} \times d_k} \) where \( \tilde{L} = L / s^2 \ll L \). The global distillation is computed as:

\begin{equation}
\mathbf{A}_{\text{global}} = \text{softmax}\left( \frac{\tilde{\mathbf{Q}} \tilde{\mathbf{K}}^\top}{\sqrt{d_k}} \right),~\mathcal{L}_{\text{global}} = \text{MSE}\left( \mathbf{A}_{\text{global}}^{\text{FP}} \parallel \mathbf{A}_{\text{global}}^{\text{quant}} \right),
\label{eq:global_distill}
\end{equation}

where MSE denotes the Mean Square Error. This approach requires only \( \mathcal{O}(\tilde{L}^2) \) complexity, which is $s^2$ times cheaper than full attention.

\textbf{Local Guidance.} While global guidance ensures structural fidelity, it fails to capture the fine-grained details crucial for high-quality video synthesis. We further observe that the attention saliency in video models is highly \textit{skewed}: only a small subset of tokens dominates the attention mass (see Fig~\ref{fig:token_saliency}). Formally, we define the token saliency as:
\rebuttal{
\begin{equation}
\mathbf{A} = \text{softmax}(\mathbf{Q}\mathbf{K}^\top / \sqrt{d_k}) \in \mathbb{R}^{h,L,L},~s_j = \sum_{h,i} A_{h,i,j} , 
\label{eq:saliency}
\end{equation}
where $h$ denotes the attention head, $i$ denotes the key token index, and $s_j$ measures the aggregate attention received by token $j$. Empirically, \( s_j \) follows a heavy-tailed distribution, with fewer tokens accounting for the majority of attention mass (\textbf{we provide more analysis in Appendix Sec.~\ref{sec:more_mssa}}). We exploit this by selecting the top-\( k \) queries \( \mathcal{I} = \{ j \mid s_j \text{ is top-}k \} \) from the FP model and computing high-resolution attention \textit{only} for these salient queries:
}

\begin{equation}
\mathbf{A}_{\text{local}} = \text{softmax}\left( \frac{\mathbf{Q}_{\mathcal{I},:} \mathbf{K}^\top}{\sqrt{d_k}} \right),~\mathcal{L}_{\text{local}} = \text{MSE}\left( \mathbf{A}_{\text{local}}^{\text{FP}} \parallel \mathbf{A}_{\text{local}}^{\text{quant}} \right),
\label{eq:local_distill}
\end{equation}

where \( \mathbf{Q}_{\mathcal{I},:} \in \mathbb{R}^{k \times d_k} \). Local distillation focuses supervision on high-impact regions at minimal cost.

\textbf{Integration and Optimization.} We combine both guidance terms into a unified distillation object:

\begin{equation}
\mathcal{L}_{\text{distill}} = \mathcal{L}_{\text{quant}} +  \lambda_{\text{global}} \mathcal{L}_{\text{global}} + \lambda_{\text{local}} \mathcal{L}_{\text{local}},
\label{eq:total_loss}
\end{equation}

where \( \lambda_{\text{global}} \) and \( \lambda_{\text{local}} \) balance the two guidance component. During PTQ calibration, we optimize the quantization parameters over \( \mathcal{D}_{\text{cal}} \) to minimize \( \mathcal{L}_{\text{distill}} \), aligning the quantized attention with its FP counterpart. As shown in Fig~\ref{fig:attn_wo_distill} and Fig~\ref{fig:attn_w_distill}, MSAD substantially reduces attention shift, enabling robust integration of quantization and sparse attention in video generation.

\subsection{Second-Order Sparse Attention Reparameterization}

\rebuttal{While the proposed MSAD mitigates the quantization-induced attention shift during calibration phase by aligning attention maps, the intrinsic bottleneck of sparse attention (i.e., the unavoidable discard of low-magnitude yet non-trivial attention connections) still exacerbates the amplified attention shift, especially under high sparsity rates~\citep{xi2025svg, zhang2025spargeattn}.} We formalize this deviation at denoising timestep $t$ in the diffusion process as: $\Delta^{(t)} = \mathbf{A}_{\text{full}}^{(t)} - \mathbf{A}_{\text{sparse}}^{(t)},$ where $\mathbf{A}_{\text{full}}$ and $\mathbf{A}_{\text{sparse}}$ denote the full-attention and sparse attention. We define this deviation $\Delta^{(t)}$ as the \textit{first-order residual}. This residual is intrinsic to sparsity and cannot be recovered through attention distillation alone. Prior work~\citep{yuan2024ditfastattn} exploits temporal coherence in video generation by assuming that residuals are invariant across timesteps:

\begin{figure}[h]
    \centering
    \subfloat[][Residual temporal difference.]{
        \includegraphics[width=0.31\linewidth]{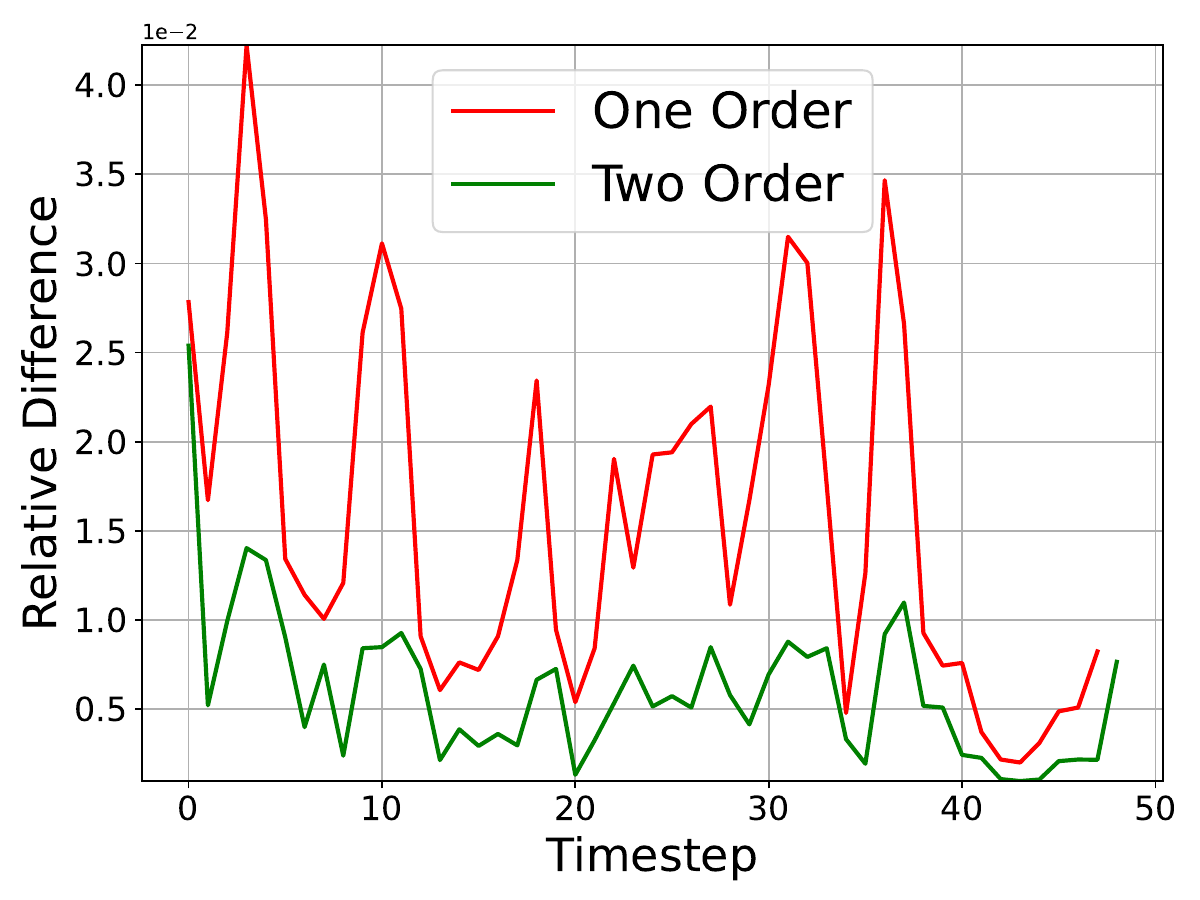}
        \label{fig:residual_temporal}
    }
    \subfloat[][Singular value distribution of all timesteps.]{
        \includegraphics[width=0.31\linewidth]{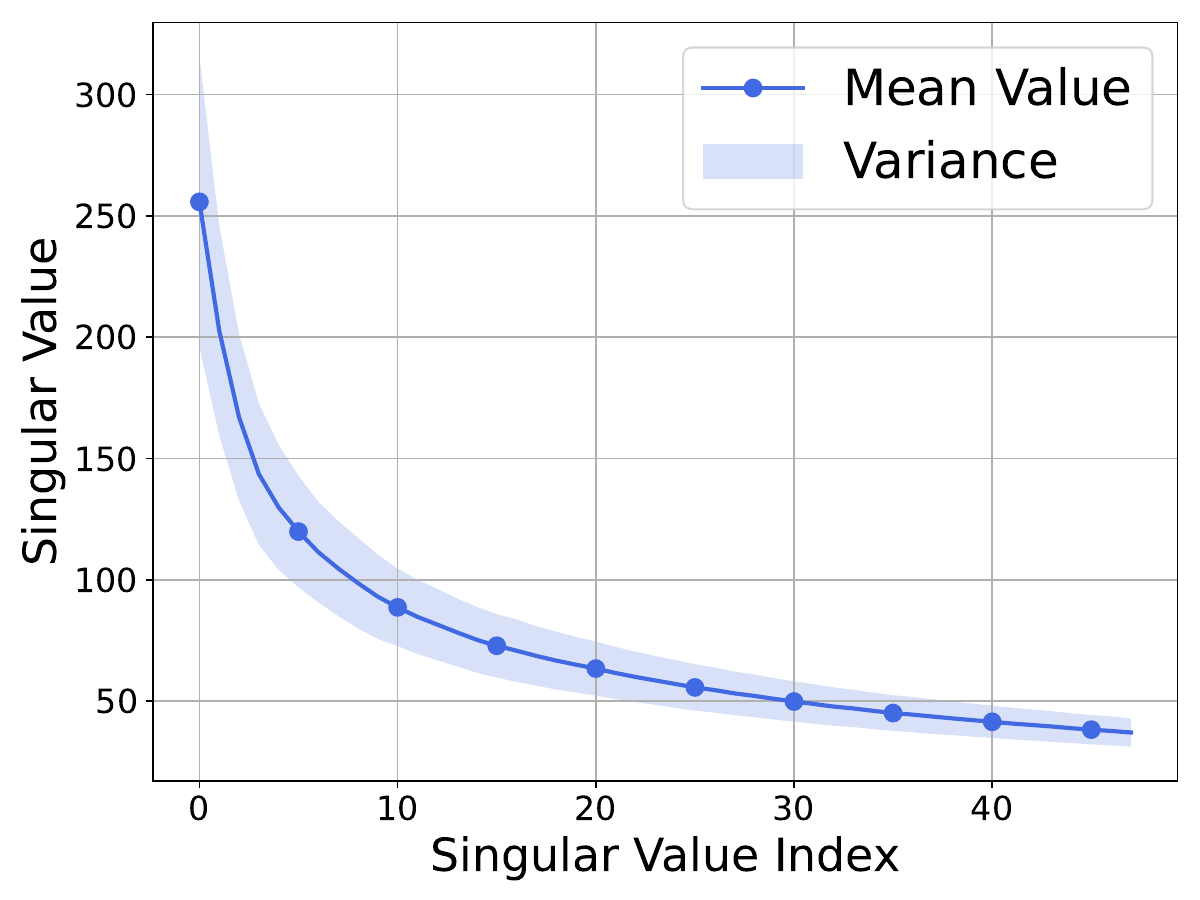}
         \label{fig:svd_distribution}
    }
    \subfloat[][Attention error comparison.]{
        \includegraphics[width=0.31\linewidth]{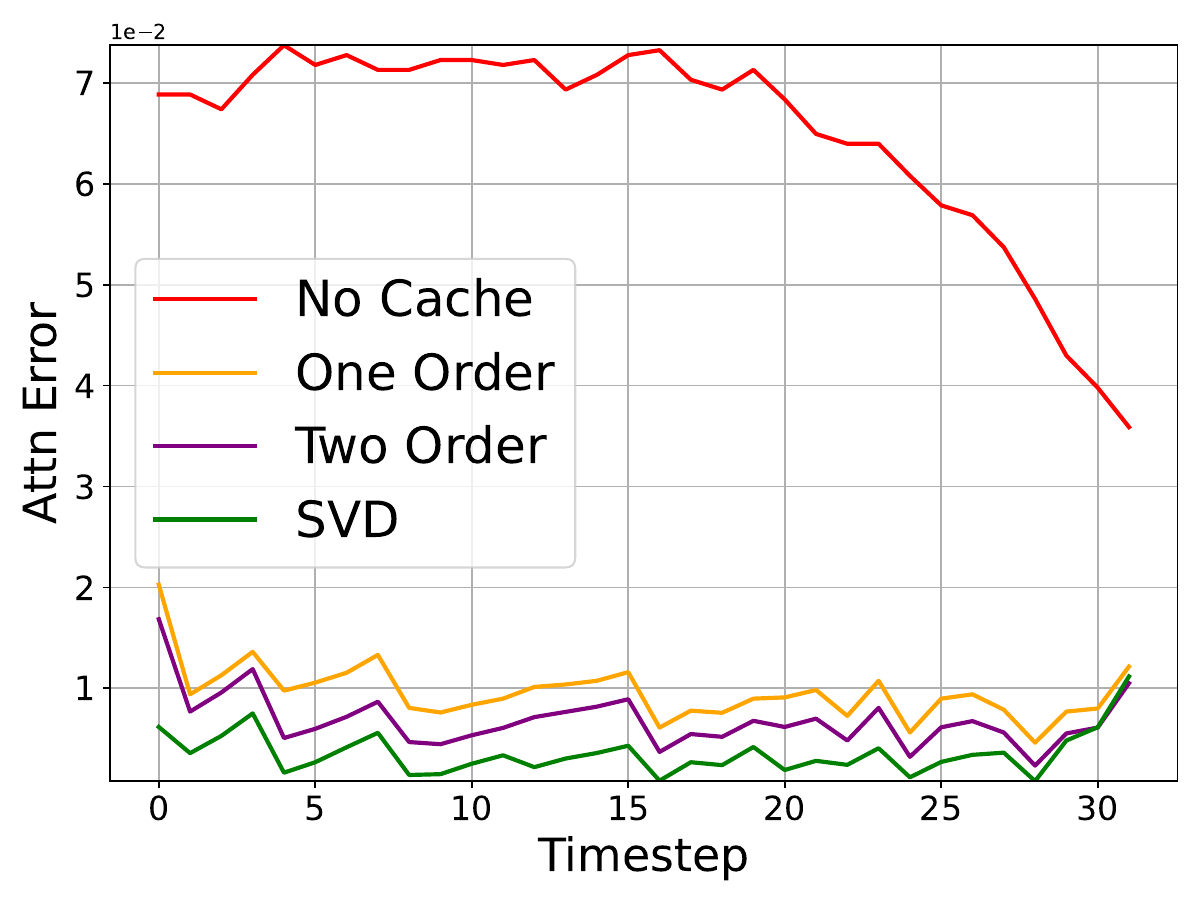}
         \label{fig:attn_error_vis}
    }
    \caption{\textbf{The motivation and effect of Second-Order Sparse Attention Reparameterization.} The results are from HunyuanVideo-13B~\citep{kong2024hunyuanvideo} \textit{single\_transformer\_block.10} under W4A8. \textbf{We provide more visualization and analysis in Appendix Sec.~\ref{sec:more_ssar}}.}
\label{fig:second_order_attn_vis}
\end{figure}

\vspace{-0.2in}

\begin{equation}
\Delta^{(t')} \approx \Delta^{(t)} \quad \forall t, t',
\label{eq:one_order_same}
\end{equation}

Under this assumption, one can cache a reference residual $\Delta^{(t_{\text{ref}})}$ from a chosen timestep and reuse it across the successive timesteps, yielding a \textit{first-order sparse attention reparameterization}:

\begin{equation}
\begin{gathered}
\mathbf{A}_{\text{full}}^{(t)} - \mathbf{A}_{\text{sparse}}^{(t)} \approx \mathbf{A}_{\text{full}}^{(t_{\text{ref}})} - \mathbf{A}_{\text{sparse}}^{(t_{\text{ref}})}=\Delta^{(t_{\text{ref}})} \Rightarrow
\hat{\mathbf{A}}^{(t)} = \mathbf{A}_{\text{sparse}}^{(t)} + \underbrace{\Delta^{(t_{\text{ref}})}}_{\text{cached}},
\end{gathered}
\label{eq:one_order_attn}
\end{equation}


\begin{proposition}
Let \( \mathbf{A}_{\emph{s, q}}^{(t)} \) denote the quantized sparse attention output. The quantization-induced perturbation \( \epsilon^{(t)} \) (as defined in Eq.~\ref{eq:quant_error}) modifies the one-order residual to:  
\begin{equation}
\begin{gathered}
\Delta_{\emph{quant}}^{(t)} = \mathbf{A}_{\emph{full}}^{(t)} - \mathbf{A}_{\emph{s, q}}^{(t)} = \Delta^{(t)} + \epsilon^{(t)} + \mathcal{O}(\|\epsilon^{(t)}\|_F \cdot \|\mathbf{M}\|_0),\\
\Rightarrow \Delta_{\emph{quant}}^{(t')} \neq \Delta_{\emph{quant}}^{(t)}, \quad \emph{for} \quad t' \neq t.
\end{gathered}
\end{equation}
\label{prop:one_order_fail}
\end{proposition}

Proposition~\ref{prop:one_order_fail} indicates that, unlike $\Delta^{(t)}$, $\Delta_{\text{quant}}^{(t)}$ varies with \( \epsilon^{(t)} \) due to the quantization noise~\citep{wu2024ptq4dit, zhao2024vidit, he2023efficientdm} which violating Eq.~\ref{eq:one_order_same}. We visualize this variance of $\Delta^{(t)}-\Delta^{(t-1)}$ in Fig.~\ref{fig:residual_temporal}. This temporal variance undermines the accuracy of Eq.~\ref{eq:one_order_attn}, causing non-negligible attention errors when \textit{first-order reparameterization} is applied after quantization. 

\begin{proposition}
Although \( \Delta_{\emph{quant}}^{(t)} \) is unstable, we observe that the second-order residual \( \hat{\Delta}_{\emph{quant}}^{(t)}:= \Delta_{\emph{quant}}^{(t)} - \Delta_{\emph{quant}}^{(t-1)}\) exhibits significantly higher temporal stability:
\begin{equation}
\mathbb{E}_{t} \left[ \left\| \hat{\Delta}_{\emph{quant}}^{(t)} - \hat{\Delta}_{\emph{quant}}^{(t')} \right\|_F \right] \leq \mathbb{E}_{t} \left[ \left\| \Delta_{\emph{quant}}^{(t)} - \Delta_{\emph{quant}}^{(t')} \right\|_F \right] \quad \emph{for} \quad |t - t'| \leq \tau.
\end{equation}
\label{prop:two_order_good}
\end{proposition}

\begin{table}[t!]
\caption{Text-to-Video generation results on Wan2.1-1.3B. Density is the attention density. Full Prec. denotes Full Precision model. \textbf{Bold}: the best result.}
\label{tab:sora_huge}
\vspace*{-0.1in}
\begin{center}
\resizebox{0.85\linewidth}{!}{
\input{tables/sora_small}}
\end{center}
\vspace*{-0.2in}
\end{table}

We visualize the empirical analysis results in Fig.~\ref{fig:residual_temporal}. This stability arises because quantization noise \( \epsilon^{(t)} \) follows a \textit{slow-varying stochastic process} in diffusion process~\citep{ma2024deepcache, liu2024teacache}: adjacent timesteps share similar distributions, rendering \( \epsilon^{(t)} - \epsilon^{(t-1)} \) approximately stationary. Leveraging this property, we propose \textit{second-order sparse attention reparameterization}:

\begin{equation}
\begin{aligned}
(\mathbf{A}_{\text{full}}^{(t)} - \mathbf{A}_{\text{s,q}}^{(t)}) - (\mathbf{A}_{\text{full}}^{(t_{\text{ref}})} - &\mathbf{A}_{\text{s,q}}^{(t_{\text{ref}})}) \approx (\mathbf{A}_{\text{full}}^{(t_{\text{ref}})} - \mathbf{A}_{\text{s,q}}^{(t_{\text{ref}})}) - (\mathbf{A}_{\text{full}}^{(t_{\text{ref}}')} - \mathbf{A}_{\text{s,q}}^{(t_{\text{ref}}')}) = \hat{\Delta}_{\text{quant}}^{(t_{\text{ref}})}, \\
\Rightarrow \tilde{\mathbf{A}}^{(t)} &= \mathbf{A}_{\text{s,q}}^{(t)} + (\mathbf{A}_{\text{full}}^{(t_{\text{ref}})} - \mathbf{A}_{\text{s,q}}^{(t_{\text{ref}})}) + \hat{\Delta}_{\text{quant}}^{(t_{\text{ref}})}, \\
&= \mathbf{A}_{\text{s,q}}^{(t)} +\underbrace{\Delta_{\text{quant}}^{(t_{\text{ref}})} + \hat{\Delta}_{\text{quant}}^{(t_{\text{ref}})}}_{\text{cached}}.
\end{aligned}
\label{eq:two_order_attn}
\end{equation}

\begin{theorem}
When Proposition~\ref{prop:two_order_good} holds, the expected approximation error of sparse attention satisfies:

\begin{equation}
\begin{gathered}
\mathbb{E}_{t} \underbrace{ \left[ \left\| \mathbf{A}_{\emph{full}}^{(t)} - \tilde{\mathbf{A}}^{(t)}_{\emph{s,q}} \right\|_F \right]}_{\emph{second-order}} \leq \mathbb{E}_{t} \underbrace{\left[ \left\| \mathbf{A}_{\emph{full}}^{(t)} - \hat{\mathbf{A}}^{(t)}_{\emph{s,q}} \right\|_F \right]}_{\emph{first-order}} \quad \emph{for} \quad |t - t'| \leq \tau.
\end{gathered}
\end{equation}

\label{theorem:less_error}
\end{theorem}

Theorem~\ref{theorem:less_error} indicates \textit{two-order} guaranteeing tighter full-attention approximation than the first-order method. Also $\Delta_{\text{quant}}^{(t_{\text{ref}})} + \hat{\Delta}_{\text{quant}}^{t_{\text{ref}}}$ in Eq.~\ref{eq:two_order_attn} can be jointly cached, without any additional storage burden compared with one-order residual. We further reduce the temporal variance of $\hat{\Delta}_{\text{quant}}$ by projecting it onto its most stable subspace. Empirically, the top-$r$ principal components from the singular value decomposition (SVD) of $\hat{\Delta}_{\text{quant}}$ capture the dominant, temporally stable patterns (see Fig.~\ref{fig:svd_distribution}). Critically, the dominant principal component exhibit exceptional temporal stability, which inspired us to project residuals onto the top-\( r \) extracted stable components:

\begin{equation}
\begin{gathered}
\text{SVD}(\hat{\Delta}_{\text{quant}}) = \mathbf{S} \mathbf{U} \mathbf{V}^\top,~\tilde{\Delta}_{\text{quant}} := \mathbf{S}_{:,:r} \mathbf{U}_{:r,:r} \mathbf{V}^\top_{:,:r}, \\
\tilde{\mathbf{A}}^{(t)} = \mathbf{A}_{\text{s,q}}^{(t)} +\underbrace{\Delta_{\text{quant}}^{(t_{\text{ref}})} + \tilde{\Delta}_{\text{quant}}^{t_{\text{ref}}}}_{\text{cached}}.
\end{gathered}
\label{eq:final_second_order}
\end{equation}

We apply the sparse attention for inference with a fixed cache-refreshing interval ($5$ in experiments) for full-attention calculation.  As visualized in Fig.~\ref{fig:attn_error_vis}, SVD-based second-order reparameterization further suppresses temporal variance, yielding accurate full-attention approximation results.

\rebuttal{
\subsection{Overall pipeline}
Our proposed QuantSparse framework consists of two component as shown in Fig.~\ref{fig:overview}: \textit{MSAD} for attention distillation during calibration and \textit{SSAR} for dynamic attention reparameterization during inference. The detailed overall pipeline is provided in Appendix Algorithm~\ref{alg:quantsparse_pipeline}.
}

%% file: tables/sora_small.tex
\begin{tabular}{lcccccccc}
\toprule
\multirow{3}{*}{\textbf{Method}} & \multirow{3}{*}{\textbf{\makecell{\#Bits \\ (W/A)}}} & \multirow{3}{*}{\textbf{Density}$_{\mathbf{\red{\downarrow}}}$} & \multicolumn{6}{c}{\textbf{Quality}} \\
\cmidrule(lr){4-9}  
& & & \multicolumn{3}{c}{\textit{Video Quality Metrics}} & \multicolumn{3}{c}{\textit{FP Diff. Metrics}} \\
\cmidrule(lr){4-6}
\cmidrule(lr){7-9} 
& & & CLIPSIM$_{\mathbf{\red{\uparrow}}}$ & VQA$_{\mathbf{\red{\uparrow}}}$ & $\Delta$FSCore$_{\mathbf{\red{\downarrow}}}$ & PSNR$_{\mathbf{\red{\uparrow}}}$ & SSIM$_{\mathbf{\red{\uparrow}}}$ & LPIPS$_{\mathbf{\red{\downarrow}}}$ \\
\midrule  

\multicolumn{9}{c}{\cellcolor[gray]{0.92}\texttt{Wan}2.1 1.3B ($\texttt{CFG}=6.0, 480\times832p, \texttt{frames}=80$)} \\
\midrule

Full Prec. & 16/16 & 100\% & 0.191 & 73.12 & 0.000 & - & - & - \\
\midrule

PTQ4DiT & 6/6 & 100\% & 0.182 & 36.79 & 2.287 & 10.20 & 0.343 & 0.598 \\
Q-DiT & 6/6 & 100\% & 0.183 & 39.21 & 2.125 & 10.36 & 0.351 & 0.577 \\
SmoothQuant & 6/6 & 100\% & 0.184 & 40.57 & 2.008 & 10.44 & 0.353 & 0.574 \\
QuaRot & 6/6 & 100\% & 0.190 & 42.81 & 1.754 & 10.71 & 0.379 & 0.571 \\
ViDiT-Q & 6/6 & 100\% & 0.190 & 50.85 & 1.253 & 11.02 & 0.385 & 0.526 \\
Q-VDiT & 6/6 & 100\% & 0.191 & 75.20 & 0.982 & 12.06 & 0.405 & 0.474 \\
QuaRot+DFT & 6/6 & 40\% & 0.183 & 36.79 & 2.297 & 11.29 & 0.321 & 0.546 \\
QuaRot+Jenga & 6/6 & 40\% & 0.184 & 38.78 & 2.165 & 11.32 & 0.329 & 0.543 \\
QuaRot+SVG & 6/6 & 40\% & 0.183 & 41.93 & 1.940 & 11.43 & 0.331 & 0.541 \\
Q-VDiT+DFT & 6/6 & 40\% & 0.188 & 47.33 & 1.377 & 11.06 & 0.345 & 0.577 \\
Q-VDiT+Jenga & 6/6 & 40\% & 0.189 & 53.52 & 1.087 & 11.21 & 0.345 & 0.583 \\
Q-VDiT+SVG & 6/6 & 40\% & 0.191 & 55.92 & 0.942 & 11.61 & 0.384 & 0.508 \\
\rowcolor{mycolor!30}\textbf{QuantSparse} & 6/6 & 40\% & \textbf{0.193} & \textbf{78.35} & \textbf{0.055} & \textbf{15.51} & \textbf{0.511} & \textbf{0.324} \\
\midrule

PTQ4DiT & 4/8 & 100\% & 0.181 & 30.26 & 2.574 & 10.00 & 0.318 & 0.603 \\
Q-DiT & 4/8 & 100\% & 0.182 & 32.57 & 2.767 & 10.11 & 0.320 & 0.594 \\
SmoothQuant & 4/8 & 100\% & 0.182 & 34.82 & 2.174 & 10.20 & 0.327 & 0.569 \\
QuaRot & 4/8 & 100\% & 0.185 & 65.15 & 1.870 & 11.72 & 0.349 & 0.514 \\
ViDiT-Q & 4/8 & 100\% & 0.186 & 63.21 & 1.698 & 11.24 & 0.351 & 0.526 \\
Q-VDiT & 4/8 & 100\% & 0.190 & 56.45 & 2.240 & 11.01 & 0.394 & 0.565 \\
QuaRot+DFT & 4/8 & 40\% & 0.187 & 32.23 & 2.329 & 10.32 & 0.360 & 0.583 \\
QuaRot+Jenga & 4/8 & 40\% & 0.191 & 32.83 & 2.148 & 10.33 & 0.346 & 0.578 \\
QuaRot+SVG & 4/8 & 40\% & 0.190 & 32.48 & 2.088 & 10.58 & 0.370 & 0.576 \\
Q-VDiT+DFT & 4/8 & 40\% & 0.185 & 45.60 & 2.907 & 10.03 & 0.331 & 0.594 \\
Q-VDiT+Jenga & 4/8 & 40\% & 0.185 & 47.61 & 3.000 & 10.04 & 0.334 & 0.596 \\
Q-VDiT+SVG & 4/8 & 40\% & 0.184 & 51.84 & 3.035 & 10.07 & 0.342 & 0.592 \\
\rowcolor{mycolor!30}\textbf{QuantSparse} & 4/8 & 40\% & \textbf{0.193} & \textbf{81.09} & \textbf{0.576} & \textbf{15.22} & \textbf{0.502} & \textbf{0.338} \\

\bottomrule
\end{tabular}

%% file: sec/4_experiments.tex
\section{Experiments}

\begin{table}[t!]
\caption{Video generation on large video generation models. \textbf{Bold}: the best result. \underline{Underline}: the second best result.}
\label{tab:sora_huge}
\vspace*{-0.1in}
\begin{center}
\resizebox{0.97\linewidth}{!}{
\input{tables/sora_wan}}
\end{center}
\vspace*{-0.25in}
\end{table}

\subsection{Experimental and Evaluation Settings}
\label{sec:main_detail}


\textbf{Evaluation Settings.} We apply QuantSparse to HunyuanVideo-13B~\citep{kong2024hunyuanvideo}, Wan2.1-1.3B and 14B~\citep{wan2025wan} with 50 sampling steps. We employ two types of metrics: (1) Multi-aspects metrics evaluation: including CLIPSIM~\citep{wu2021godiva}, VQA~\citep{wu2023dover}, FlowScore~\citep{liu2024evalcrafter}, PSNR, SSIM, and LPIPS~\citep{zhang2018lpips}. All metrics are evaluated on the prompt sets used in~\citep{zhao2024vidit, feng2025qvdit} (2) Benchmark evaluation: We select 8 major dimensions from Vbench~\citep{huang2024vbench} following prior works~\citep{zhao2024vidit, chen2024qdit, feng2025qvdit}. For bit setting, we use W6A6 and W4A8 following prior work~\citep{zhao2024vidit, chen2024qdit, wu2024ptq4dit}, since they can bring more compression effects and ensure the performance.

\textbf{Baseline Methods.}
We select PTQ4DiT~\citep{wu2024ptq4dit}, Q-DiT~\citep{chen2024qdit}, ViDiT-Q~\citep{zhao2024vidit}, and Q-VDiT~\citep{feng2025qvdit} for diffusion baseline. We also compare with strong LLM baseline SmoothQuant~\citep{xiao2023smoothquant} and QuaRot~\citep{ashkboos2024quarot}. For sparsification, we compare with DiTFastAttn (DFT)~\citep{yuan2024ditfastattn} (cache-based), Jenga~\citep{zhang2025jenga} (dynamic pattern), and SparseVideoGen (SVG)~\citep{xi2025svg} (static pattern).

\textbf{Implementation Detail.} Same with prior works~\citep{zhao2024vidit, ashkboos2024quarot, feng2025qvdit}, we adopt channel-wise weight quantization and dynamic token-wise activation quantization. We follow block-wise post-training strategy used in~\citep{wu2024ptq4dit, chen2024qdit, sun2024flatquant} for calibration. \textbf{More details can be found in Appendix~\ref{sec:more_expe_detail}}.


\subsection{Main Results}

\begin{wraptable}{r}{0.55\linewidth}
  \centering
  \begin{minipage}[b]{\linewidth}
    \caption{Ablation results of each component.}
      \resizebox{\linewidth}{!}{
      \input{tables/ablation}}
        \label{tab:ablation}
    \end{minipage}
     \vspace{-0.3in}
\end{wraptable}

\begin{figure}[h]
  \centering
  \includegraphics[width=1.0\linewidth]{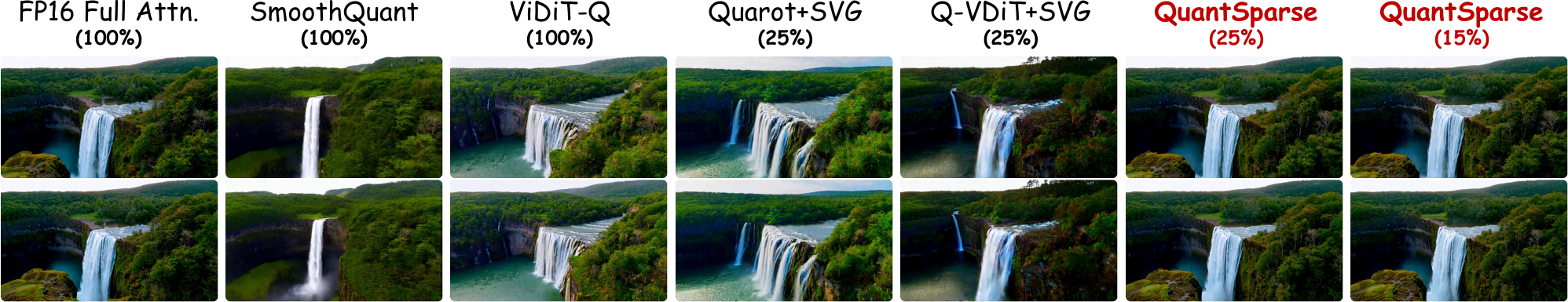}
  \caption{Visual comparison on Wan2.1-14B under W4A8 quantization setting. We uniformly sample two frames for visualization. `(xx\%)' denotes the attention density.}
  \label{fig:compare}
\end{figure}

We present multi-aspects metrics evaluation results on HunyuanVideo~\citep{kong2024hunyuanvideo} and Wan2.1-14B~\citep{wan2025wan} in Tab.~\ref{tab:sora_huge}. It can be seen that the existing SOTA quantization methods have a significant performance degradation after applying sparse attention. But QuantSparse still maintains high generation performance even at high sparsity. It is worth mentioning that QuantSparse even surpasses the existing quantization-only methods under the low-bit settings of W6A6 and W4A8. Compared with the Full-Precision (FP) model, QuantSparse even maintains almost lossless performance. For example, for HunyuanVideo under W6A6, QuantSparse achieved a VQA score of 82.26 with only 15\% attention density, far exceeding current SOTA method Q-VDiT~\citep{feng2025qvdit} of 73.68, and even surpassing the FP model of 81.23. \textbf{We present more baseline methods comparison in Appendix Sec.~\ref{sec:more_eval}, and comprehensive VBench evaluation results in Appendix Sec.~\ref{sec:vbench}.} \rebuttal{We also observed that QuantSparse slightly outperforms Full Precision model on certain metrics. This slight outperformance of QuantSparse can be attributed to its focus on task-critical tokens and reduced attention to noisy or irrelevant tokens, as shown in our saliency analysis. Additionally, the SSAR module stabilizes sparse attention, reducing quantization noise and improving temporal consistency. These effects, combined with targeted compression, allow QuantSparse to maintain near-lossless quality while offering substantial compression and acceleration.} We also visualized the generated videos in Fig.~\ref{fig:compare}. Compared with FP model, QuantSparse achieves almost lossless generation performance while other methods have notable quality degradation. \textbf{We provide more visual comparison results in Appendix Sec.~\ref{sec:more_visual}.}

\begin{table}[t!]
\caption{\rebuttal{Detailed efficiency comparison.}}
\label{tab:efficiency}
\vspace*{-0.1in}
\begin{center}
\resizebox{0.9\linewidth}{!}{
\input{tables/efficiency}}
\end{center}
\vspace*{-0.2in}
\end{table}

\subsection{Ablation Study}

We conduct ablation study on proposed Multi-Scale Salient Attention Distillation (MSAD) and Second-Order Sparse Attention Reparameterization (SSAR) on Wan2.1-14B under W4A8 in Tab.~\ref{tab:ablation}.

\textbf{Effect of attention distillation.}
Compared with no distillation, both proposed attention guidance can enhance the model performance. The combined MSAD further improves PSNR from 14.35 to 18.72, demonstrating the effect of attention distillation.

\textbf{Effect of attention reparameterization.}
Compared with naive sparse attention, first-order residual can reduce the attention error, demonstrating the effectiveness of attention reparameterization. Our proposed SSAR achieves the best approximation performance by reducing both the quantization-induced error and temporal variance.

\rebuttal{\textbf{Effect of cache-interval.} We also supplement the ablation and the results are shown in Tab.~\ref{tab:abla_interval}. While shorter intervals yield higher PSNR and SSIM, indicating better performance, they also result in a reduced speedup (1.65$\times$ and 1.69$\times$ respectively). For instance, interval=3 achieves the highest PSNR (18.86) but sacrifices a noticeable amount of the potential speedup (9\%). Longer intervals increasing the interval to 6 provides a slightly higher speedup (1.76$\times$). However, this comes at the cost of a degradation in performance (PSNR drops to 17.72). We choose interval=5 is based on its optimal balance between model performance and inference speedup. But we highlight that this is a trade-off based on computational resource and all interval settings offer reasonable results and notable acceleration.
}

\textbf{More ablation study about pooling stride $s$, salient token $k$, weight factor $\lambda$, and SVD rank $r$ in Eq.~\ref{eq:total_loss} and Eq.~\ref{eq:final_second_order} in Appendix Sec.~\ref{sec:more_ablation}.}

\subsection{Efficiency Analysis}

\begin{wraptable}{r}{0.5\linewidth}
    \vspace{-0.2in}
  \centering
  \begin{minipage}[b]{\linewidth}
    \caption{\rebuttal{Ablation study of cache-fresh interval and attention density on W4A8 Wan2.1-14B.}}
      \resizebox{\linewidth}{!}{
      \begin{tabular}{c|cccc}
\toprule
- & PSNR$_{\mathbf{\red{\uparrow}}}$ & SSIM$_{\mathbf{\red{\uparrow}}}$ & LPIPS$_{\mathbf{\red{\downarrow}}}$ & Speedup$_{\mathbf{\red{\uparrow}}}$ \\
\midrule  

\multicolumn{5}{c}{\cellcolor[gray]{0.92}Interval Analysis} \\
\midrule

Interval=3 & 18.86 & 0.631 & 0.243 & 1.65$\times$ \\
Interval=4 & 18.48 & 0.617 & 0.260 & 1.69$\times$ \\
Interval=5 & 18.22 & 0.605 & 0.272 & 1.74$\times$ \\
Interval=6 & 17.72 & 0.566 & 0.321 & 1.76$\times$ \\
\midrule

\multicolumn{5}{c}{\cellcolor[gray]{0.92}Density Analysis} \\
\midrule

Density=25\% & 18.72 & 0.630 & 0.240 & 1.55$\times$ \\
Density=20\% & 18.45 & 0.622 & 0.252 & 1.63$\times$ \\
Density=15\% & 18.22 & 0.605 & 0.272 & 1.74$\times$ \\
Density=10\% & 17.73 & 0.589 & 0.288 & 1.80$\times$ \\

\bottomrule
\end{tabular}}
        \label{tab:abla_interval}
    \end{minipage}
     \vspace{-0.3in}
\end{wraptable}

We present the deployment efficiency in Tab.~\ref{tab:efficiency}. \rebuttal{All the experiments are conducted on a single NVIDIA A800 80G GPU with CUDA 12.4. We use CUTLASS~\citep{cutlass} on top of PyTorch for performing INT matrix multiplication.} Existing quantization methods can bring higher model compression, but the effect of inference acceleration is limited. Sparse attention brings significant acceleration, but has almost no model compression, and even brings more memory consumption. QuantSparse combines the advantages of both quantization and sparse attention, bringing significant model compression and acceleration. For Wan2.1-14B~\citep{wan2025wan}, QuantSparse (15\% density) brings 3.80$\times$ storage compression, 1.51$\times$ memory saving, and 1.74$\times$ end-to-end acceleration. \textbf{We further report the calibration resource consumption in Appendix Sec.~\ref{sec:calib_efficiency} and report the performance combined with other acceleration methods in Appendix Sec.~\ref{sec:combine_more_accele}.}

%% file: tables/sora_wan.tex
\begin{tabular}{l|c|c|cccccccc}
\toprule
\multirow{3}{*}{\textbf{Method}} & \multirow{3}{*}{\textbf{\makecell{\#Bits \\ (W/A)}}} & \multirow{3}{*}{\textbf{Density}$_{\mathbf{\red{\downarrow}}}$} & \multicolumn{6}{c}{\textbf{Quality}} & \multicolumn{2}{c}{\textbf{Latency \& Speed}} \\
\cmidrule(lr){4-9}
\cmidrule(lr){10-11}
& & & \multicolumn{3}{c}{\textit{Video Quality Metrics}} & \multicolumn{3}{c}{\textit{FP Diff. Metrics}} & \multirow{2}{*}{DiT Time$_{\mathbf{\red{\downarrow}}}$} & \multirow{2}{*}{Speedup$_{\mathbf{\red{\uparrow}}}$} \\
\cmidrule(lr){4-6}
\cmidrule(lr){7-9} 
& & & CLIPSIM$_{\mathbf{\red{\uparrow}}}$ & VQA$_{\mathbf{\red{\uparrow}}}$ & $\Delta$FSCore$_{\mathbf{\red{\downarrow}}}$ & PSNR$_{\mathbf{\red{\uparrow}}}$ & SSIM$_{\mathbf{\red{\uparrow}}}$ & LPIPS$_{\mathbf{\red{\downarrow}}}$ &  \\
\midrule  

\multicolumn{11}{c}{\cellcolor[gray]{0.92}HunyuanVideo 13B ($\texttt{CFG}=6.0, 720\times1280p, \texttt{frames}=60$)} \\
\midrule

Full Prec. & 16/16 & 100\% & 0.184 & 81.23 & 0.000 & - & - & - & 1264s & 1.00$\times$ \\
\midrule

SmoothQuant & 4/8 & 100\% & 0.178 & 42.21 & 1.194 & 15.44 & 0.479 & 0.583 & 1149s & 1.10$\times$ \\
QuaRot & 4/8 & 100\% & 0.180 & 42.89 & 0.708 & 15.46 & 0.502 & 0.528 & 1149s & 1.10$\times$ \\
ViDiT-Q & 4/8 & 100\% & 0.181 & 49.82 & 1.254 & 15.75 & 0.534 & 0.489 & 1149s & 1.10$\times$ \\
Q-VDiT & 4/8 & 100\% & 0.182 & 67.95 & 1.168 & 16.85 & 0.605 & 0.461 & 1155s & 1.09$\times$ \\
QuaRot+SVG & 4/8 & 25\% & 0.181 & 43.34 & 0.900 & 15.39 & 0.497 & 0.530 & 731s & 1.73$\times$ \\
Q-VDiT+SVG & 4/8 & 25\% & 0.182 & 70.99 & 1.379 & 16.71 & 0.595 & 0.458 & 743s & 1.70$\times$ \\
QuaRot+SVG & 4/8 & 15\% & 0.181 & 41.40 & 1.004 & 15.34 & 0.494 & 0.536 & 671s & 1.88$\times$ \\
Q-VDiT+SVG & 4/8 & 15\% & 0.182 & 76.30 & 1.393 & 16.66 & 0.591 & 0.460 & 687s & 1.84$\times$ \\
\rowcolor{mycolor!30}\textbf{QuantSparse} & 4/8 & 25\% & \underline{0.183} & \underline{79.05} & \textbf{0.014} & \underline{20.86} & \underline{0.675} & \textbf{0.272} & 731s & 1.73$\times$ \\
\rowcolor{mycolor!30}\textbf{QuantSparse} & 4/8 & 15\% & \textbf{0.184} & \textbf{81.19} & \underline{0.016} & \textbf{20.88} & \textbf{0.678} & \underline{0.273} & 671s & 1.88$\times$ \\
\midrule

\multicolumn{11}{c}{\cellcolor[gray]{0.92}\texttt{Wan}2.1 14B ($\texttt{CFG}=5.0, 720\times1280p, \texttt{frames}=80$)} \\
\midrule

Full Prec. & 16/16 & 100\% & 0.182 & 90.79 & 0.000 & - & - & - & 4031s & 1.00$\times$ \\
\midrule

SmoothQuant & 4/8 & 100\% & 0.180 & 73.11 & 0.875 & 13.70 & 0.423 & 0.510 & 3425s & 1.18$\times$ \\
QuaRot & 4/8 & 100\% & \underline{0.182} & 85.91 & 0.753 & 13.79 & 0.431 & 0.494 & 3425s & 1.18$\times$ \\
ViDiT-Q & 4/8 & 100\% & \underline{0.182} & 83.13 & 0.496 & 15.12 & 0.487 & 0.425 & 3425s & 1.18$\times$ \\
Q-VDiT & 4/8 & 100\% & \underline{0.182} & 83.76 & 0.343 & 15.85 & 0.512 & 0.398 & 3457s & 1.17$\times$ \\
QuaRot+SVG & 4/8 & 25\% & \underline{0.182} & 85.66 & 0.134 & 13.70 & 0.427 & 0.487 & 2594s & 1.55$\times$ \\
Q-VDiT+SVG & 4/8 & 25\% & \underline{0.182} & 87.89 & 0.310 & 15.48 & 0.507 & 0.409 & 2635s & 1.53$\times$ \\
QuaRot+SVG & 4/8 & 15\% & \underline{0.182} & 81.93 & 0.152 & 13.40 & 0.415 & 0.494 & 2315s & 1.74$\times$ \\
Q-VDiT+SVG & 4/8 & 15\% & 0.181 & 82.31 & 0.411 & 15.18 & 0.493 & 0.429 & 2372s & 1.70$\times$ \\
\rowcolor{mycolor!30}\textbf{QuantSparse} & 4/8 & 25\% & \textbf{0.183} & \textbf{91.98} & \underline{0.056} & \textbf{18.72} & \textbf{0.630} & \textbf{0.240} & 2594s & 1.55$\times$ \\
\rowcolor{mycolor!30}\textbf{QuantSparse} & 4/8 & 15\% & \underline{0.182} & \underline{90.73} & \textbf{0.042} & \underline{18.22} & \underline{0.605} & \underline{0.272} & 2315s & 1.74$\times$ \\
 
\bottomrule
\end{tabular}

%% file: tables/ablation.tex
\begin{tabular}{l|llll}
\toprule
\textbf{Method} & VQA$_{\mathbf{\red{\uparrow}}}$ &  PSNR$_{\mathbf{\red{\uparrow}}}$ & SSIM$_{\mathbf{\red{\uparrow}}}$ & LPIPS$_{\mathbf{\red{\downarrow}}}$ \\
\midrule  

\multicolumn{5}{c}{\cellcolor[gray]{0.92} Distillation Analysis} \\
None & 81.92 & 14.35 & 0.486 & 0.425 \\
\textit{Global} & 85.26 & 16.01 & 0.547 & 0.349 \\
\textit{Local} & 86.95 & 16.82 & 0.561 & 0.325 \\
\rowcolor{mycolor!30} \textbf{MSAD} & \textbf{91.98}$_{\mathbf{\red{+10.06}}}$ & \textbf{18.72}$_{\mathbf{\red{+4.37}}}$ & \textbf{0.630}$_{\mathbf{\red{+0.144}}}$ & \textbf{0.240}$_{\mathbf{\red{-0.185}}}$ \\
\midrule

\multicolumn{5}{c}{\cellcolor[gray]{0.92} Cache Analysis} \\
None & 68.00 & 14.16 & 0.470 & 0.445 \\
\textit{First} & 70.82 & 17.08 & 0.572 & 0.285 \\
\textit{Second} & 89.73 & 18.68 & 0.616 & 0.258 \\
\rowcolor{mycolor!30} \textbf{SSAR} & \textbf{91.98}$_{\mathbf{\red{+23.98}}}$ & \textbf{18.72}$_{\mathbf{\red{+4.56}}}$ & \textbf{0.630}$_{\mathbf{\red{+0.160}}}$ & \textbf{0.240}$_{\mathbf{\red{-0.205}}}$ \\
\bottomrule
\end{tabular}

%% file: tables/efficiency.tex
\begin{tabular}{l|c|c|llcc}
\toprule
\multirow{2}{*}{\textbf{Method}} & \multirow{2}{*}{\textbf{\makecell{\#Bits \\ (W/A)}}} & \multirow{2}{*}{\textbf{Density$_{\mathbf{\red{\downarrow}}}$}} & \multicolumn{2}{c}{\textbf{Model Overload}} & \multicolumn{2}{c}{\textbf{Latency \& Speed}} \\
\cmidrule(lr){4-5}
\cmidrule(lr){6-7}
& & & Model Storage$_{\mathbf{\red{\downarrow}}}$ & Memory Consumption$_{\mathbf{\red{\downarrow}}}$ & DiT Time$_{\mathbf{\red{\downarrow}}}$ & Speedup$_{\mathbf{\red{\uparrow}}}$ \\
\midrule  

\multicolumn{7}{c}{\cellcolor[gray]{0.92}HunyuanVideo 13B ($\texttt{CFG}=6.0, 720\times1280p, \texttt{frames}=60$)} \\
\midrule

Full Prec. & 16/16 & 100\% & 23.88GB & 35.79GB & 1264s & 1.00$\times$ \\
\midrule

QuaRot & 4/8 & 100\% & 6.49GB & 24.34GB & 1149s & 1.10$\times$  \\
Q-VDiT & 4/8 & 100\% & 6.50GB & 24.89GB & 1155s & 1.09$\times$  \\
DFT & 16/16 & 25\% & 23.88GB & 40.11GB & 792s & 1.60$\times$ \\
Jenga & 16/16 & 25\% & 23.88GB & 36.92GB & 846s & 1.49$\times$ \\
SVG & 16/16 & 25\% & 23.88GB & 40.10GB & 786s & 1.61$\times$ \\
SVG & 16/16 & 15\% & 23.88GB & 40.10GB & 707s & 1.79$\times$ \\
\rowcolor{mycolor!30}\textbf{QuantSparse} & 4/8 & 25\% & 6.49GB$_{\mathbf{\red{\downarrow3.68\times}}}$ & 27.02GB$_{\mathbf{\red{\downarrow1.32\times}}}$ & 731s & 1.73$\times$ \\
\rowcolor{mycolor!30}\textbf{QuantSparse} & 4/8 & 15\% & 6.49GB$_{\mathbf{\red{\downarrow3.68\times}}}$ & 27.02GB$_{\mathbf{\red{\downarrow1.32\times}}}$ & 671s & 1.88$\times$ \\
\midrule

\multicolumn{7}{c}{\cellcolor[gray]{0.92}\texttt{Wan}2.1 14B ($\texttt{CFG}=5.0, 720\times1280p, \texttt{frames}=80$)} \\
\midrule

Full Prec. & 16/16 & 100\% & 26.61GB & 42.48GB & 4031s & 1.00$\times$ \\
\midrule

QuaRot & 4/8 & 100\% & 7.00GB & 26.04GB & 3425s & 1.18$\times$ \\
Q-VDiT & 4/8 & 100\% & 7.02GB & 26.73GB & 3457s & 1.17$\times$ \\
DFT & 16/16 & 25\% & 26.61GB & 44.86GB & 3015s & 1.34$\times$ \\
Jenga & 16/16 & 25\% & 26.61GB & 42.62GB & 3087s & 1.31$\times$ \\
SVG & 16/16 & 25\% & 26.61GB & 44.07GB & 2987s & 1.35$\times$ \\
SVG & 16/16 & 15\% & 26.61GB & 44.07GB & 2661s & 1.51$\times$ \\
\rowcolor{mycolor!30}\textbf{QuantSparse} & 4/8 & 25\% & 7.00GB$_{\mathbf{\red{\downarrow3.80\times}}}$ & 28.14GB$_{\mathbf{\red{\downarrow1.51\times}}}$ & 2594s & 1.55$\times$ \\
\rowcolor{mycolor!30}\textbf{QuantSparse} & 4/8 & 15\% & 7.00GB$_{\mathbf{\red{\downarrow3.80\times}}}$ & 28.14GB$_{\mathbf{\red{\downarrow1.51\times}}}$ & 2315s & 1.74$\times$ \\

\bottomrule
\end{tabular}

%% file: sec/5_conclusion.tex
\section{Conclusion}

In this paper, we propose QuantSparse, a unified 
compression framework that effectively combines model quantization and sparse attention. To address the amplified attention shift, we propose Multi-Scale Salient Attention Distillation to efficiently align the attention shift. To address the intrinsic sparsity loss, we propose Second-Order Sparse Attention Reparameterization to utilize decomposed second-order residual for attention approximation. Extensive experiments shown that QuantSparse achieves lossless performance while bringing significant model compression and acceleration. 

%% file: sec/6_appendix.tex
\section{Proof of Theorem~\ref{theorem:less_error}.}
\label{sec:proof}

\textit{Proof of Theorem~\ref{theorem:less_error}.}

For $\tilde{\mathbf{A}}^{(t)}_{\text{s,q}}$, we have:

\begin{equation}
\begin{aligned}
(\mathbf{A}_{\text{full}}^{(t)} - &\mathbf{A}_{\text{s,q}}^{(t)}) - (\mathbf{A}_{\text{full}}^{(t_{\text{ref}})} - \mathbf{A}_{\text{s,q}}^{(t_{\text{ref}})}) = \hat{\Delta}_{\text{quant}}^{t} \\
\Rightarrow \mathbf{A}_{\text{full}}^{(t)} &= \mathbf{A}^{(t)}_{\text{s,q}} + (\mathbf{A}_{\text{full}}^{(t_{\text{ref}})} - \mathbf{A}_{\text{s,q}}^{(t_{\text{ref}})}) + \hat{\Delta}_{\text{quant}}^{t} \\
&= \mathbf{A}^{(t)}_{\text{s,q}} + \Delta^{(t_{\text{ref}})}_{\text{quant}} + \hat{\Delta}_{\text{quant}}^{t}.
\end{aligned}
\end{equation}

Given this, we further have:

\begin{equation}
\begin{aligned}
\mathbf{A}_{\text{full}}^{(t)} - \tilde{\mathbf{A}}_{\text{s,q}}^{(t)} &= (\mathbf{A}^{(t)}_{\text{s,q}} + \Delta^{(t_{\text{ref}})}_{\text{quant}} + \hat{\Delta}_{\text{quant}}^{(t)}) - \tilde{\mathbf{A}}_{\text{s,q}}^{(t)} \\
&= (\mathbf{A}^{(t)}_{\text{s,q}} + \Delta^{(t_{\text{ref}})}_{\text{quant}} + \hat{\Delta}_{\text{quant}}^{(t)}) - (\mathbf{A}^{(t)}_{\text{s,q}} + \Delta^{(t_{\text{ref}})}_{\text{quant}} + \hat{\Delta}_{\text{quant}}^{(t_{\text{ref}})}) \\
&= \hat{\Delta}_{\text{quant}}^{(t)} - \hat{\Delta}_{\text{quant}}^{(t_{\text{ref}})}.
\end{aligned}
\end{equation}

Similarly, for $\hat{\mathbf{A}}^{(t)}_{\text{s,q}}$, we also have:

\begin{equation}
\begin{aligned}
\mathbf{A}_{\text{full}}^{(t)} - \hat{\mathbf{A}}_{\text{s,q}}^{(t)} &= (\mathbf{A}^{(t)}_{\text{s,q}} + \Delta^{(t)}_{\text{quant}}) - \hat{\mathbf{A}}_{\text{s,q}}^{(t)} \\
&= (\mathbf{A}^{(t)}_{\text{s,q}} + \Delta^{(t)}_{\text{quant}}) - (\mathbf{A}^{(t)}_{\text{s,q}} + \Delta^{(t_{\text{ref}})}_{\text{quant}}) \\
&= \Delta_{\text{quant}}^{(t)} - \Delta_{\text{quant}}^{(t_{\text{ref}})}.
\end{aligned}
\end{equation}

Based on Proposition~\ref{prop:two_order_good}, we have:

\begin{equation}
\mathbb{E}_{t} \underbrace{ \left[ \left\| \mathbf{A}_{\text{full}}^{(t)} - \tilde{\mathbf{A}}^{(t)}_{\text{s,q}} \right\|_F \right]}_{\text{second-order}}
= \mathbb{E}_{t} \left[ \left\| \hat{\Delta}_{\text{quant}}^{(t)} - \hat{\Delta}_{\text{quant}}^{(t_{\text{ref}})} \right\|_F \right]
\leq
\mathbb{E}_{t} \left[ \left\| \Delta_{\text{quant}}^{(t)} - \Delta_{\text{quant}}^{(t_{\text{ref}})} \right\|_F \right]
=
\mathbb{E}_{t} \underbrace{\left[ \left\| \mathbf{A}_{\text{full}}^{(t)} - \hat{\mathbf{A}}^{(t)}_{\text{s,q}} \right\|_F \right]}_{\text{first-order}}.
\end{equation}

Therefore, Theorem~\ref{theorem:less_error} holds.

\begin{algorithm}[t]
    \caption{QuantSparse: Calibration to Inference Pipeline}
    \label{alg:quantsparse_pipeline}
    \begin{algorithmic}[1]
        \Require{Pre-trained video diffusion transformer $M$ (FP16), calibration dataset $\mathcal{D}_{\text{cal}}$, target bit-width (W/A), denoising steps $T$, cache interval $\tau$}
        \Ensure{Quantized-sparse model $M_{QS}$, generated video $Y$}
        
        \State \textbf{Calibration Phase:}
        \State \quad Initialize quantization parameters $\{s, z\}$ for weights (W) and activations (A)
        \State \quad Input $X \in \mathcal{D}_{cal}$ to $M$
        \State \quad Compute token saliency $s_j$ using Eq.~\ref{eq:saliency} for FP model $M$
        \State \quad Select top-$k$ salient tokens $I = \{j \mid s_j \text{ is top-}k\}$
        \State \quad \textbf{Global Guidance Distillation:}
        \State \quad \quad Calculate $\mathcal{L}_{\text{global}}$ using Eq.~\ref{eq:global_distill}
        \State \quad \textbf{Local Guidance Distillation:}
        \State \quad \quad Calculate $\mathcal{L}_{\text{local}}$ using Eq.~\ref{eq:local_distill}
        \State \quad Optimize quantization parameters using  Eq.~\ref{eq:total_loss} with $\mathcal{L}_{\text{global}}$ and $\mathcal{L}_{\text{local}}$
        \State \quad Obtain quantized model $M_{\text{quant}}$ with optimized $\{s, z\}$
        
        \State \textbf{Inference Phase:}
        \State \quad Load $M_{\text{quant}}$ and input prompt $P$.
        \State \quad Input $P$ into $M_{\text{quant}}$ and initialize cached residuals $\{\Delta_{\text{quant}}^{(t_{\text{ref}})}, \hat{\Delta}_{\text{quant}}^{(t_{\text{ref}})}\}$
        \State \quad \textbf{for} t \textbf{in} $T$
        \State \quad \quad Compute quantized sparse attention:
        \[
        A_{\text{s,q}}^{(t)} = \text{SparseAttention}(Q_{\text{quant}}, K_{\text{quant}}, V_{\text{quant}}; M)
        \]
        \State \quad \quad \textbf{if} {$t - t_{\text{ref}} \leq \tau$}
            \State \quad \quad \quad Reuse cached residuals: $\Delta_{\text{curr}} = \Delta_{\text{quant}}^{(t_{\text{ref}})} + \hat{\Delta}_{\text{quant}}^{(t_{\text{ref}})}$
        \State \quad \quad \textbf{else} 
            \State \quad \quad \quad Update $t_{\text{ref}} = t$, recompute and cache residuals
        \State \quad \quad \textbf{endif}
        \State \quad \quad Refine attention using Eq.~\ref{eq:final_second_order}
        \State \quad \textbf{endfor}
        \State \quad Generate video $Y$
        \Return $Y$
    \end{algorithmic}
\end{algorithm}

\section{Details of selected evaluation metrics}

\subsection{Multi-aspects metrics evaluation}

This evaluation suite includes absolute quality of videos and relative difference metrics that quantify the difference between FP16 generation. 

\textbf{Absolute Quality.} Consistent with prior quantization works~\citep{zhao2024vidit, feng2025qvdit}, we adopt CLIPSIM, VQA, and FlowScore to measure text-video alignment, quality, and temporal consistency, respectively.

\textbf{Relative Difference Metrics.} Following prior sparse attention works~\citep{xi2025svg, yuan2024ditfastattn, ren2025grouping, zhang2025jenga}, we adopt Peak Signal-to-Noise Ratio (PSNR), Structural Similarity Index Measure (SSIM), and Learned Perceptual Image Patch Similarity (LPIPS) for pixel-space differences, structural similarity, and high-level patch similarity, respectively.

All the evaluations are conducted on high-resolution generation tasks. Due to the computational overhead, we use the OpenSORA prompt sets used in~\citep{zhao2024vidit, feng2025qvdit} for video generation.

\subsection{Benchmark evaluation}

To further provide benchmark evaluation, we follow previous works~\citep{feng2025qvdit, zhao2024vidit}. We select 8 major dimensions from Vbench~\citep{huang2024vbench}, including frame-wise quality, temporal quality, and semantic evaluation. 

For \textbf{Frame-wise Quality}, we select \textit{Imaging Quality} and \textit{Aesthetic Quality} for distortion assessment and artistic and beauty evaluation. For \textbf{Temporal Quality}, we use \textit{Dynamic Degree}, \textit{Motion Smoothness}, \textit{Subject Consistency}, and \textit{Background Consistency} for degree of dynamics, physical law smoothness, subject’s appearance consistent, and temporal consistency of the background, respectively. For \textbf{Semantic Evaluation}, we use \textit{Scene} and \textit{Overall Consistency} for text prompt scene consistency and overall video-text consistency.

The evaluation follows the suite provided by VBench~\citep{huang2024vbench}. We generate one video for each prompt, same as previous works~\citep{zhao2024vidit, feng2025qvdit}. Due to the large prompt sets used in VBench, we slightly decrease the resolution for computational efficiency. In addition, this experimental setup also provides an additional evaluation of multi-resolution video generation performance, which proves the generalization and effectiveness of our method in different application scenarios.

\begin{table}[t!]
\caption{Text-to-Video generation experiments on more huge models.}
\label{tab:sora_small}
\begin{center}
\resizebox{0.9\linewidth}{!}{
\input{tables/sora_hy}}
\end{center}
\end{table}

\begin{table}[t!]
\caption{Performance of text-to-video generation under VBench evaluation benchmark suite. We evaluate on Imaging Quality (IQ), Aesthetic Quality (AQ), Motion Smoothness (MS), Dynamic Degree (DD), Background Consistency (BC), Subject Consistency (SuC), Scene Consistency (ScC), and Overall Consistency (OC). Higher ($\mathbf{\red{\uparrow}}$) metrics represent better performance. \textbf{Bold}: the best result. \underline{Underline}: The second best result.}
\label{tab:vbench}
\begin{center}
\resizebox{0.9\linewidth}{!}{
\input{tables/vbench}}
\end{center}
\end{table}

\section{Experiment Settings}
\label{sec:more_expe_detail}

Same with prior works~\citep{zhao2024vidit, ashkboos2024quarot, feng2025qvdit}, we adopt channel-wise weight quantization and dynamic token-wise activation quantization. And we use uniform symmetry quantization for both weight and activation for better hardware acceleration and memory saving. For fair comparison, we apply the same quantization granularity for all quantization methods. We adopt channel-wise scale used in ~\citep{xiao2023smoothquant, wu2024ptq4dit, zhao2024vidit, feng2025qvdit} and rotation-based matrix used in~\citep{ashkboos2024quarot, zhao2024vidit, sun2024flatquant} for quantization. We follow block-wise post-training strategy used in~\citep{wu2024ptq4dit, chen2024qdit, sun2024flatquant} for calibration. All the experiments are conducted on a single NVIDIA A800 GPU.

During calibration, we set channel-wise scale, rotation matrix, and quantization scale as learnable following~\citep{feng2025qvdit, sun2024flatquant}. We use 20 random generated samples and train 15 epoch for each transformer block. We apply the same calibration samples and epochs for all methods for fair comparison. We use AdamW~\citep{loshchilov2017adamw} optimizer and cosine learning rate scheduler. For the channel-wise scale and rotation matrix, we use a learning rate of $5e^{-3}$. For the learnable quantization scale, we use a learning rate of $5e^{-2}$. For distillation, we use $r=128$ for global distillation pooling, $k=256$ for salient query selection, and $\lambda_{\text{global}}=1e^{-4}, \lambda_{\text{global}}=1e^{-4}$ for Wan2.1-1.3B, Wan2.1-14B, and $\lambda_{\text{global}}=1.0, \lambda_{\text{global}}=1e^{2}$ for HunyuanVideo, respectively. The selection of distillation balancing factor is based on the order of magnitude of the loss. For sparse attention, we use a fixed cache refreshing interval of $5$, and use $k=16$ for SVD.

For deployment, we quantize the weight and absorb all the quantization parameters following~\citep{zhao2024vidit, sun2024flatquant, feng2025qvdit, ashkboos2024quarot}. For activation, we use dynamic online quantization same as~\citep{feng2025qvdit, sun2024flatquant, zhao2024vidit}.

\section{More evaluation results on Wan2.1-1.3B}
\label{sec:more_eval}

We present comprehensive evaluation on Wan2.1-1.3B~\citep{wan2025wan} in Tab.~\ref{tab:sora_small}. Since Wan2.1-1.3B has less computation budget and we find that it will suffer from serious performance degradation under high sparsity, we uniformly adopt 40 density in sparse attention to ensure its performance. 

Different quantization methods have obvious performance degradation, especially under W4A8. Among them, the quantization method specially designed for video model Q-VDiT~\citep{feng2025qvdit} and the strong LLM quantization method Quarot~\citep{ashkboos2024quarot} show relatively stronger performance. For a broader and fair comparison, we add existing sparse attention methods to Q-VDiT and Quarot to verify the effect of naive combination of model quantization and sparse attention. We find that when combining Q-VDiT and Quarot with different sparse attention methods, the performance decreases to varying degrees, and the performance of SVG~\citep{xi2025svg} decreases the least. Therefore, we chose SVG as our baseline sparse attention in all other experiments.

Compared with all existing methods, QuantSparse achieves SOTA performance under all bit settings, and is almost lossless compared with the FP model. It is worth mentioning that QuantSparse even surpasses all quantization-only methods. It not only achieves better compression effect, but also has better performance, which fully demonstrates the effectiveness of our method.

\section{VBench Evaluation results}
\label{sec:vbench}

We present the VBench~\citep{huang2024vbench} evaluation results in Tab.~\ref{tab:vbench}. Under the comprehensive evaluation of all 8 dimensions, the naive combination of Q-VDiT~\citep{feng2025qvdit}, Quarot~\citep{ashkboos2024quarot} and SVG~\citep{xi2025svg} all show significant performance degradation, which fully demonstrates the disadvantage of simply combining existing quantization and sparse attention methods. While QuantSparse achieves comprehensive SOTA performance in all bit settings of all models, and is almost lossless compared with FP model, even better in some dimensions. For Wan2.1-14B~\citep{wan2025wan} under W4A8, QuantSparse achieves 63.55 and 63.81 under 25\% and 15\% attention density, respectively, surpassing 63.38 of FP model.

\begin{figure}[h]
    \centering
    \subfloat[][Block14 head1.]{
        \includegraphics[width=0.18\linewidth]{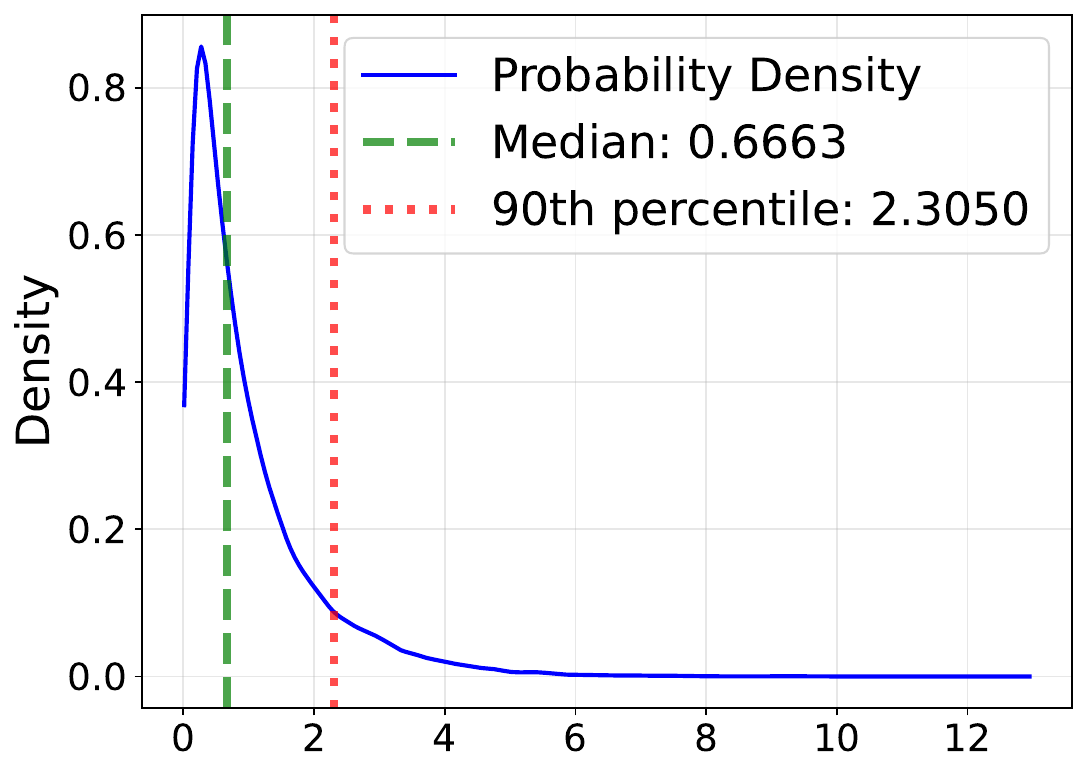}
    }
    \subfloat[][Block15 head1.]{
        \includegraphics[width=0.18\linewidth]{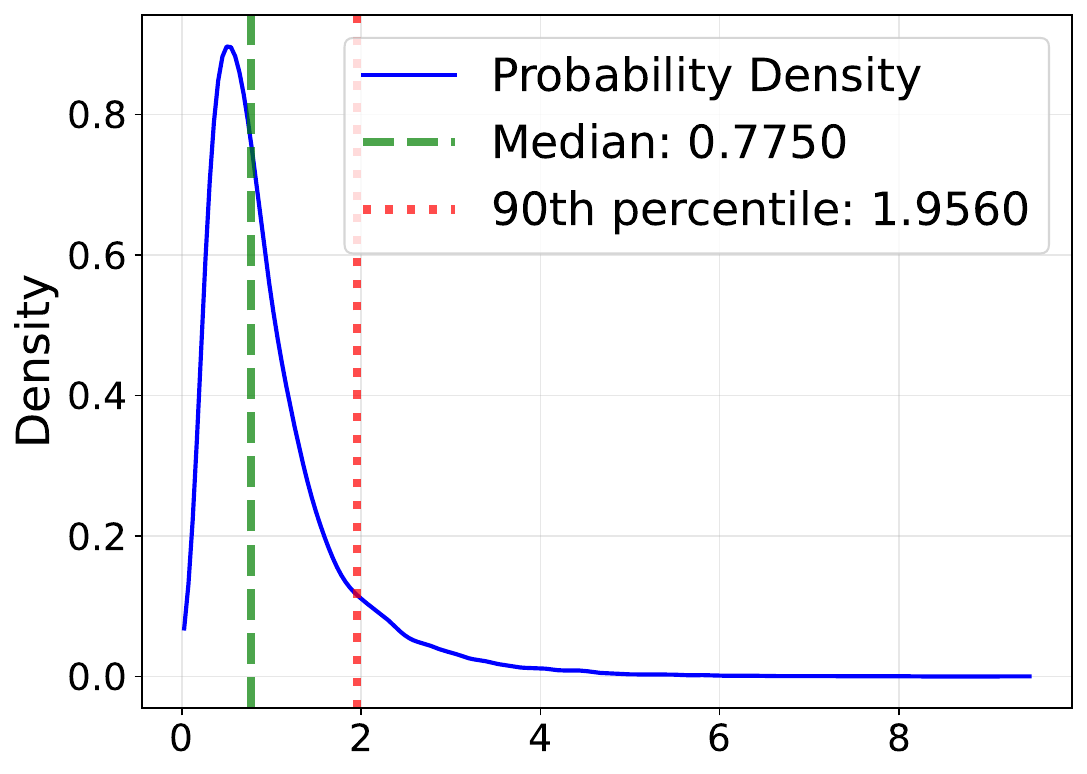}
    }
    \subfloat[][Block16 head1.]{
        \includegraphics[width=0.18\linewidth]{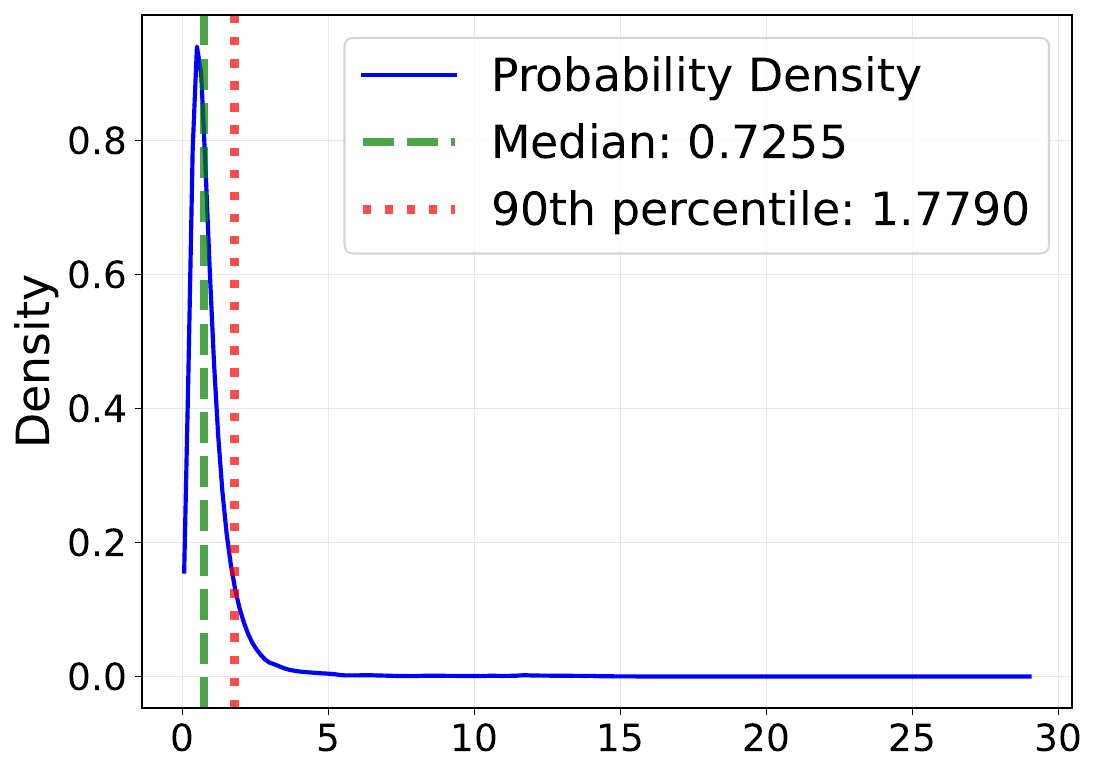}
    }
    \subfloat[][Block17 head1.]{
        \includegraphics[width=0.18\linewidth]{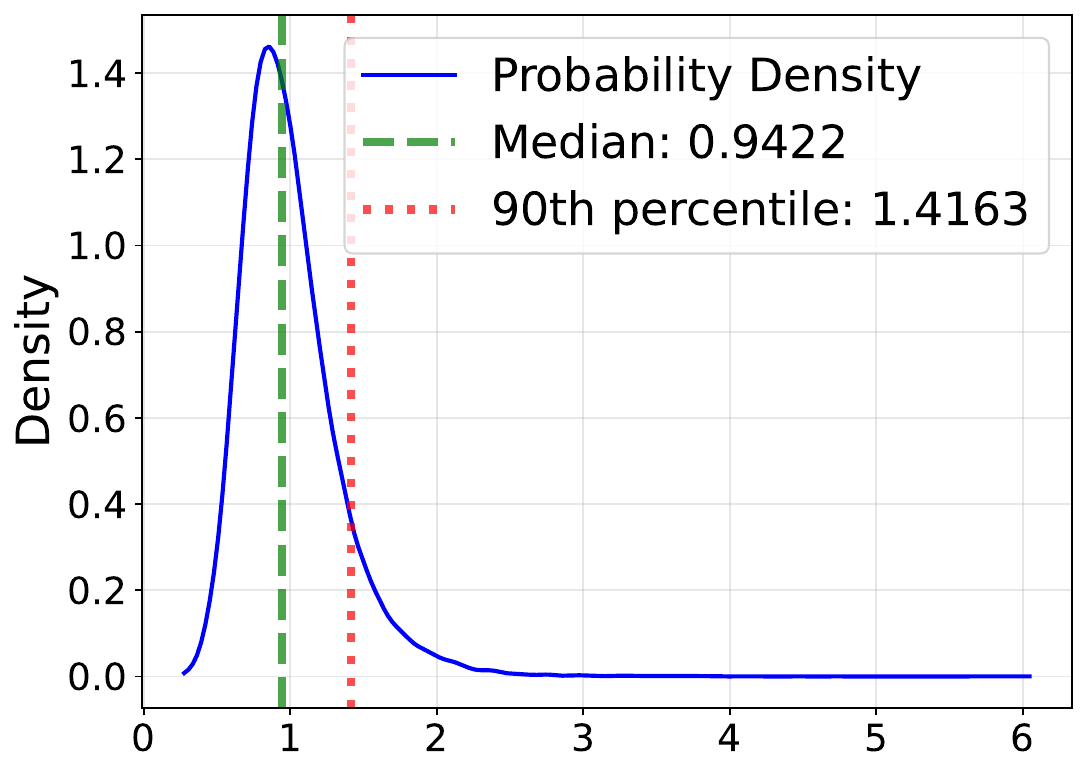}
    }
    \subfloat[][Block18 head1.]{
        \includegraphics[width=0.18\linewidth]{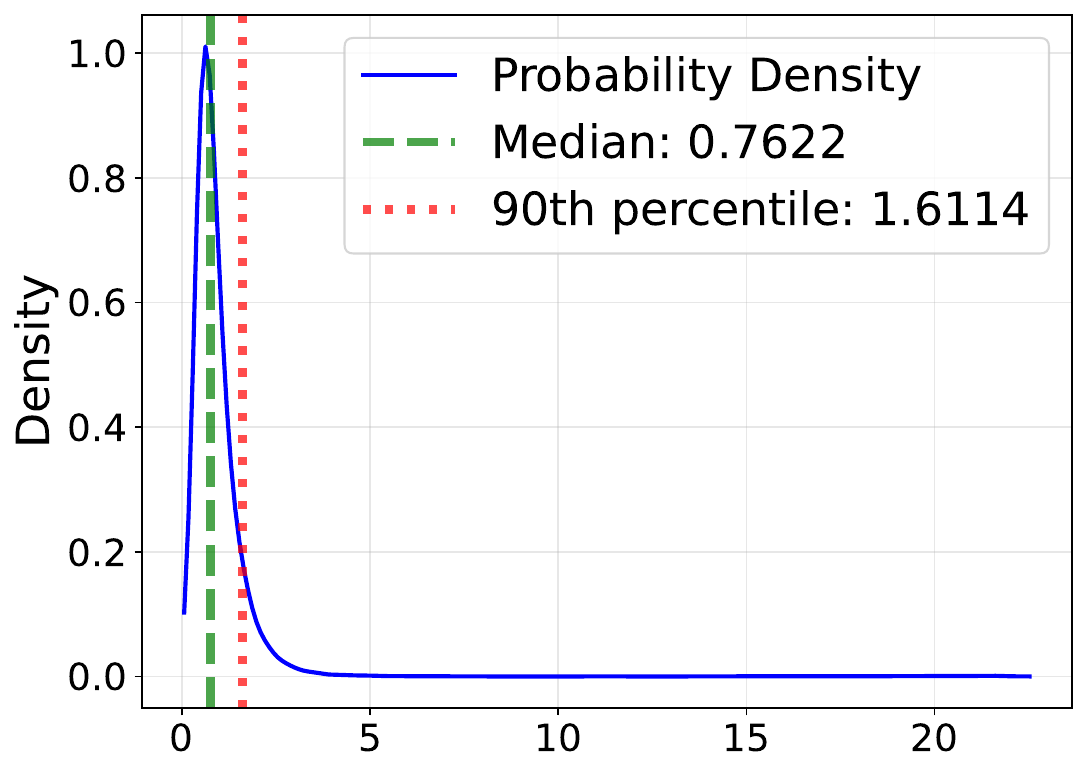}
    }
    \caption{More token saliency distribution of Wan2.1-1.3B~\citep{wan2025wan}.}
\label{fig:more_token_saliency_wan}
\end{figure}

\begin{figure}[h]
    \centering
    \subfloat[][\textit{Block14 head1}.]{
        \includegraphics[width=0.18\linewidth]{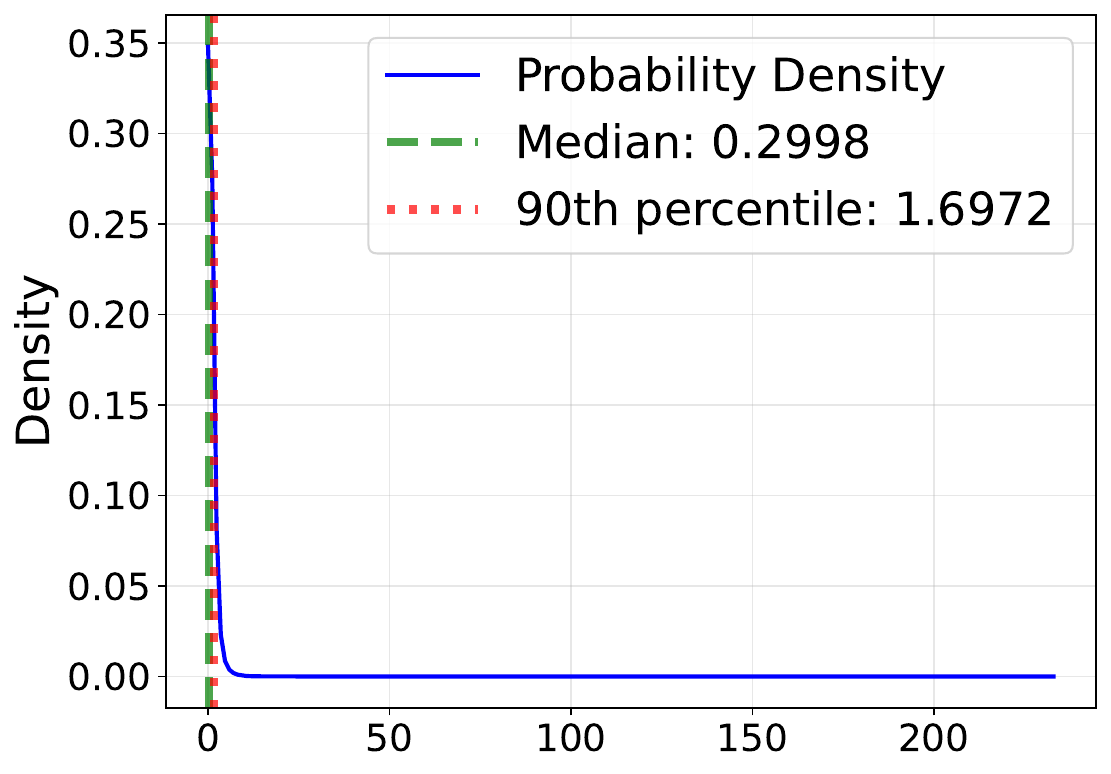}
    }
    \subfloat[][\textit{Block15 head1}.]{
        \includegraphics[width=0.18\linewidth]{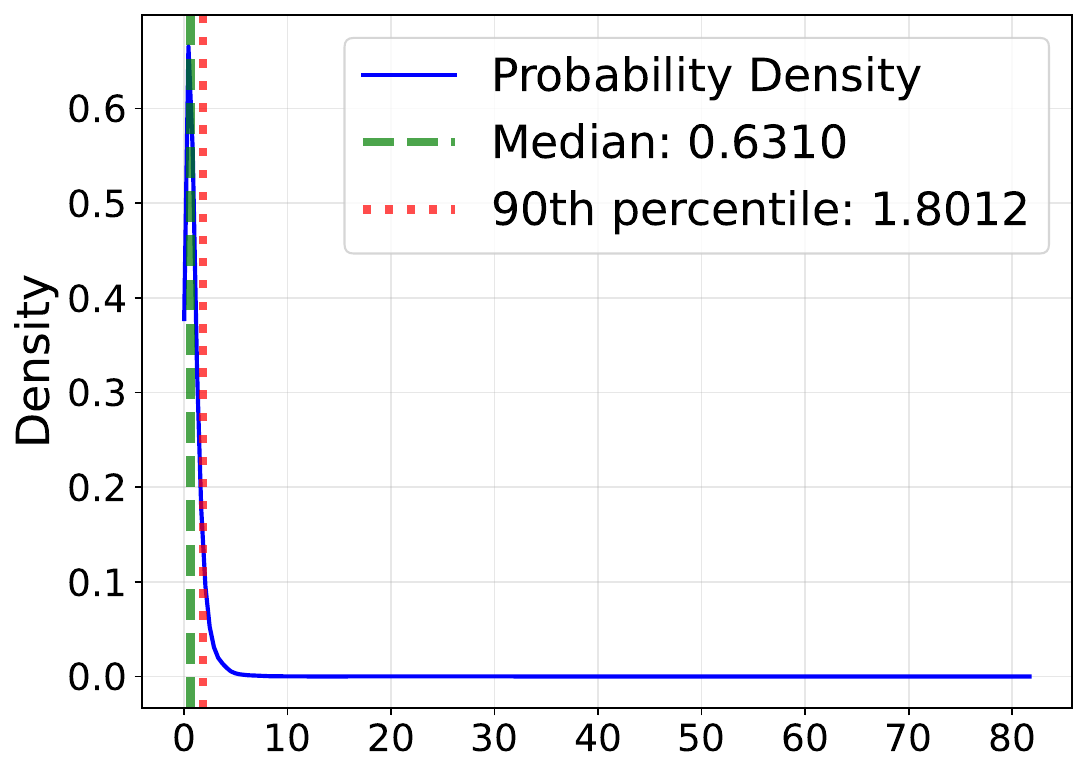}
    }
    \subfloat[][\textit{Block16 head1}.]{
        \includegraphics[width=0.18\linewidth]{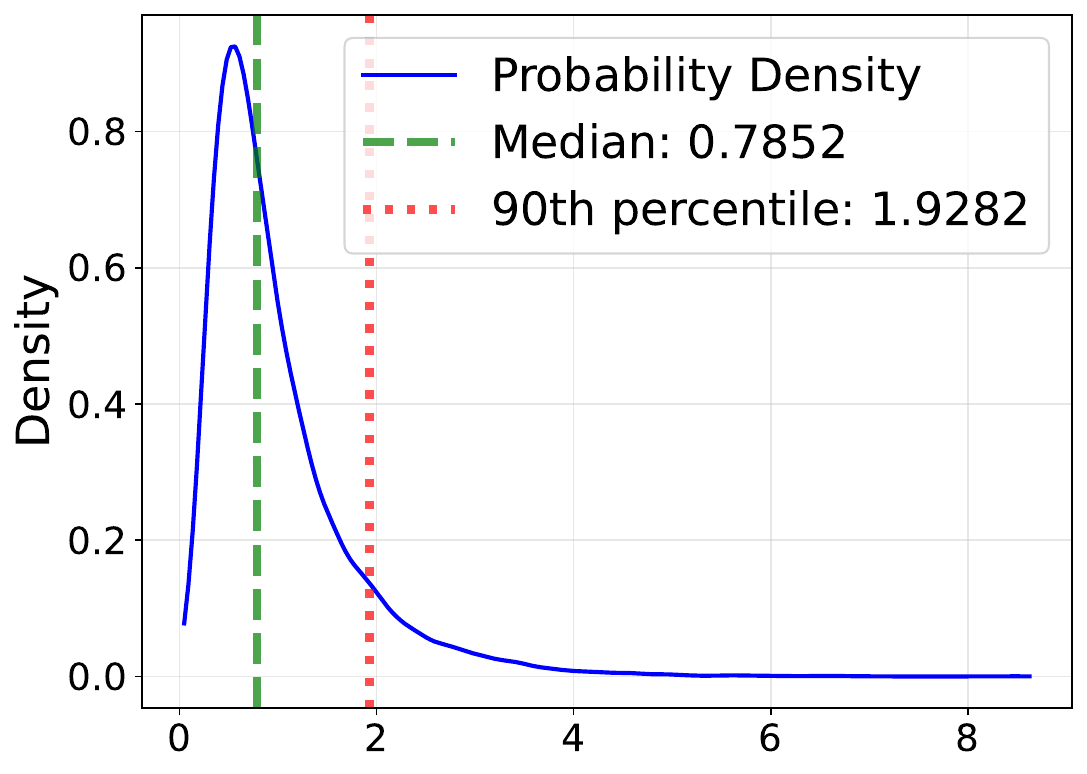}
    }
    \subfloat[][\textit{Block17 head1}.]{
        \includegraphics[width=0.18\linewidth]{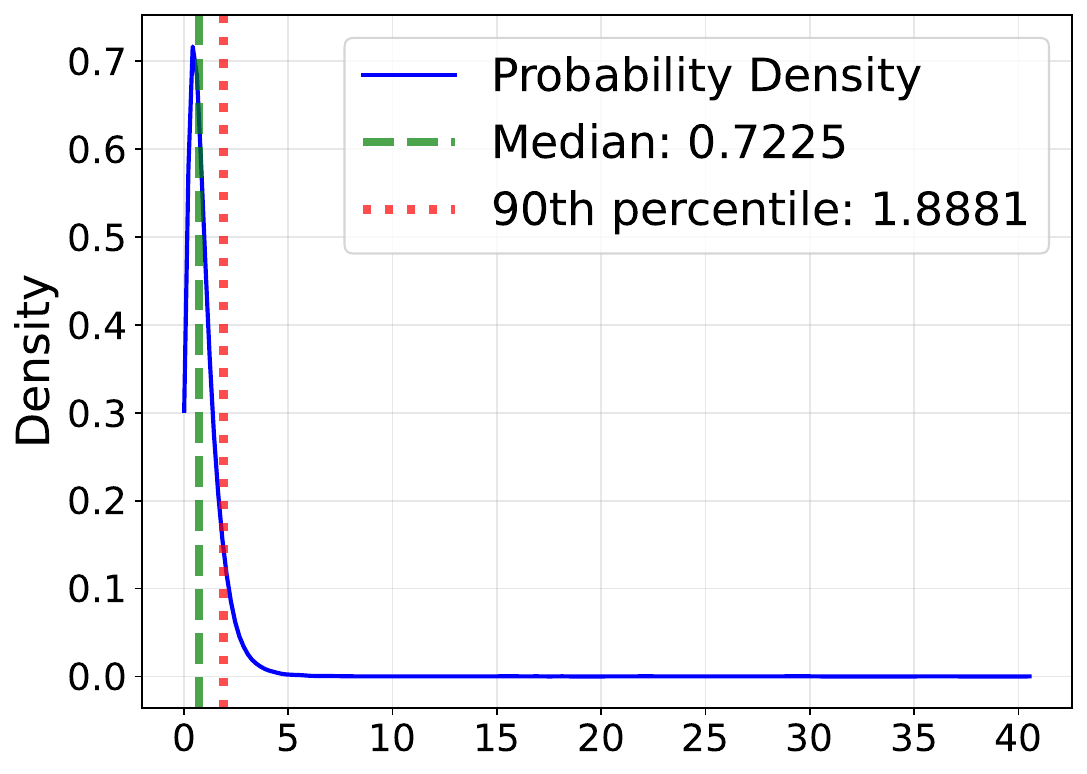}
    }
    \subfloat[][\textit{Block18 head1}.]{
        \includegraphics[width=0.18\linewidth]{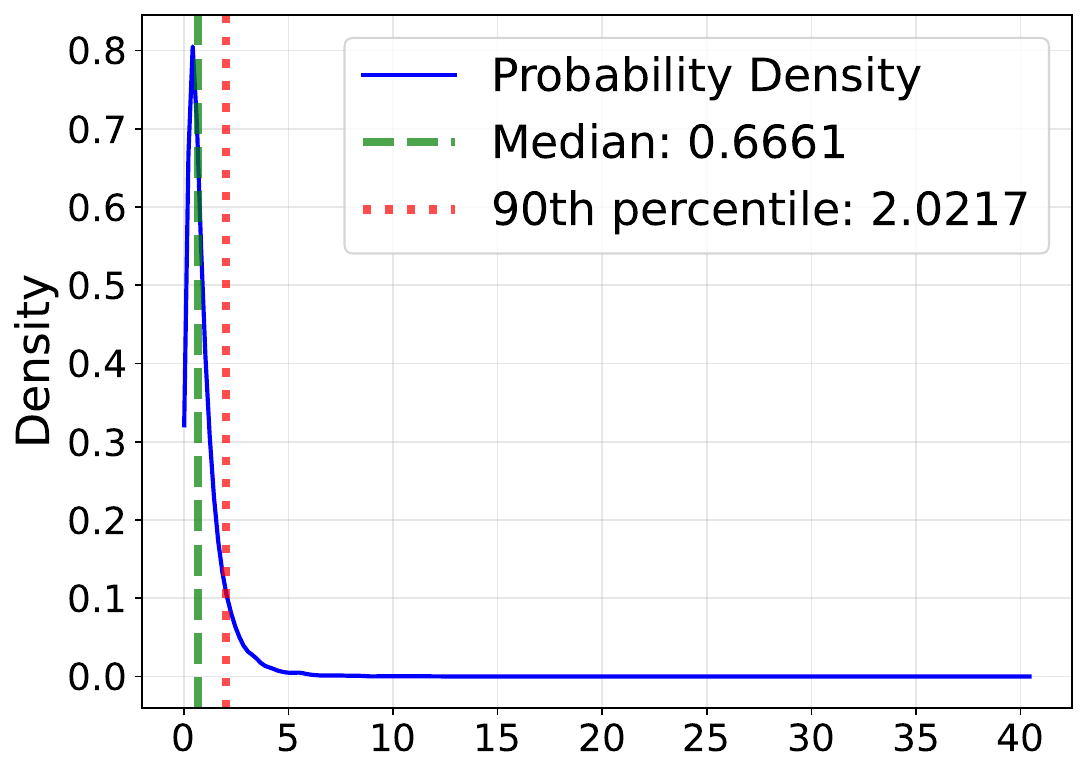}
    }
    \caption{More token saliency distribution of HunyuanVideo-13B~\citep{kong2024hunyuanvideo}.}
\label{fig:more_token_saliency_hy}
\end{figure}

\section{More analysis of Multi-Scale Salient Attention Distillation}
\label{sec:more_mssa}

We present more analysis of the proposed Multi-Scale Salient Attention Distillation (MSAD) here. 

\rebuttal{
We conducted quantitative experiments to test the impact of quantization and sparsification on attention shift by measuring the attention Mean Square Error (MSE). The results are collected from 1000 random samples on Wan2.1-1.3B~\citep{wan2025wan} under W4A8 and 40\% attention density. The results are presented in Tab.~\ref{tab:attention_shift}. The attention shift caused by the simple combination of quantization and sparsification methods is much greater than the sum of individual shifts. This proves the joint effect of quantization and sparsification on attention error, and our core motivation "amplified attention shift".
}

\begin{table}[h!]
\caption{\rebuttal{Quantitative experiment on attention shift caused by different compression techniques.}}
\label{tab:attention_shift}
\begin{center}
\resizebox{0.6\linewidth}{!}{
\begin{tabular}{l|c}
\toprule
\textbf{Method} & Attention Shift \\
\midrule  

Quantization (QuaRot~\citep{ashkboos2024quarot}) & 0.216 \\
Sparsification (SVG~\citep{xi2025svg}) & 0.134 \\
\textbf{Quantization+Sparsification} & \textbf{0.685} \\

\bottomrule
\end{tabular}}
\end{center}
\end{table}

\begin{figure}[h!]
  \centering
    \includegraphics[width=0.95\linewidth]{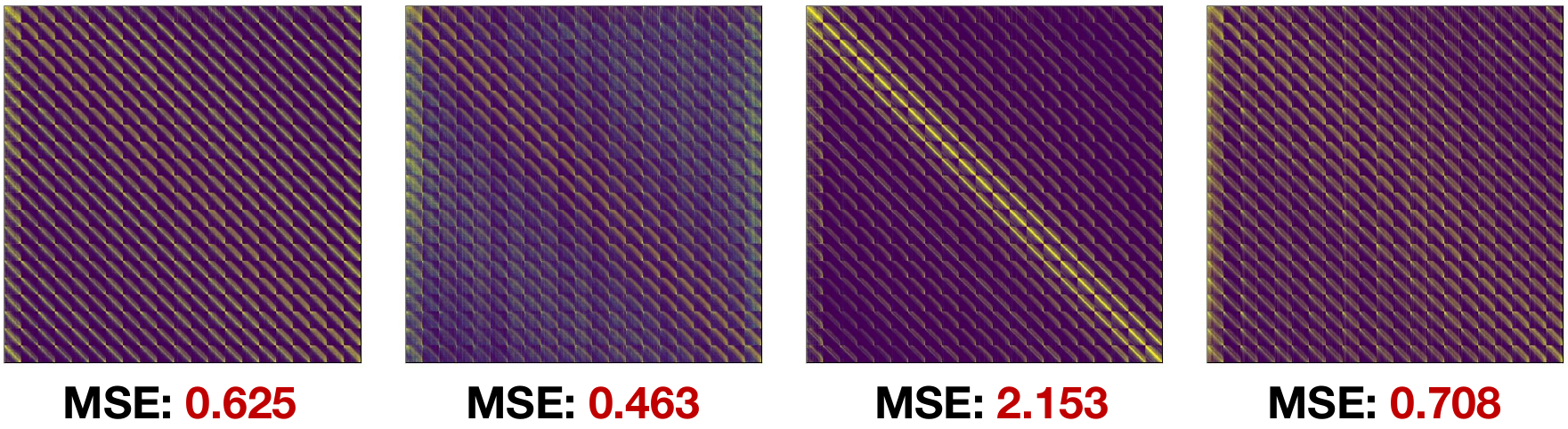}
  \caption{\rebuttal{Attention differences between FP model and quantized model without distillation.}}
  \label{fig:more_attn_map_wo}
\end{figure}

\begin{figure}[h!]
  \centering
    \includegraphics[width=0.95\linewidth]{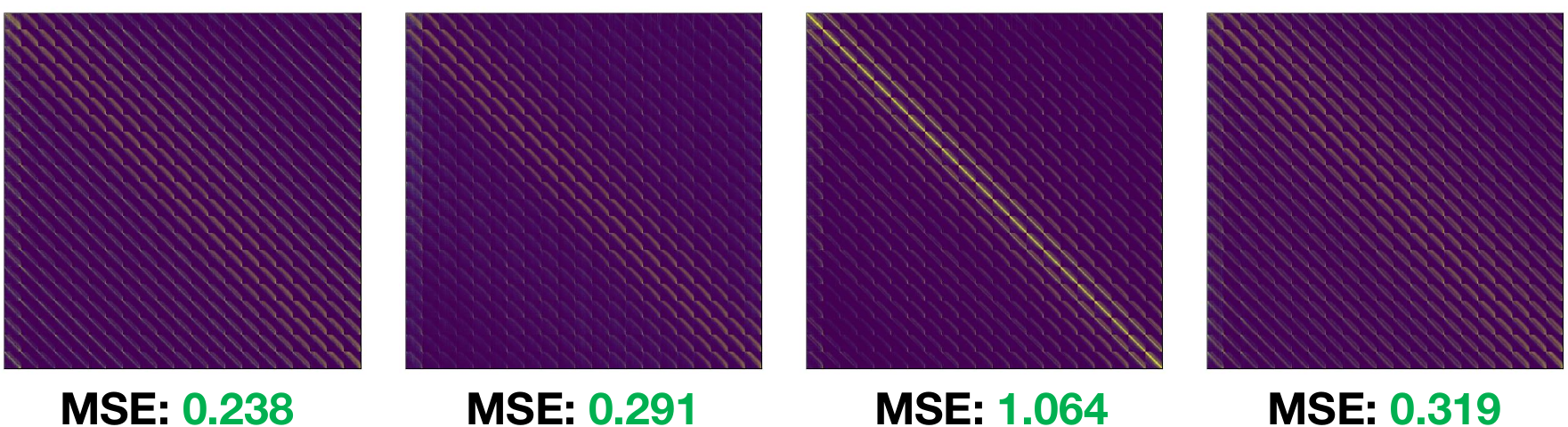}
  \caption{\rebuttal{Attention differences between FP model and quantized model with distillation.}}
  \label{fig:more_attn_map_w}
\end{figure}

\rebuttal{
We supplement 4 additional attention map comparisons in Fig.~\ref{fig:more_attn_map_wo} and Fig.~\ref{fig:more_attn_map_w}, showing the attention distribution difference between the FP model and quantized model. The results are collected from Wan2.1-1.3B under W4A8.
}

\rebuttal{
Each column in Fig.~\ref{fig:more_attn_map_wo} and Fig.~\ref{fig:more_attn_map_w} corresponds to the attention difference between the same attention map before and after the proposed distillation MSAD. This indicates that our MSAD effectively alleviates the attention shift.
}

We present more visualization of heavy-tail token saliency distribution in Fig.~\ref{fig:more_token_saliency_wan} and Fig.~\ref{fig:more_token_saliency_hy}. It can be seen that a significantly heavy-tailed token saliency phenomenon appears in different blocks of Wan2.1~\citep{wan2025wan} and HunyuanVideo~\citep{kong2024hunyuanvideo}, which fully shows that our salient local distillation is meaningful.

\begin{wraptable}{r}{0.45\linewidth}
    \vspace{-0.2in}
  \centering
  \begin{minipage}[b]{\linewidth}
    \caption{Ablation results of local distillation.}
      \resizebox{\linewidth}{!}{
      \input{tables/abla_local}}
        \label{tab:ablation_local_distill}
    \end{minipage}
     \vspace{-0.4in}
\end{wraptable}

To further prove the effect of top-$k$ salient queries selection, we compare with random selection methods and present the results in Tab.~\ref{tab:ablation_local_distill}. Compared with random selection, our top-$k$ salient selection further improves the PSNR from 15.49 to 16.82, fully demonstrating the effectiveness of our local distillation.

\begin{figure}[h]
    \centering
    \subfloat[][\textit{block.11}.]{
        \includegraphics[width=0.18\linewidth]{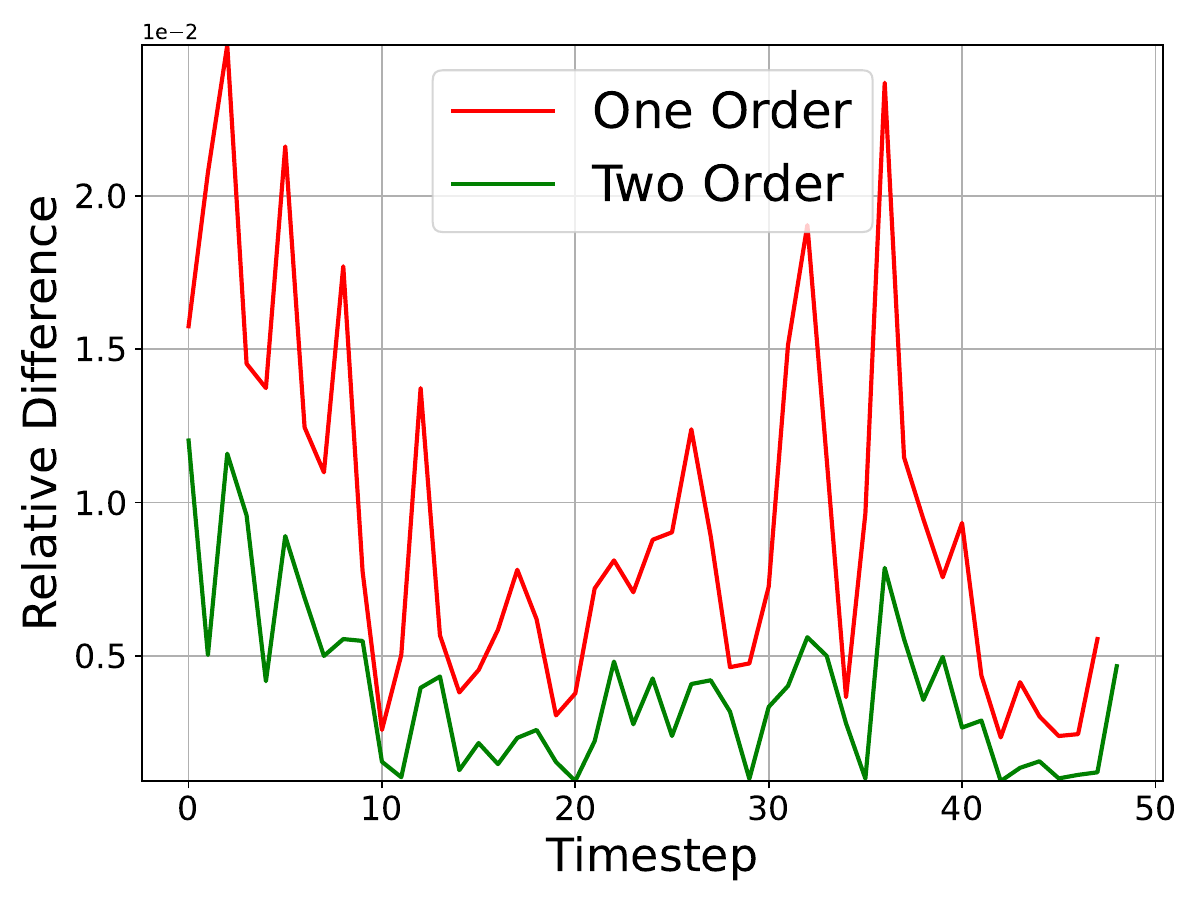}
    }
    \subfloat[][\textit{block.12}.]{
        \includegraphics[width=0.18\linewidth]{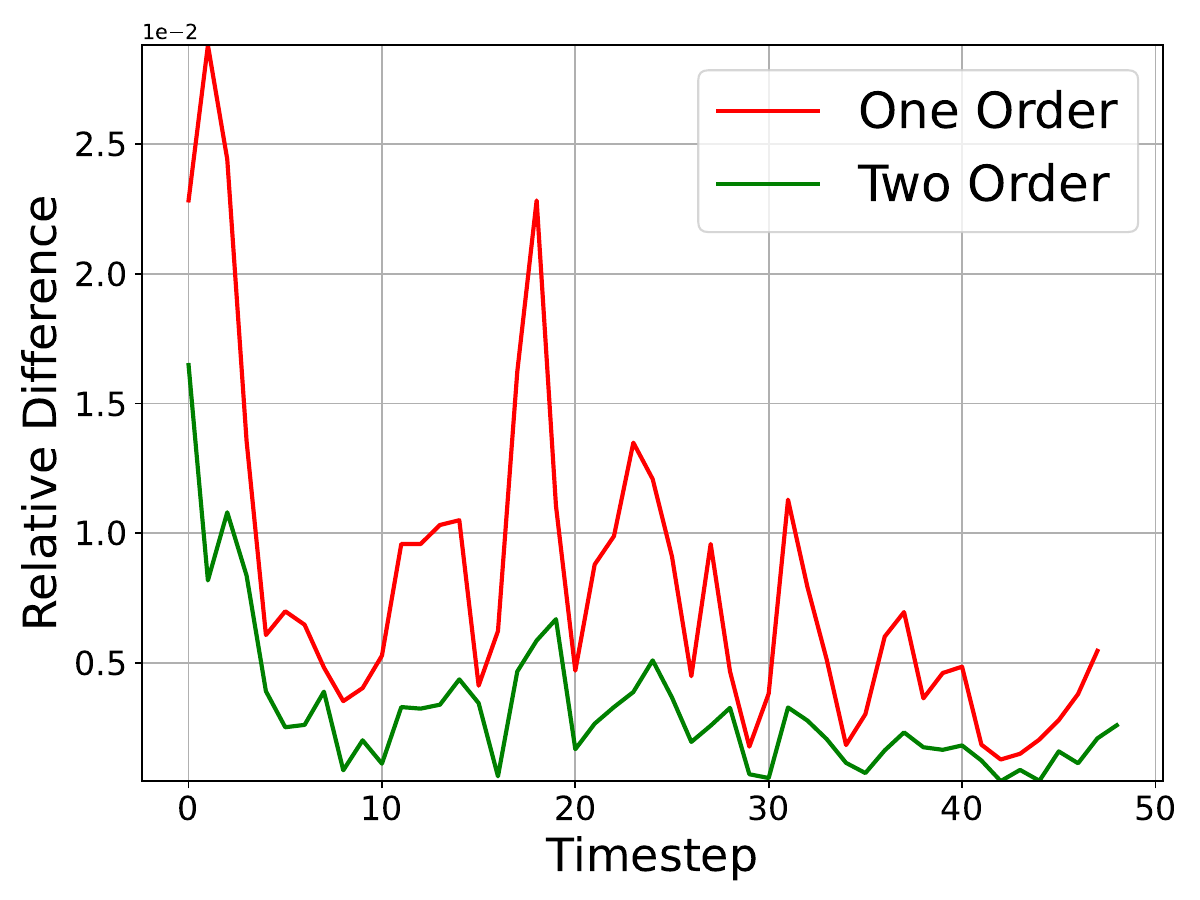}
    }
    \subfloat[][\textit{block.13}.]{
        \includegraphics[width=0.18\linewidth]{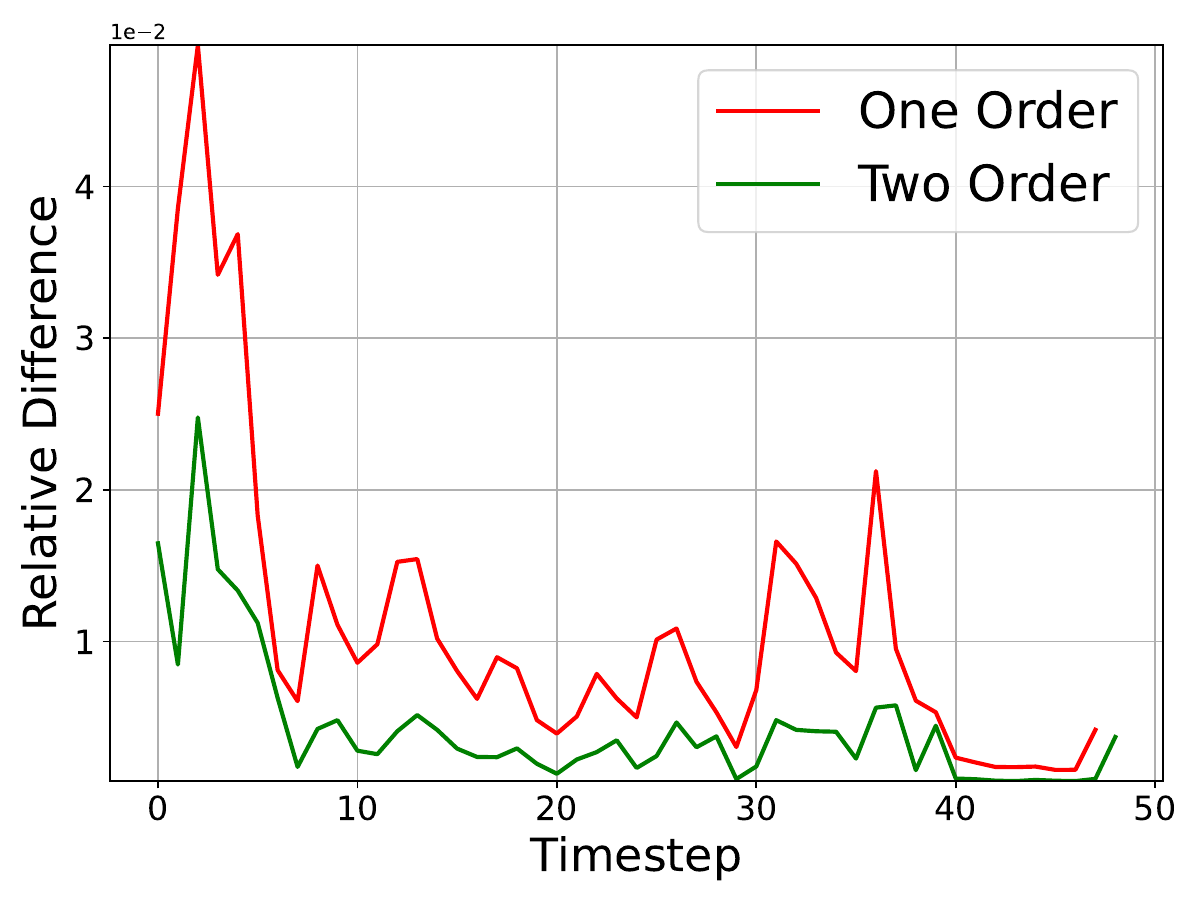}
    }
    \subfloat[][\textit{block.14}.]{
        \includegraphics[width=0.18\linewidth]{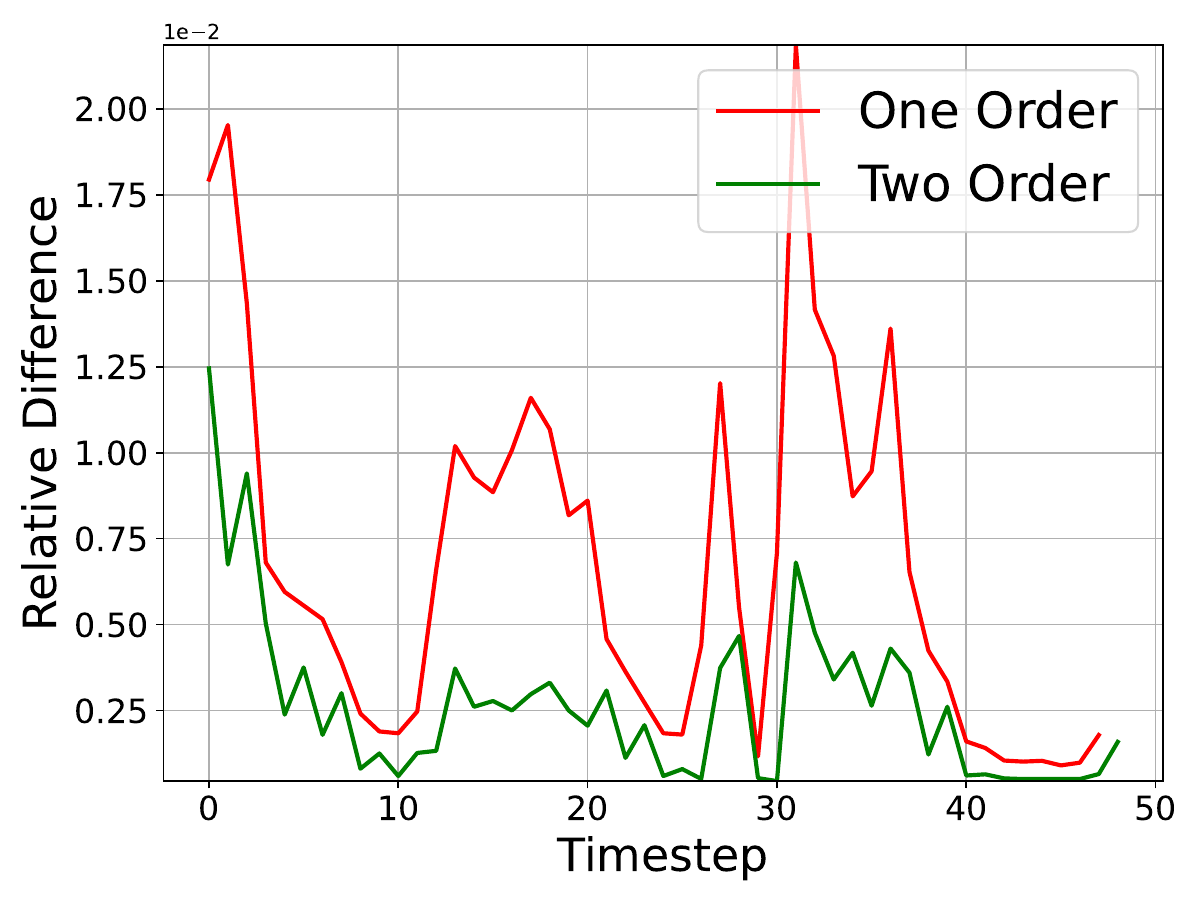}
    }
    \subfloat[][\textit{block.15}.]{
        \includegraphics[width=0.18\linewidth]{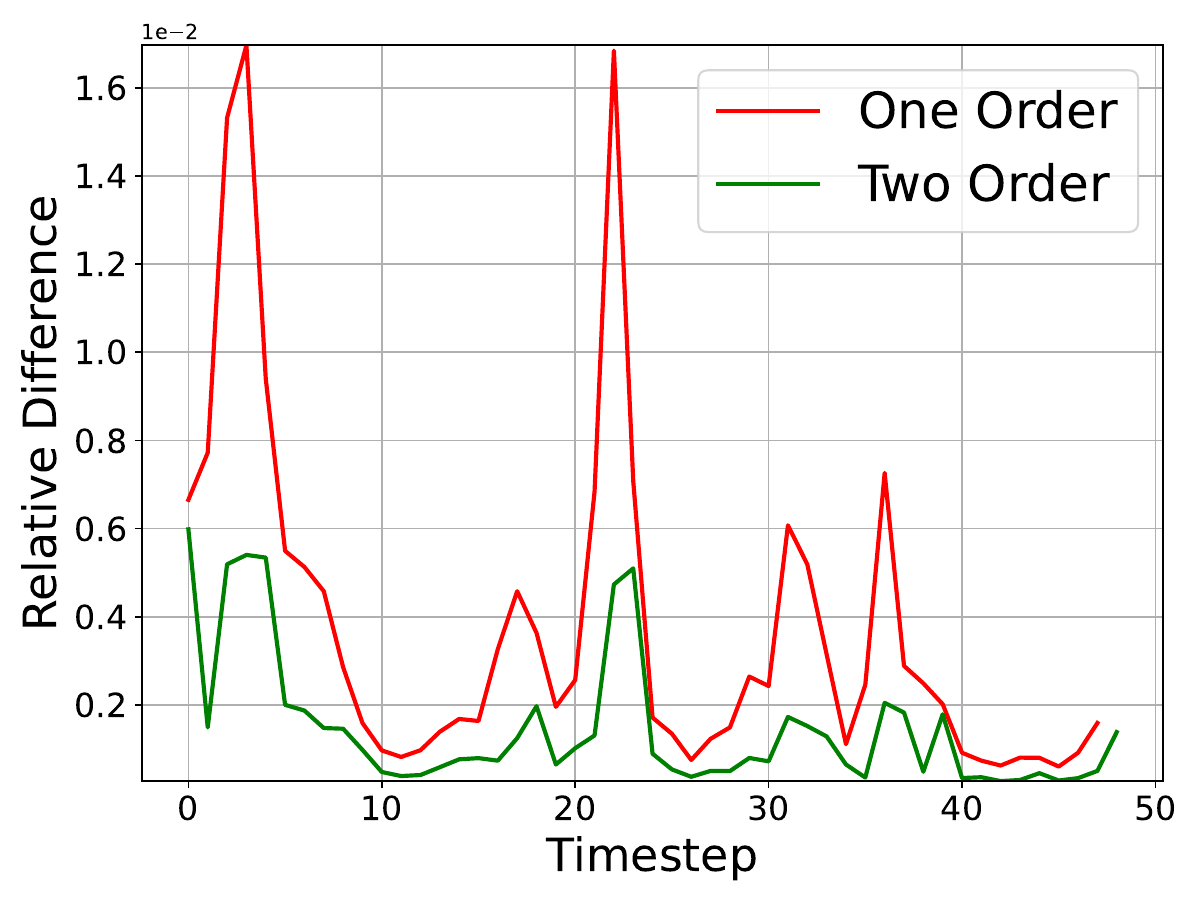}
    }
    \caption{More residual temporal difference distribution of HunyuanVideo-13B~\citep{kong2024hunyuanvideo}.}
\label{fig:more_temporal_vari}
\end{figure}

\section{More analysis of Second-Order Sparse Attention Reparameterization}
\label{sec:more_ssar}

We present more analysis of the proposed Second-Order Sparse Attention Reparameterization (SSAR) here. We present more visualization of residual temporal difference in Fig.~\ref{fig:more_temporal_vari}. It can be seen that after the introduction of quantization, the numerical difference of the first-order residuals of adjacent time steps cannot be simply ignored. However, the numerical difference of the second-order residual is significantly smaller than that of the first-order residual, so the use of the second-order residual has a better approximation effect.

\begin{figure}[h]
    \centering
    \subfloat[][\textit{block.11}.]{
        \includegraphics[width=0.18\linewidth]{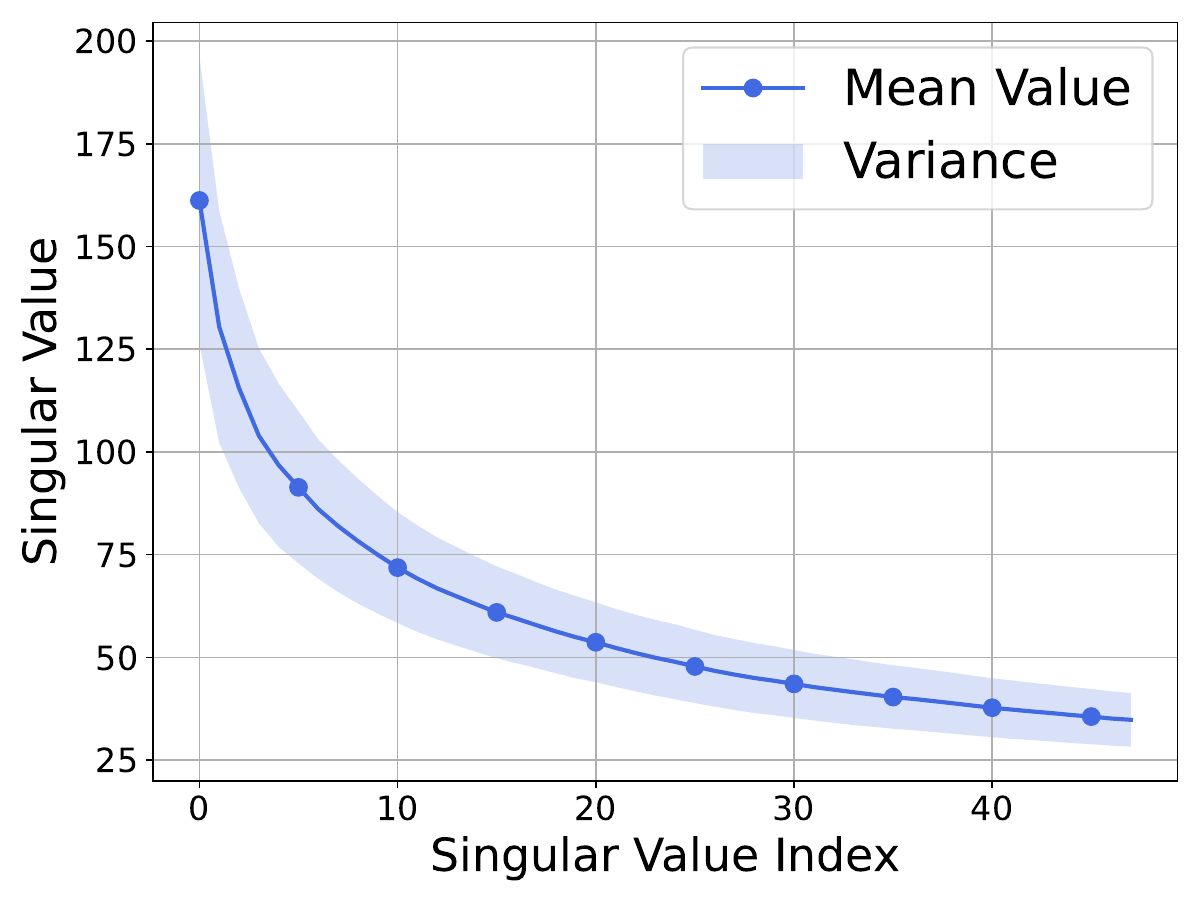}
    }
    \subfloat[][\textit{block.12}.]{
        \includegraphics[width=0.18\linewidth]{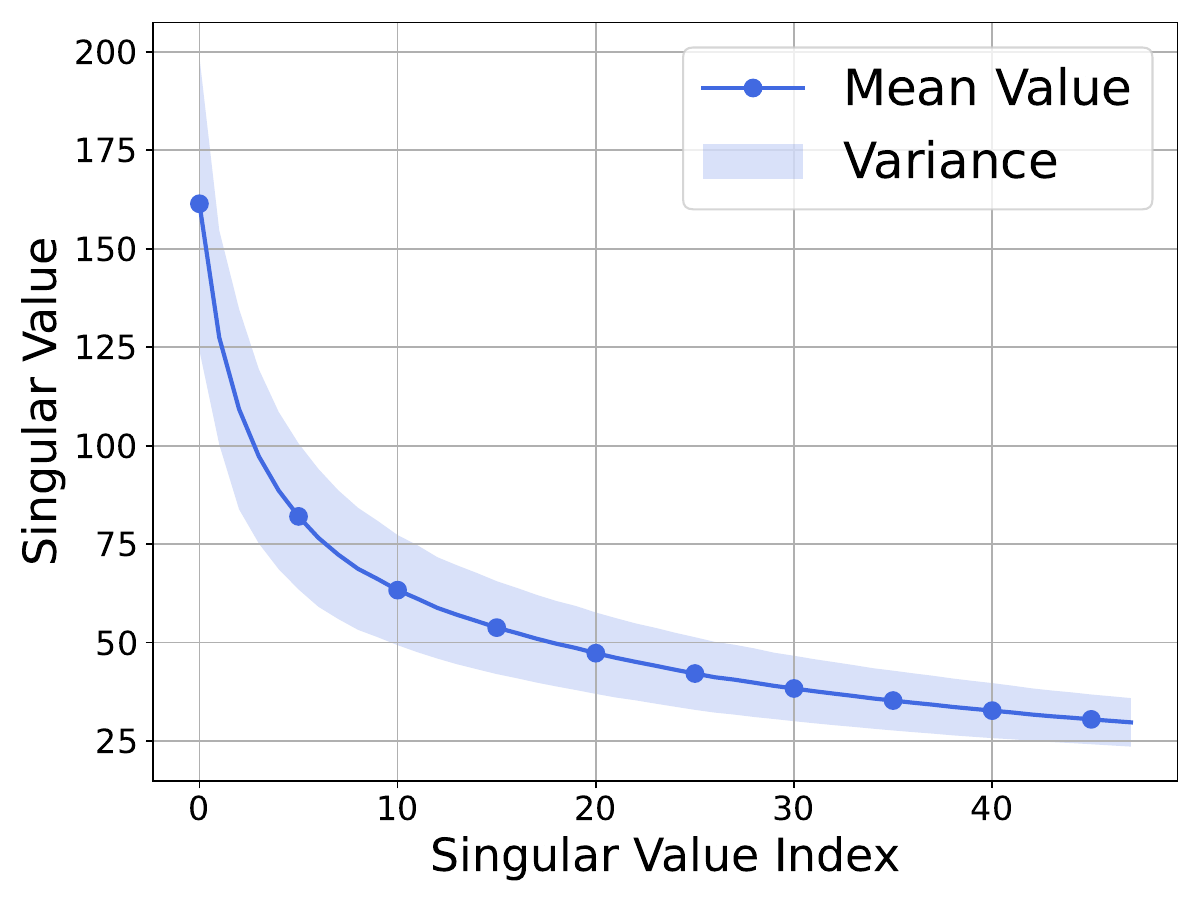}
    }
    \subfloat[][\textit{block.13}.]{
        \includegraphics[width=0.18\linewidth]{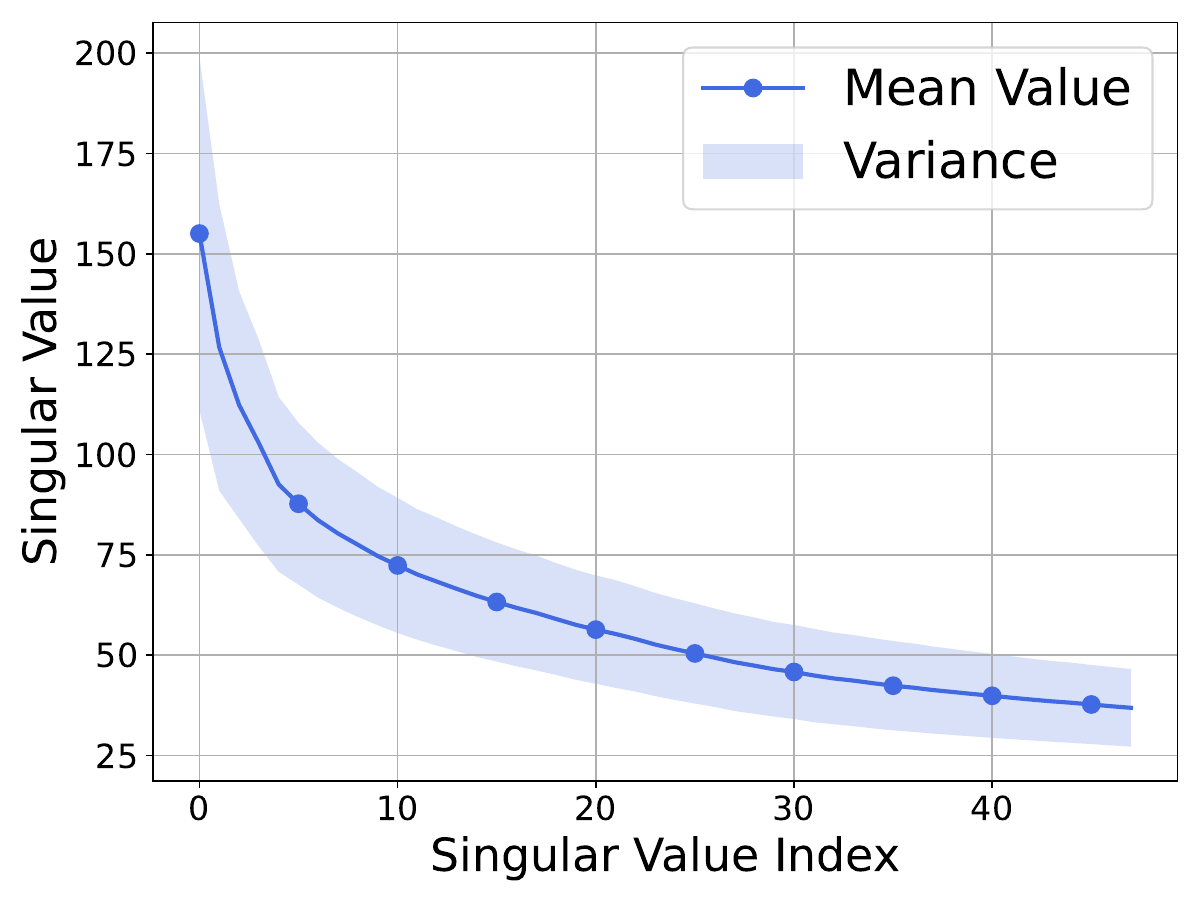}
    }
    \subfloat[][\textit{block.14}.]{
        \includegraphics[width=0.18\linewidth]{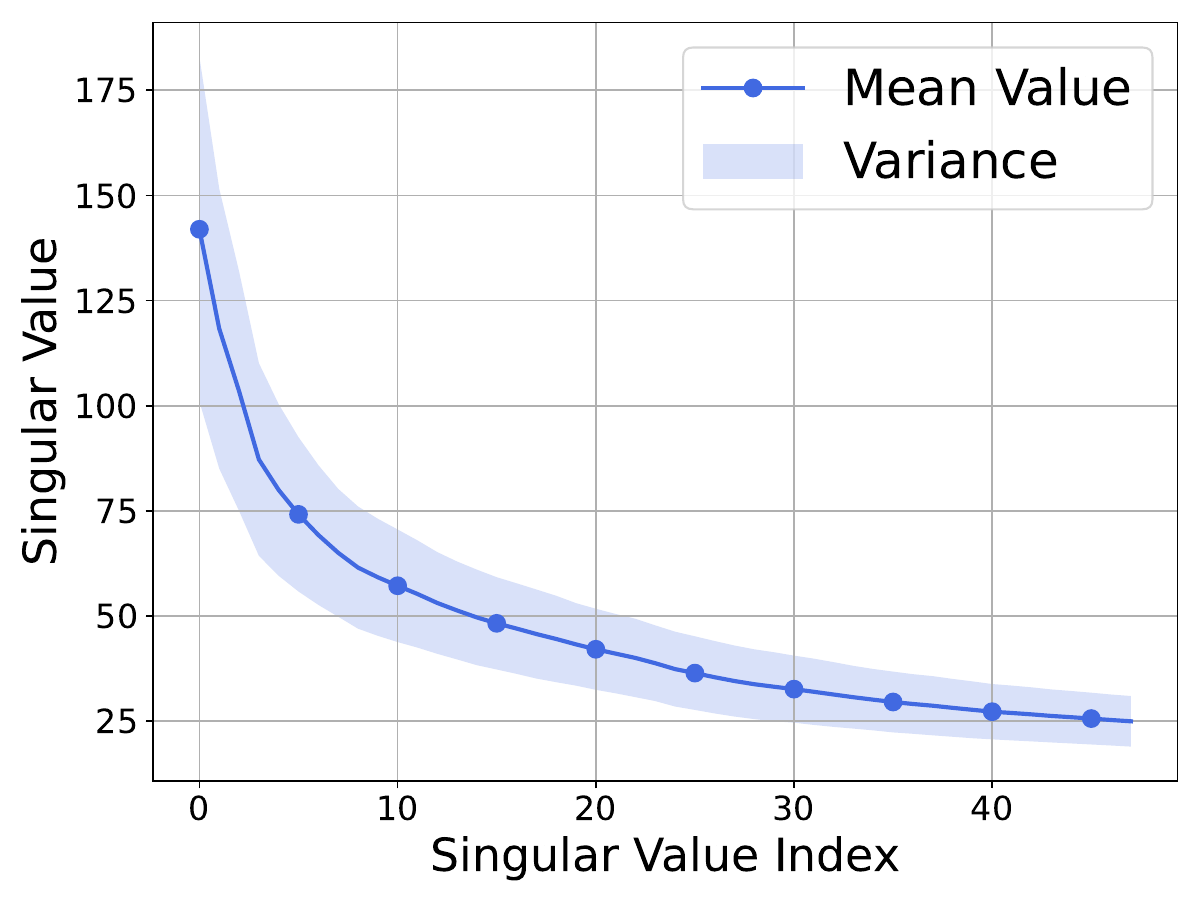}
    }
    \subfloat[][\textit{block.15}.]{
        \includegraphics[width=0.18\linewidth]{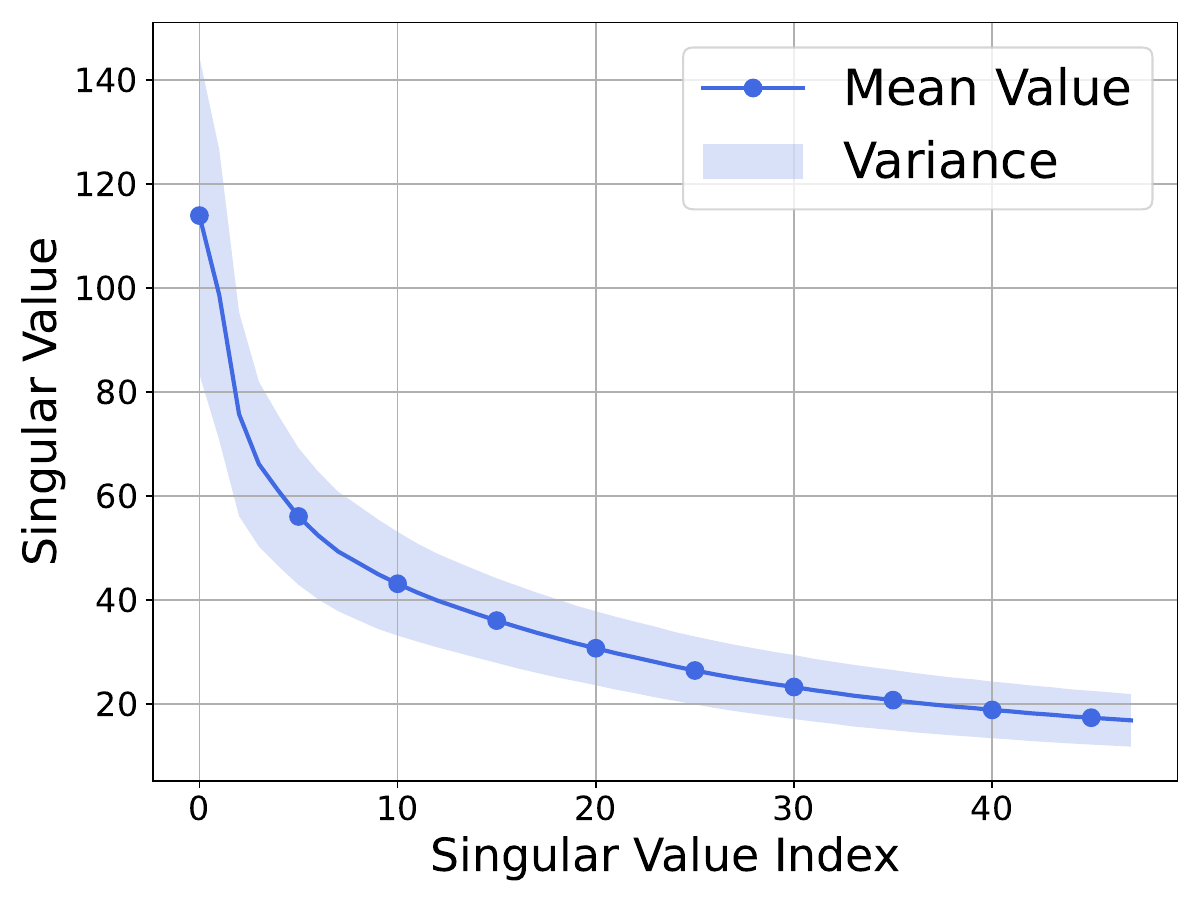}
    }
    \caption{More singular value distribution of all
    timesteps of HunyuanVideo-13B~\citep{kong2024hunyuanvideo}.}
\label{fig:more_svd}
\end{figure}

To verify the motivation of using the temporal-stable component of the second-order residual, we visualize more singular value distribution of all timesteps in Fig.~\ref{fig:more_svd}. It can be seen that in different blocks of different models, the second-order residuals at different time steps show considerable stability. Therefore, the second-order residual after SVD can retain the characteristics of time stability, further reduce the variance caused by different time steps, and have better approximation effect.

\begin{figure}[h]
    \centering
    \subfloat[][\textit{block.11}.]{
        \includegraphics[width=0.18\linewidth]{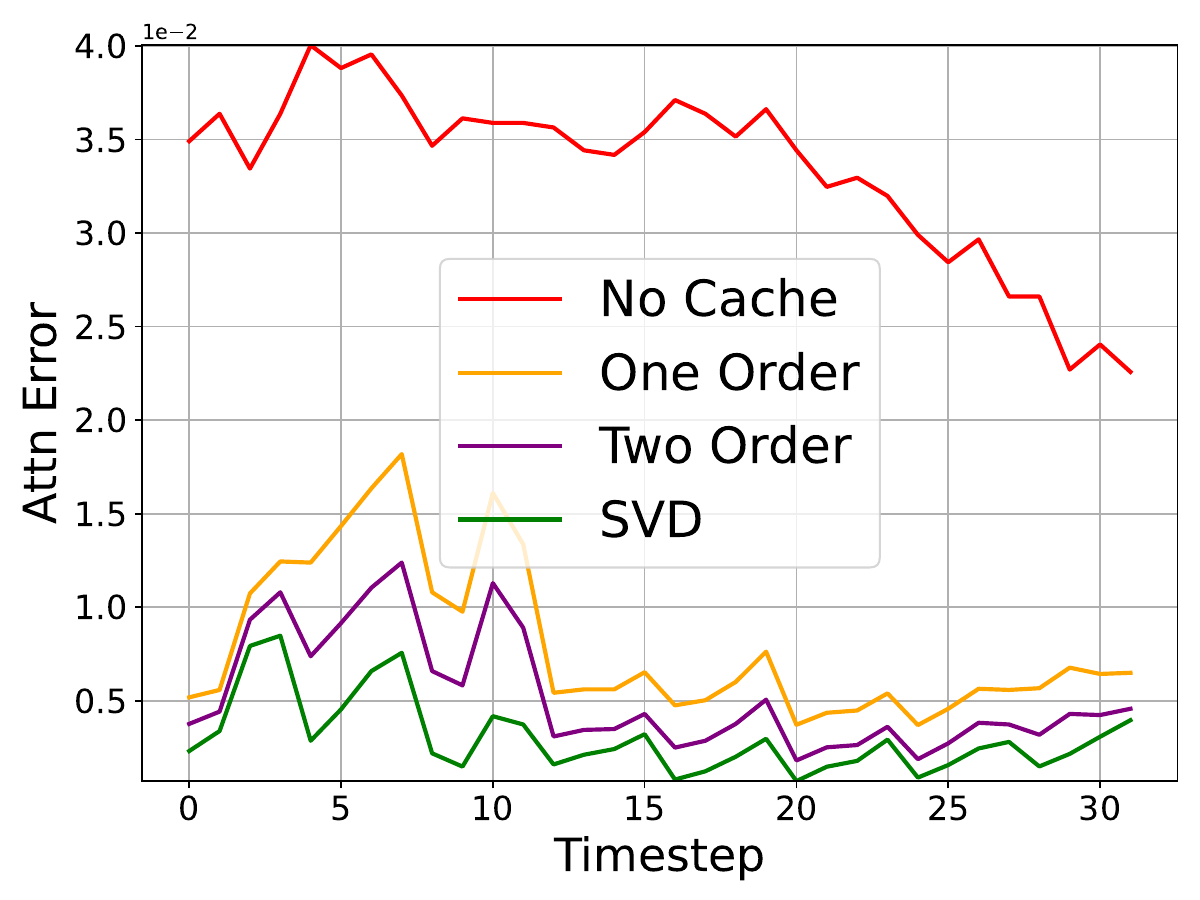}
    }
    \subfloat[][\textit{block.12}.]{
        \includegraphics[width=0.18\linewidth]{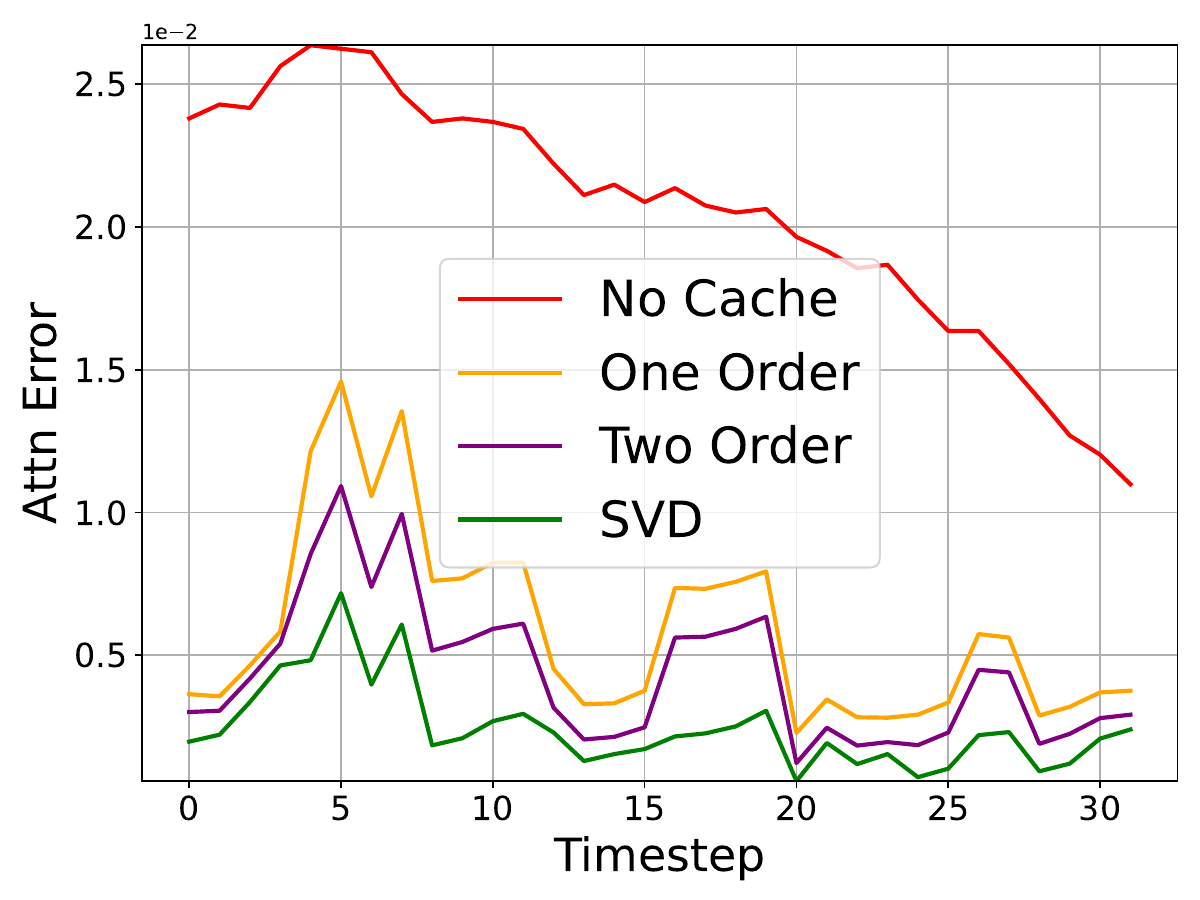}
    }
    \subfloat[][\textit{block.13}.]{
        \includegraphics[width=0.18\linewidth]{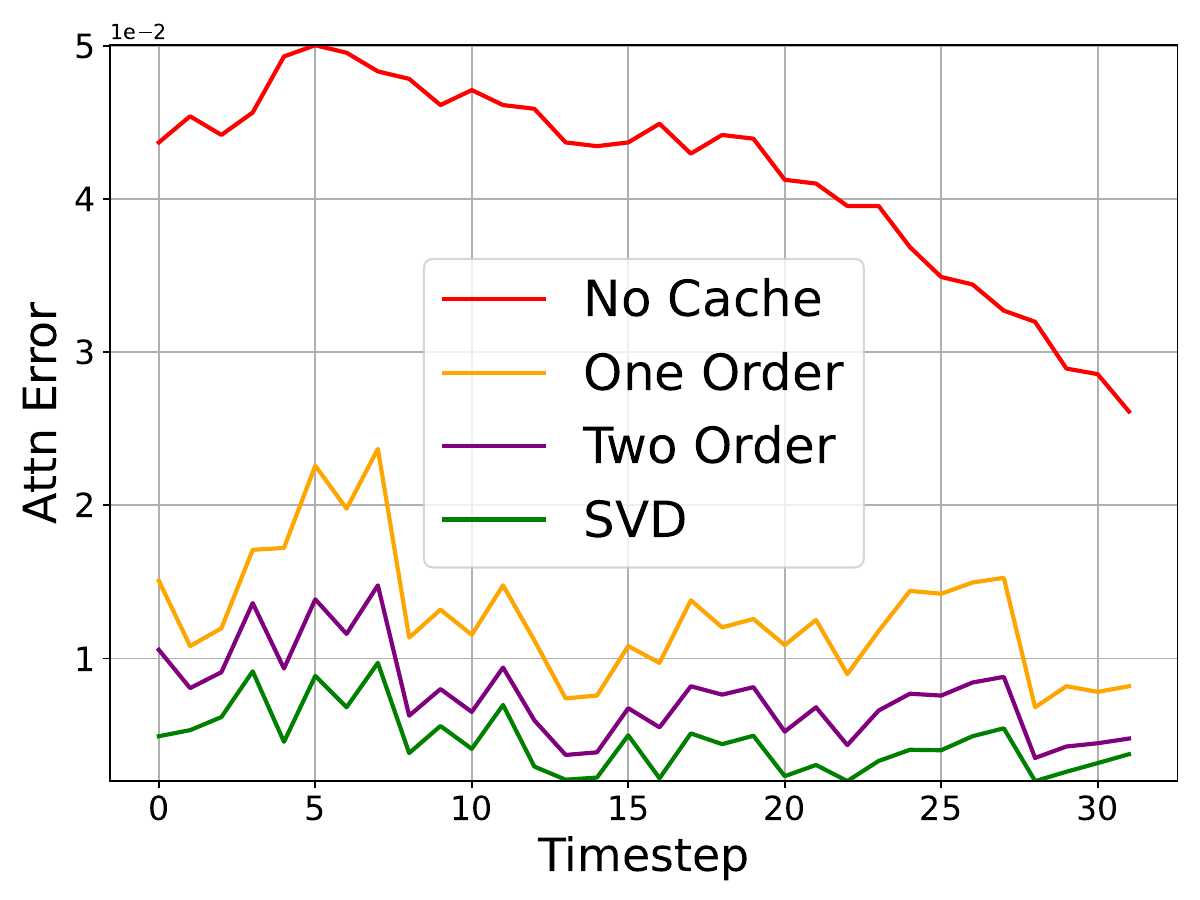}
    }
    \subfloat[][\textit{block.14}.]{
        \includegraphics[width=0.18\linewidth]{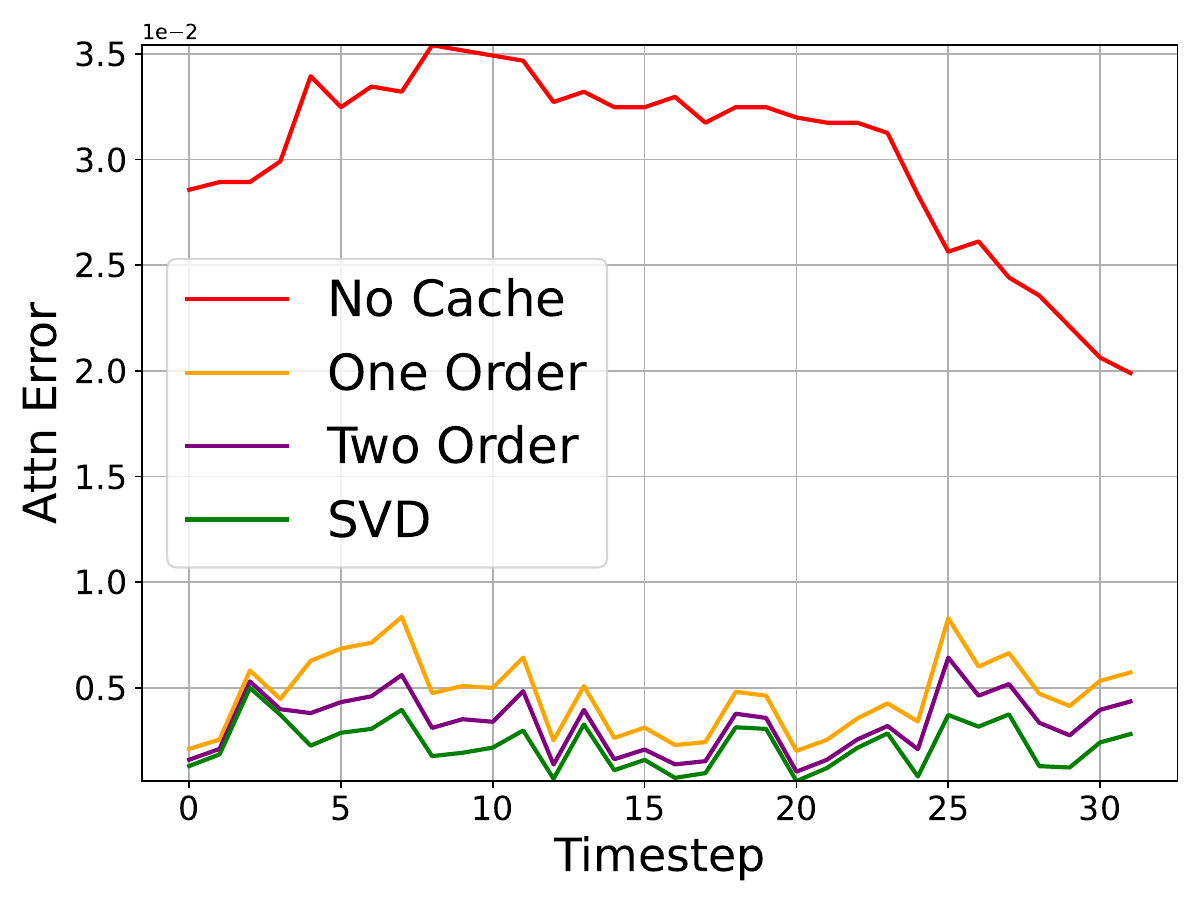}
    }
    \subfloat[][\textit{block.15}.]{
        \includegraphics[width=0.18\linewidth]{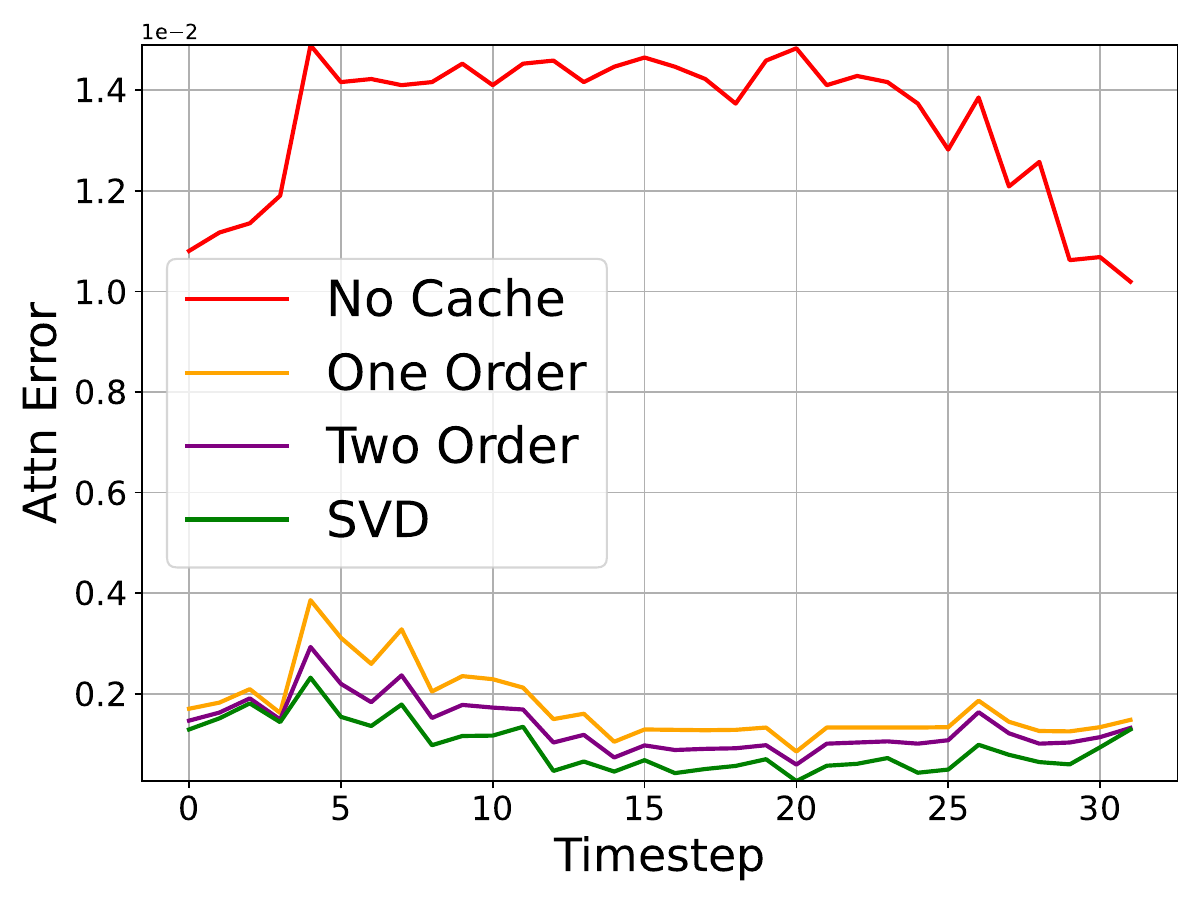}
    }
    \caption{More attention error comparison of HunyuanVideo-13B~\citep{kong2024hunyuanvideo}.}
\label{fig:more_attn_error}
\end{figure}

We further visualize more attention error comparison in Fig.~\ref{fig:more_attn_error}. It can be seen that the residual mechanism significantly reduces the attention error, which proves the importance of sparse attention reparameterization. At the same time, compared with the first-order residual, the second-order residual further reduces the attention error, which proves the necessity of introducing the second-order residual after quantization. Also, the second-residual after using SVD can further reduce the attention error, which proves that we have indeed extracted the temporally stable component and achieved the best attention approximation effect.

\section{More ablation study}
\label{sec:more_ablation}

Here, we provide more ablation study about the proposed Multi-Scale Salient Attention Distillation (MSAD) and Second-Order Sparse Attention Reparameterization (SSAR).

\begin{wraptable}{r}{0.3\linewidth}
    \vspace{-0.2in}
  \centering
  \begin{minipage}[b]{\linewidth}
    \caption{\rebuttal{Ablation on $s$ and $k$ used in attention distillation.}}
      \resizebox{\linewidth}{!}{
      \input{tables/abla_distill}}
        \label{tab:ablation_single_distill}
    \end{minipage}
     \vspace{-0.3in}
\end{wraptable}

\rebuttal{We first study the pooling stride $s$ used in Eq.~\ref{eq:global_distill} and salient token $k$ in Eq.~\ref{eq:local_distill} to verify the hyperparameter selection of both global and local distillation. We present the results in Tab.~\ref{tab:ablation_single_distill}. It can be seen that different hyperparameters can improve the distillation performance. This shows that our distillation method is both effective and robust, which is insensitive to hyperparameters. This also demonstrates that the memory-efficient distillation are effective enough and we do not have to use the giant complete attention map to supervise the attention module. Higher $s$ and lower $k$ can reduce memory, but typically harm performance. Yet we identify that decreasing $s$ and increasing $k$ also brings little improvement. Since $s=128$ and $k=256$ are both effective and efficient as shown in Fig.~\ref{fig:attn_distill_memory}, we choose this balanced selection.}

\begin{table}[h!]
\caption{\rebuttal{Ablation on SVD used in SSAR.}}
\label{tab:ablation_svd}
\begin{center}
\resizebox{0.6\linewidth}{!}{
\input{tables/abla_svd}}    
\end{center}
\end{table}

\rebuttal{We then study the top-$r$ components in SVD used in Eq.~\ref{eq:final_second_order}, and present the results in Tab.~\ref{tab:ablation_svd}. Compared with the original second-order residual, it can be seen that the different selection of $r$ in SVD can improve the temporal stability of the second-order residual and bring better performance. In our experiment, we chose $r=16$ as it achieves good performance.} We further explore higher-order residual effectiveness on attention approximation. Compared with the Second-Order residual, Third-Order residual only slightly improves PSNR from 18.70 to 18.68 and decreases the performance on VQA, SSIM, and LPIPS. This indicates that the stability brought by higher-order residuals will gradually saturate, and we attribute it to the additional noise brought by longer time series information on higher-order residuals. The second-order residual not only stabilizes the first-order residual, but SVD can further reduce spatiotemporal noise.

\begin{wraptable}{r}{0.3\linewidth}
    \vspace{-0.2in}
  \centering
  \begin{minipage}[b]{\linewidth}
    \caption{Ablation on $\lambda_{*}$ used in Eq.~\ref{eq:total_loss}.}
      \resizebox{\linewidth}{!}{
      \input{tables/distill_hyper}}
        \label{tab:ablation_distill_hyper}
    \end{minipage}
     \vspace{-0.3in}
\end{wraptable}

We then study the weight factor used in Eq.~\ref{eq:total_loss} to verify the distillation robustness of hyperparameters. We present the results in Tab.~\ref{tab:ablation_distill_hyper}. The values are selected by controlling the distillation term to be of the same order of magnitude as $\mathcal{L}_{\text{quant}}$. It can be seen that different weight factors improve the model performance. This shows that our distillation method is not only effective but also insensitive to the choice of hyperparameters, indicating its generalization and effectiveness. Since $\lambda_{\text{global}}=1e-4$ and $\lambda_{\text{local}}=1e-4$ are good enough, and the hyperparameter selection is robust, we do not further fine-tune the hyperparameter selection.

\rebuttal{
We further compare our Multi-Scale Salient Attention Distillation (MSAD) with full-attention distillation (using the complete FP attention map as the target) on Wan2.1-1.3B~\citep{wan2025wan} under W4A8 quantization. The results are shown in Tab.~\ref{tab:full_attention}. MSAD achieves nearly identical performance to full-attention distillation. The results highlight MSAD’s efficiency advantages while maintaining comparable performance.
}

\begin{table}[h!]
\caption{\rebuttal{Ablation study on full-attention distillation.}}
\label{tab:full_attention}
\begin{center}
\resizebox{0.9\linewidth}{!}{
\begin{tabular}{l|ccccc}
\toprule
\textbf{Method} & Resolution & PSNR$_{\mathbf{\red{\uparrow}}}$ & LPIPS$_{\mathbf{\red{\downarrow}}}$ & Attention Memory Cost (GB)$_{\mathbf{\red{\downarrow}}}$ & Calibration Time (Hours)$_{\mathbf{\red{\downarrow}}}$ \\
\midrule  

\multicolumn{6}{c}{\cellcolor[gray]{0.92}\texttt{Wan}2.1 1.3B} \\
\midrule

Full Attention & (17472, 17472) & 15.25 & 0.338 & 6.82 & 1.86 \\
\rowcolor{mycolor!30} \textbf{MSAD ($s=64$)} & (273, 273) & 15.23 & 0.338 & 0.17 & 0.66 \\
\rowcolor{mycolor!30} 
\textbf{MSAD ($s=128$)} & (137, 137) & 15.22 & 0.338 & 0.14 & 0.64 \\
\rowcolor{mycolor!30} 
\textbf{MSAD ($s=256$)} & (69, 69) & 15.21 & 0.339 & 0.13 & 0.63 \\

\bottomrule
\end{tabular}}
\end{center}
\end{table}

\rebuttal{
\textbf{Effect of attention density.} We conduct an ablation study on attention density, analyzing the trade-off between performance and inference speed. The results are presented in Tab.~\ref{tab:abla_interval}. As shown, a 25\% density offers a good balance, achieving a significant 1.55$\times$ speedup with minimal performance degradation (PSNR of 18.72). A 15\% density further boosts the speedup to 1.74$\times$ while maintaining acceptable performance (PSNR of 18.22). Based on these results, we selected 25\% and 15\% density for the experiments presented in the main paper. The 25\% density provides a strong baseline for high performance with good acceleration, while the 15\% density demonstrates the potential for even greater inference speedup at a slightly decreased performance trade-off.
}

\begin{table}[h!]
\caption{Calibration computation resource report. PTQ denotes naive Post-Training Quantization without attention distillation.}
\label{tab:calib_resource}
\begin{center}
\resizebox{0.7\linewidth}{!}{
\input{tables/calib}}
\end{center}
\end{table}

\section{Calibration Computation Resource}
\label{sec:calib_efficiency}

We study the calibration computation resource of each of our proposed methods and the overall pipeline. As Second-Order Sparse Attention Reparameterization (SSAR) is used for only inference, for calibration, we only add Multi-Scale Salient Attention Distillation (MSAD) compared to naive Post-Training Quantization (PTQ). We present the calibration resource in Tab.~\ref{tab:calib_resource}. Compared with naive PTQ, our \textit{Global Distillation} only brings an average of 0.8\% extra time burden and almost no additional memory consumption because of its efficient low-resolution attention operation. Also, our \textit{Local Distillation} only needs to calculate the token saliency distribution once before each block calibration and reuse the salient token index in each optimization iteration, which is also very efficient. These two distillation methods are not only efficient but also can effectively alleviate the attention shift caused by quantization and improve the video generation effect.  QuantSparse has significantly improved the model performance by combining two distillation methods, while ensuring high efficiency.

To further prove the effectiveness of proposed Second-Order Sparse Attention Reparameterization (SSAR), we present the inference burden brought by SSAR in Tab.~\ref{tab:sparse_attn_resource}. Compared with Non-Reparameterization, the cache-based method only requires one additional matrix addition operation for the sparse attention output, which is very efficient. Therefore, the cache-based method will only bring little additional latency and memory burden. Furthermore, the second-order residual can store and calculate the second-order term and the first-order term together. Therefore, compared with the first-order residual, the second-order residual only requires an additional second order term calculation, but significantly improves the sparse attention performance under quantization, and improves the PSNR from 17.08 to 18.68 under Wan2.1-14B~\citep{wan2025wan}. In addition, using SVD to extract the temporally stable component of second-order residuals brings almost no additional consumption, but can further improve the effect of second-order residuals, which further decreases LPIPS from 0.258 to 0.240 under Wan2.1-14B.

\begin{table}[h!]
\caption{Sparse attention reparameterization resource report. `None' denotes Non-Reparameterization.}
\label{tab:sparse_attn_resource}
\begin{center}
\resizebox{0.7\linewidth}{!}{
\input{tables/sparse_attn}}
\end{center}
\end{table}

\section{Combination with other acceleration techniques}
\label{sec:combine_more_accele}

To further validate the integration ability of QuantSparse with other acceleration techniques, we combined it with existing attention quantization techniques SageAttention~\citep{zhang2024sageattention} and cache techniques TeaCache~\citep{liu2024teacache}, and presented the results in Tab.~\ref{tab:efficiency_more}. All the experiments are conducted on Wan2.1-14B~\citep{wan2025wan} under W4A8 quantization setting. We apply SageAttention by quantizing attention into 8-bit. For TeaCache, we set the threshold as 0.16 to ensure performance. 

It can be seen that, despite retaining only 15\% attention density under W4A8 quantization, the combination of QuantSparse and SageAttention still incurs almost no performance loss. This indicates that QuantSparse is highly friendly to sparsification and quantization, fully demonstrating the necessity of attention distillation and second-order reparameterization. Although further adding TeaCache may result in a slight performance decrease, it can bring significant additional inference acceleration. This provides a further trade-off between performance and inference speed, and also demonstrates the effectiveness of combining QuantSparse with cache-based methods.

\begin{figure}[h!]
  \centering
  \includegraphics[width=1.0\linewidth]{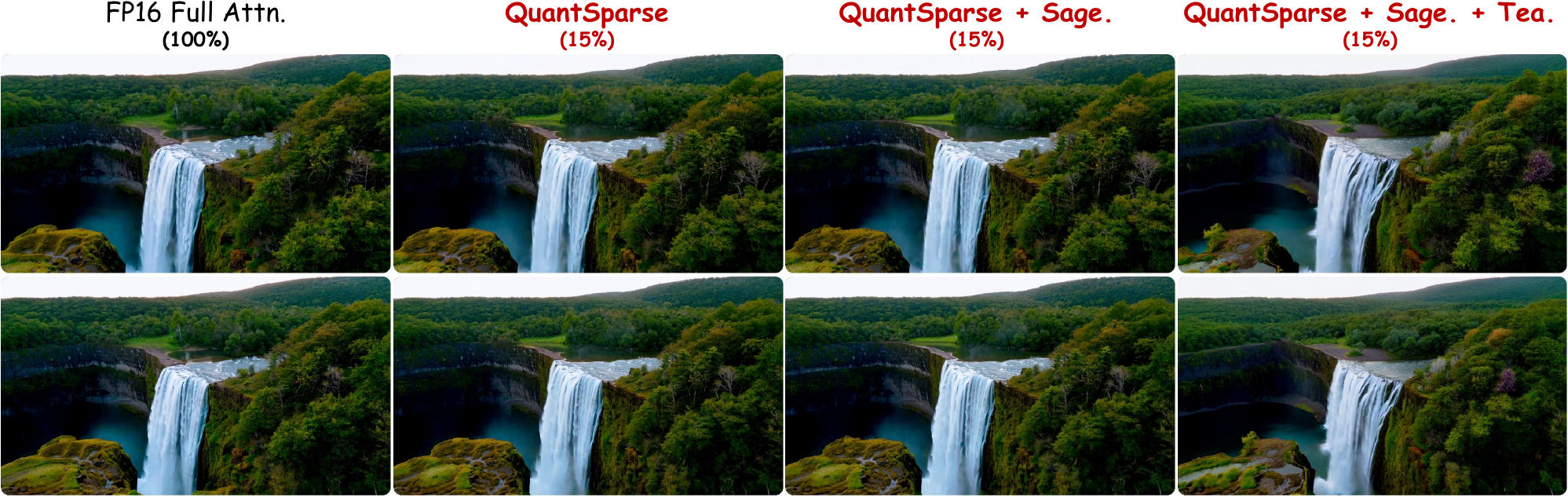}
  \caption{Combining with other acceleration techniques visualization on Wan2.1-14B under W4A8 quantization setting.}
  \label{fig:more_compare}
\end{figure}

We further provide more visualization results in Fig.~\ref{fig:more_compare}. It can be seen that the combination of QuantSparse and other acceleration techniques not only shows almost no decrease in metrics but also maintains good visual effects without producing any decrease in visual quality.

\begin{table}[h!]
\caption{More efficiency comparison under W4A8 quantization setting. Sage. denotes SageAttention~\citep{zhang2024sageattention}. Tea. denotes TeaCache~\citep{liu2024teacache}.}
\label{tab:efficiency_more}
\begin{center}
\resizebox{0.9\linewidth}{!}{
\input{tables/efficiency_more}}
\end{center}
\end{table}

\rebuttal{
\section{Image generation experiment}
QuantSparse is designed as a general framework for Diffusion Transformers (DiTs) and is not limited to video generation. To validate its generalizability, we conducted an experiment on Hunyuan-DiT~\citep{li2024hunyuandit}, a 1.5B parameters model targeting image generation. We evaluate on DrawBench~\citep{saharia2022drawbench} under W4A8 quantization and present the results in Tab.~\ref{tab:image_generation}. Even for image-generation DiTs, QuantSparse outperforms SOTA quantization baselines QuaRot~\citep{ashkboos2024quarot} and Q-VDiT~\citep{feng2025qvdit} while using only 40\% attention density. This confirms that our framework generalizes to DiT-based visual generation tasks and not limited to video generation.
}

\begin{table}[h!]
\caption{\rebuttal{Image generation experiment results on Hunyuan-DiT.}}
\label{tab:image_generation}
\begin{center}
\resizebox{0.6\linewidth}{!}{
\begin{tabular}{l|cccc}
\toprule
\textbf{Method} & Density$_{\mathbf{\red{\downarrow}}}$ &  PSNR$_{\mathbf{\red{\uparrow}}}$ & SSIM$_{\mathbf{\red{\uparrow}}}$ & LPIPS$_{\mathbf{\red{\downarrow}}}$ \\
\midrule  

\multicolumn{5}{c}{\cellcolor[gray]{0.92}\texttt{Hunyuan-DiT}} \\
\midrule

QuaRot & 100\% & 17.30 & 0.627 & 0.460 \\
Q-VDiT & 100\% & 19.32 & 0.658 & 0.347 \\
\rowcolor{mycolor!30} \textbf{QuantSparse} & 40\% & \textbf{20.34} & \textbf{0.692} & \textbf{0.289} \\

\bottomrule
\end{tabular}}
\end{center}
\end{table}

\section{The use of large language models (LLMs)}

In this paper, Large Language Models are only used as general-purpose auxiliary tools, primarily for document-level auxiliary tasks such as grammar checking and expression refinement. LLMs did not participate in the core conceptualization, method derivation, or experimental design of this research, nor did they contribute to any core writing content.

\section{More visual comparison}
\label{sec:more_visual}

In the following pages, we provide more visual comparisons of different-scale video-generation models. `Full Prec.' denotes the Full Precision model. (xx\%) denotes the attention density.

\begin{figure}[h!]
  \centering
  \includegraphics[width=0.93\linewidth]{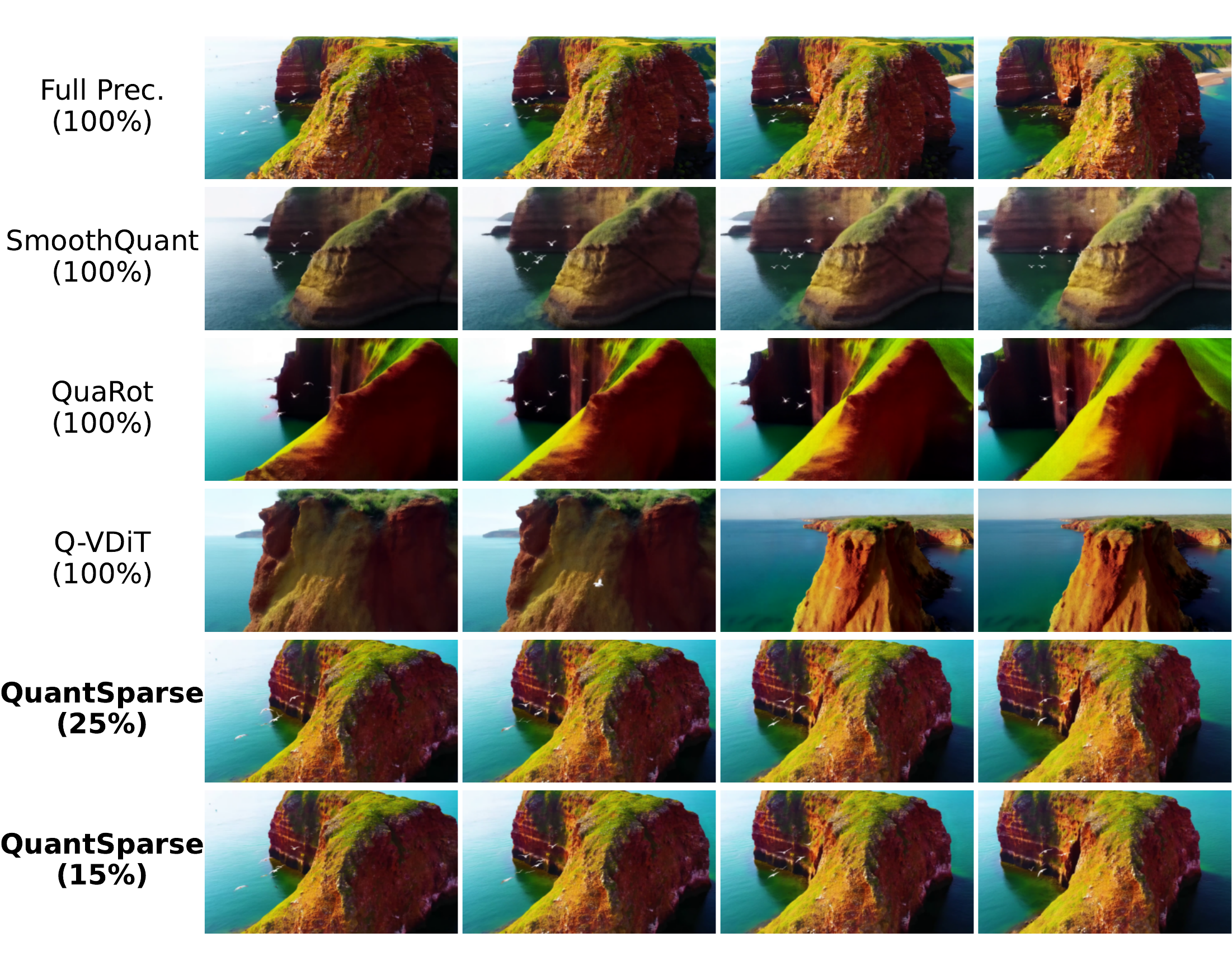}
  \caption{HunyuanVideo-13B results. 
  }
  \label{fig:hy_1}
\end{figure}

\begin{figure}[h!]
  \centering
  \includegraphics[width=0.93\linewidth]{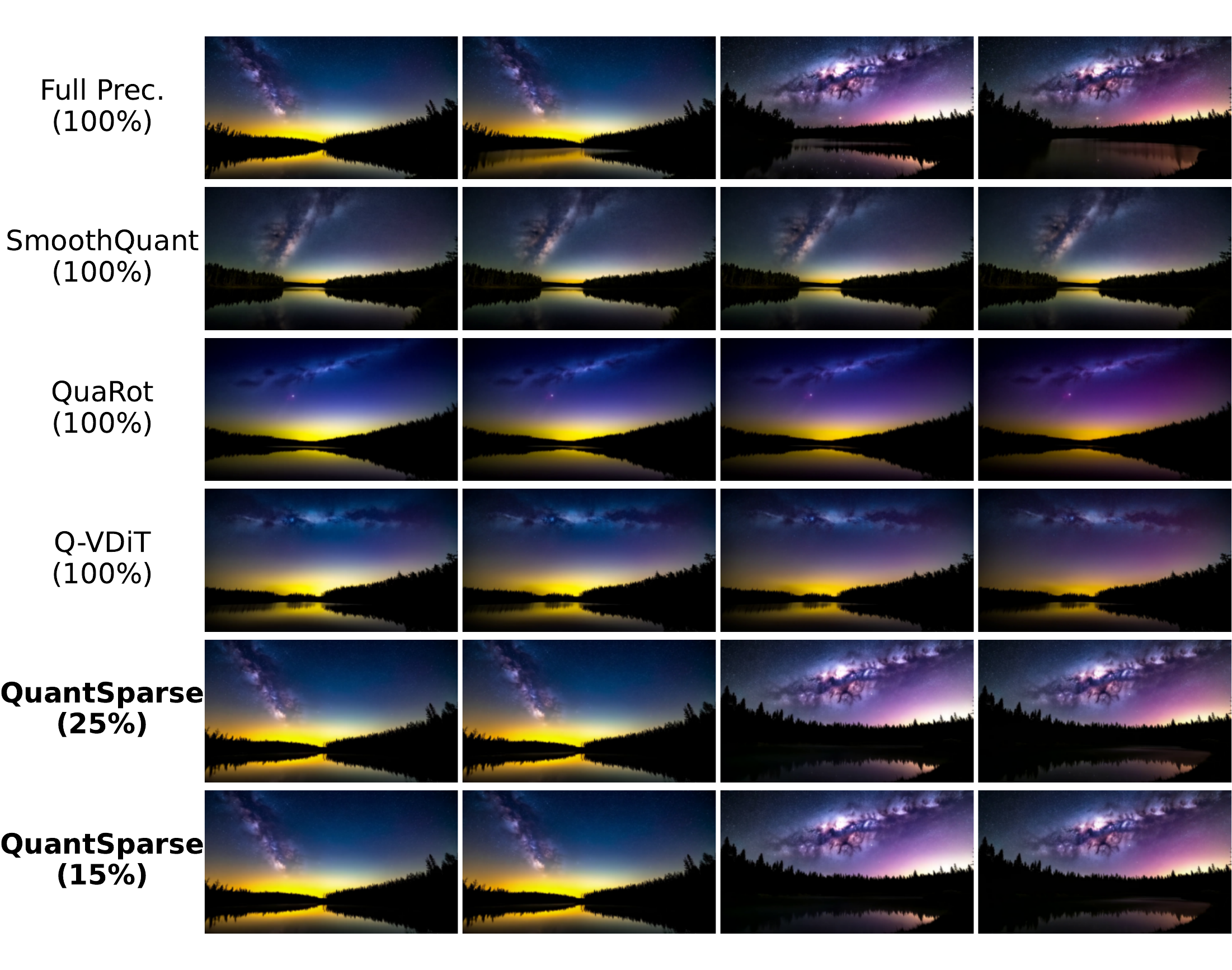}
   \vspace{-0.2in}
  \caption{HunyuanVideo-13B results. 
  }
  \label{fig:hy_2}
\end{figure}

\begin{figure}[h!]
  \centering
  \includegraphics[width=0.95\linewidth]{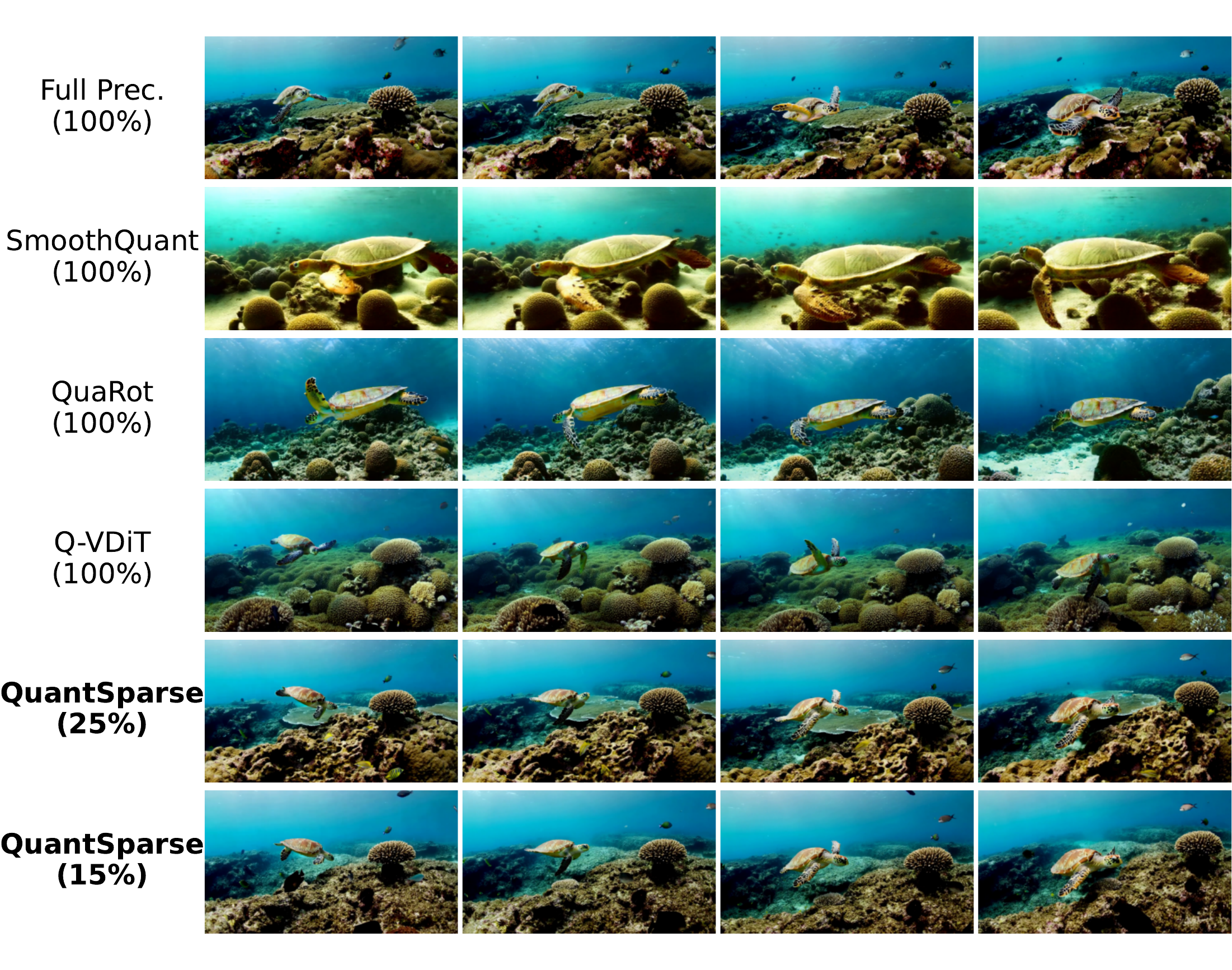}
   \vspace{-0.2in}
  \caption{Wan2.1-14B results. 
  }
  \label{fig:wan14_1}
\end{figure}

\begin{figure}[h!]
  \centering
  \includegraphics[width=0.95\linewidth]{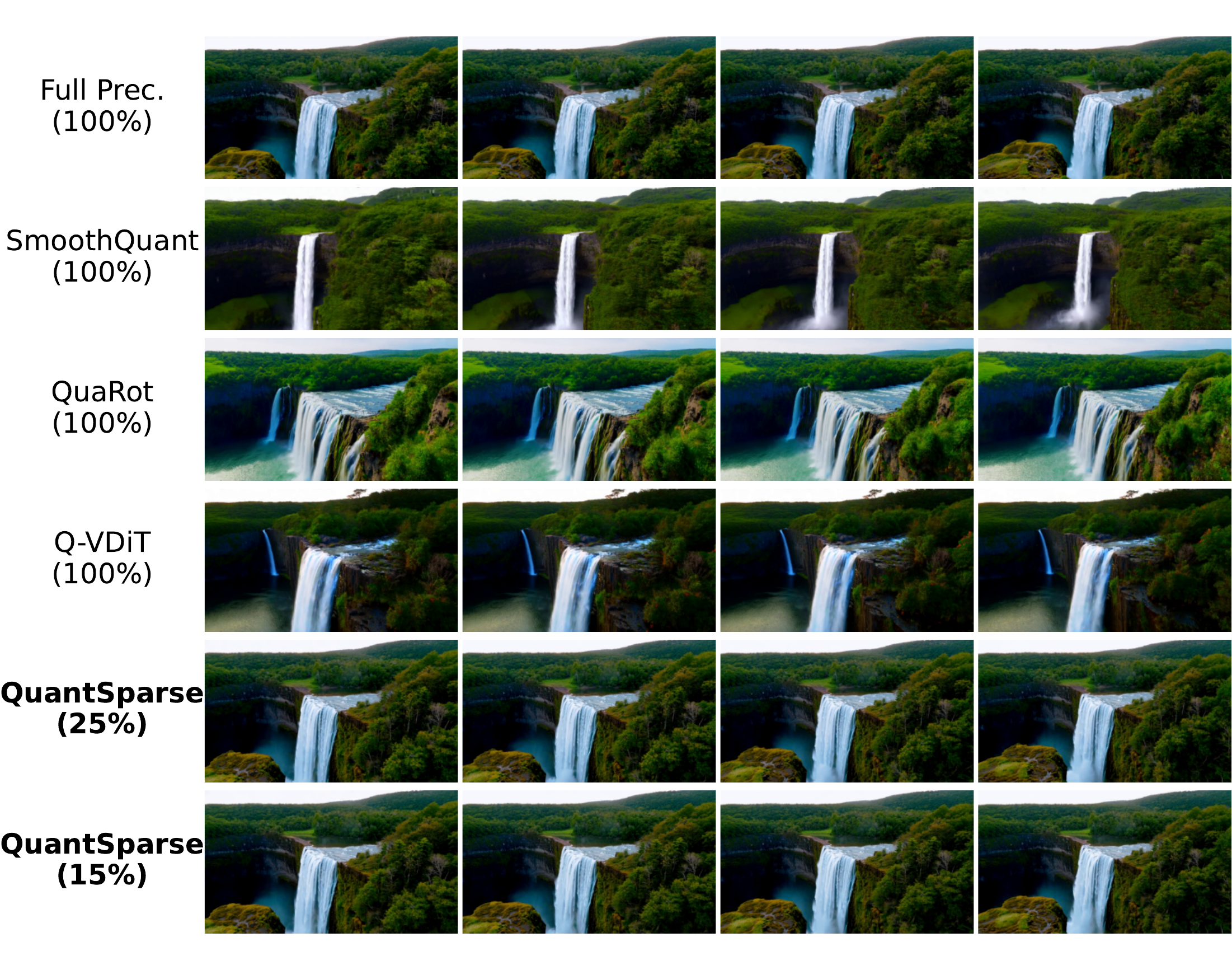}
   \vspace{-0.2in}
  \caption{Wan2.1-14B results. 
  }
  \label{fig:wan14_2}
\end{figure}

\begin{figure}[h!]
  \centering
  \includegraphics[width=1.0\linewidth]{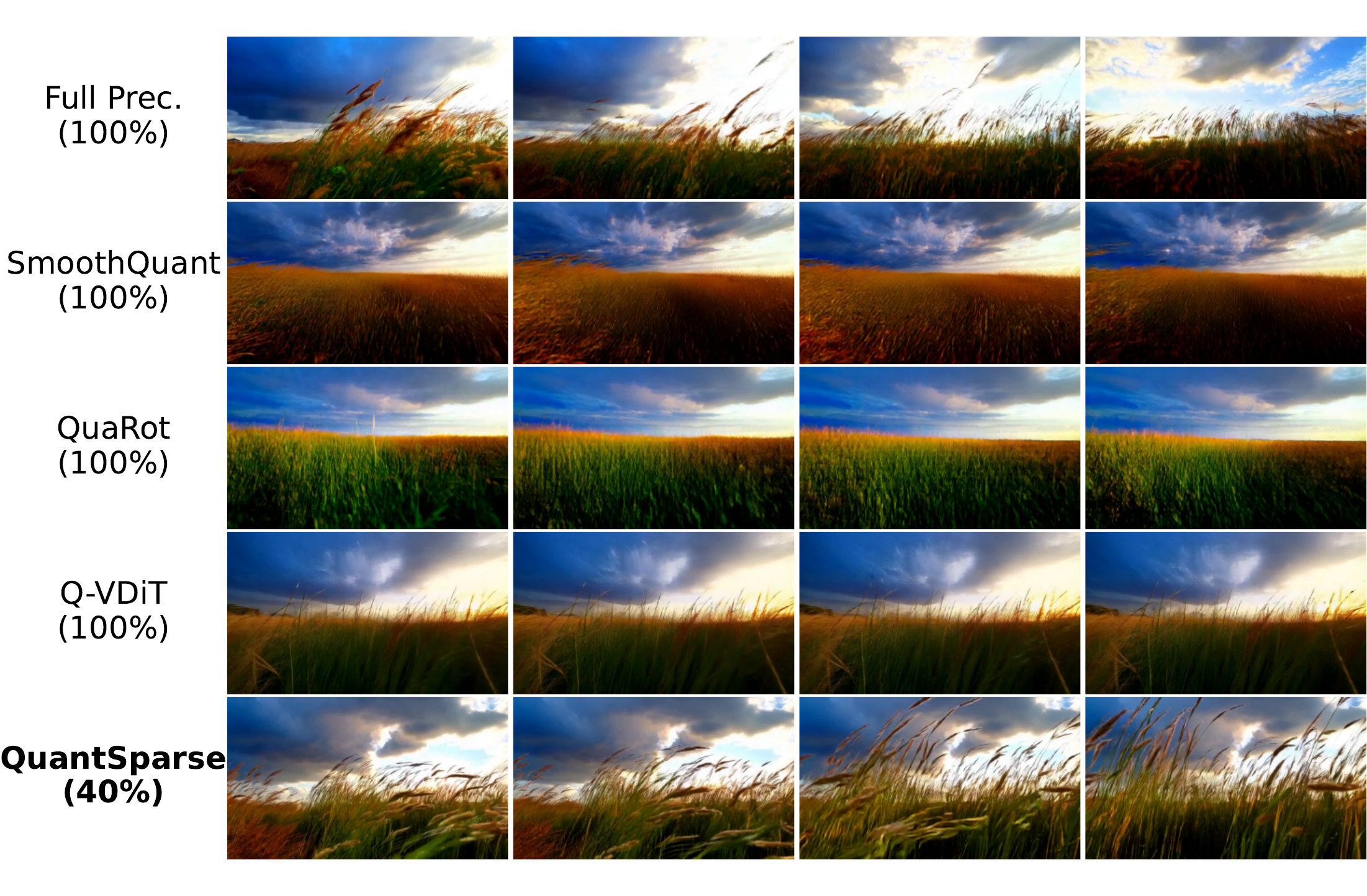}
   \vspace{-0.2in}
  \caption{Wan2.1-1.3B results. 
  }
  \label{fig:wan13_1}
\end{figure}

\begin{figure}[h!]
  \centering
  \includegraphics[width=1.0\linewidth]{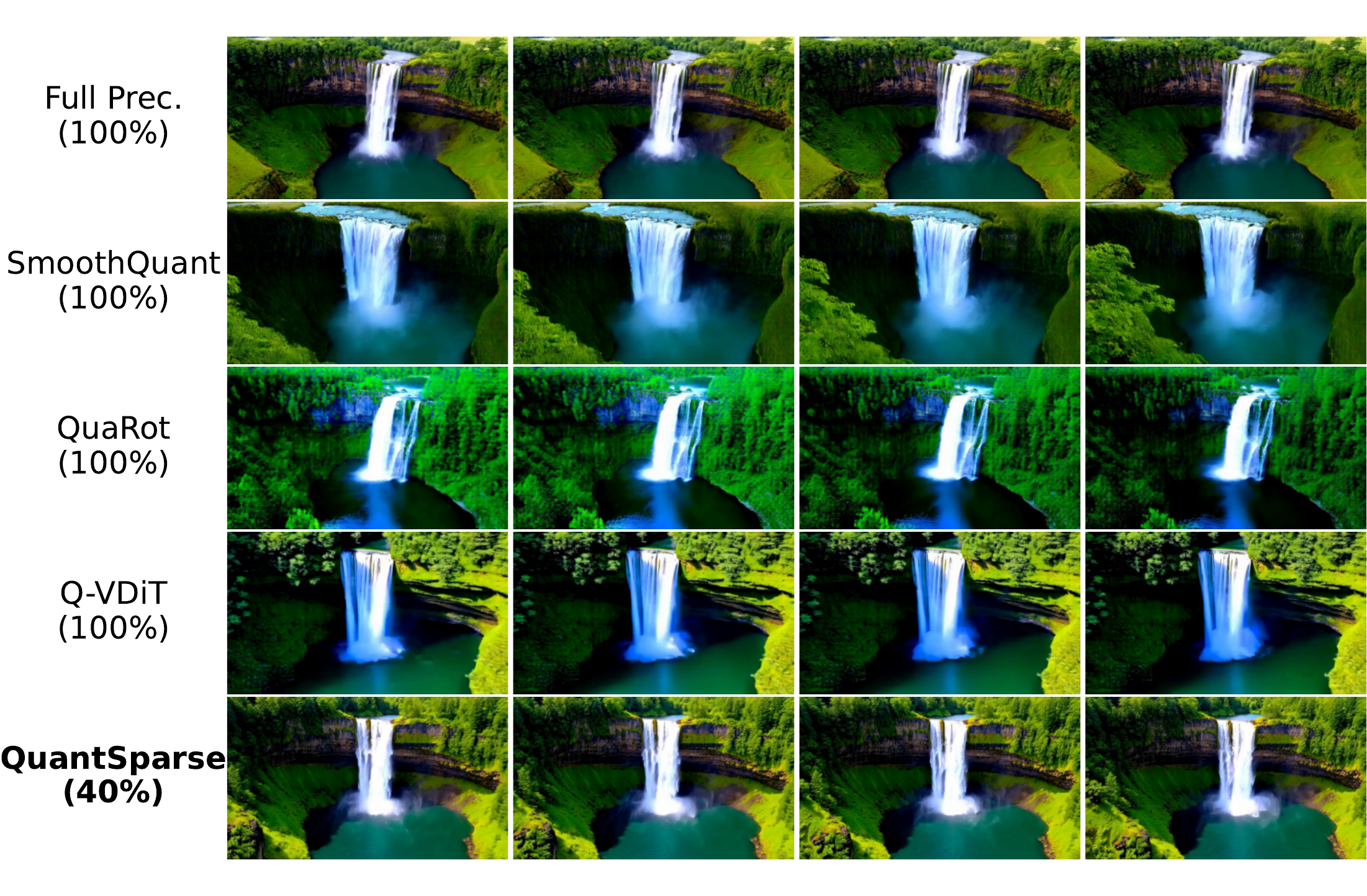}
   \vspace{-0.2in}
  \caption{Wan2.1-1.3B results. 
  }
  \label{fig:wan13_2}
\end{figure}

We also provide the used text prompt for each figure:

\begin{enumerate}
    \item Fig.~\ref{fig:hy_1}: \textit{A soaring drone footage captures the majestic beauty of a coastal cliff, its red and yellow stratified rock faces rich in color and against the vibrant turquoise of the sea. Seabirds can be seen taking flight around the cliff's precipices. As the drone slowly moves from different angles, the changing sunlight casts shifting shadows that highlight the rugged textures of the cliff and the surrounding calm sea. The water gently laps at the rock base and the greenery that clings to the top of the cliff, and the scene gives a sense of peaceful isolation at the fringes of the ocean. The video captures the essence of pristine natural beauty untouched by human structures.}
    \item Fig.~\ref{fig:hy_2}: \textit{A serene night scene in a forested area. The first frame shows a tranquil lake reflecting the star-filled sky above. The second frame reveals a beautiful sunset, casting a warm glow over the landscape. The third frame showcases the night sky, filled with stars and a vibrant Milky Way galaxy. The video is a time-lapse, capturing the transition from day to night, with the lake and forest serving as a constant backdrop. The style of the video is naturalistic, emphasizing the beauty of the night sky and the peacefulness of the forest.}
    \item Fig.~\ref{fig:wan14_1}: \textit{A serene underwater scene featuring a sea turtle swimming through a coral reef. The turtle, with its greenish-brown shell, is the main focus of the video, swimming gracefully towards the right side of the frame. The coral reef, teeming with life, is visible in the background, providing a vibrant and colorful backdrop to the turtle's journey. Several small fish, darting around the turtle, add a sense of movement and dynamism to the scene. The video is shot from a slightly elevated angle, providing a comprehensive view of the turtle's surroundings. The overall style of the video is calm and peaceful, capturing the beauty and tranquility of the underwater world.}
    \item Fig.~\ref{fig:wan14_2}: \textit{The video captures the majestic beauty of a waterfall cascading down a cliff into a serene lake. The waterfall, with its powerful flow, is the central focus of the video. The surrounding landscape is lush and green, with trees and foliage adding to the natural beauty of the scene. The camera angle provides a bird's eye view of the waterfall, allowing viewers to appreciate the full height and grandeur of the waterfall. The video is a stunning representation of nature's power and beauty.}
    \item Fig.~\ref{fig:wan13_1}: \textit{The dynamic movement of tall, wispy grasses swaying in the wind. The sky above is filled with clouds, creating a dramatic backdrop. The sunlight pierces through the clouds, casting a warm glow on the scene. The grasses are a mix of green and brown, indicating a change in seasons. The overall style of the video is naturalistic, capturing the beauty of the landscape in a realistic manner. The focus is on the grasses and their movement, with the sky serving as a secondary element. The video does not contain any human or animal elements.}
    \item Fig.~\ref{fig:wan13_2}: \textit{The video captures the majestic beauty of a waterfall cascading down a cliff into a serene lake. The waterfall, with its powerful flow, is the central focus of the video. The surrounding landscape is lush and green, with trees and foliage adding to the natural beauty of the scene. The camera angle provides a bird's eye view of the waterfall, allowing viewers to appreciate the full height and grandeur of the waterfall. The video is a stunning representation of nature's power and beauty.}
\end{enumerate}

\newpage

%% file: tables/sora_hy.tex
\begin{tabular}{l|c|c|cccccc}
\toprule
\multirow{3}{*}{\textbf{Method}} & \multirow{3}{*}{\textbf{\makecell{\#Bits \\ (W/A)}}} & \multirow{3}{*}{\textbf{Density}$_{\mathbf{\red{\downarrow}}}$} & \multicolumn{6}{c}{\textbf{Quality}} \\
\cmidrule(lr){4-9}  
& & & \multicolumn{3}{c}{\textit{Video Quality Metrics}} & \multicolumn{3}{c}{\textit{FP Diff. Metrics}} \\
\cmidrule(lr){4-6}
\cmidrule(lr){7-9} 
& & & CLIPSIM$_{\mathbf{\red{\uparrow}}}$ & VQA$_{\mathbf{\red{\uparrow}}}$ & $\Delta$FSCore$_{\mathbf{\red{\downarrow}}}$ & PSNR$_{\mathbf{\red{\uparrow}}}$ & SSIM$_{\mathbf{\red{\uparrow}}}$ & LPIPS$_{\mathbf{\red{\downarrow}}}$ \\
\midrule  

\multicolumn{9}{c}{\cellcolor[gray]{0.92}HunyuanVideo 13B ($\texttt{CFG}=6.0, 720\times1280p, \texttt{frames}=60$)} \\
\midrule

Full Prec. & 16/16 & 100\% & 0.184 & 81.23 & 0.000 & - & - & -  \\
\midrule

SmoothQuant & 6/6 & 100\% & 0.180 & 69.55 & 1.406 & 15.91 & 0.553 & 0.411 \\
QuaRot & 6/6 & 100\% & \underline{0.182} & 72.28 & 0.546 & 16.99 & 0.590 & 0.378 \\
ViDiT-Q & 6/6 & 100\% & \underline{0.182} & 72.36 & 0.937 & 18.24 & 0.623 & 0.335 \\
Q-VDiT & 6/6 & 100\% & \underline{0.182} & 73.68 & 1.232 & 21.02 & \underline{0.675} & 0.264 \\
QuaRot+SVG & 6/6 & 25\% & 0.181 & 72.57 & 0.718 & 16.85 & 0.581 & 0.385 \\
Q-VDiT+SVG & 6/6 & 25\% & 0.181 & 72.59 & 1.405 & 20.38 & 0.658 & 0.284 \\
QuaRot+SVG & 6/6 & 15\% & 0.181 & 72.60 & 0.997 & 16.85 & 0.578 & 0.394 \\
Q-VDiT+SVG & 6/6 & 15\% & 0.181 & 72.04 & 1.763 & 19.94 & 0.644 & 0.307 \\
\rowcolor{mycolor!30}\textbf{QuantSparse} & 6/6 & 25\% & \textbf{0.183} & \underline{81.17} & \underline{0.435} & \textbf{22.71} & \textbf{0.720} & \textbf{0.221} \\
\rowcolor{mycolor!30}\textbf{QuantSparse} & 6/6 & 15\% & \textbf{0.183} & \textbf{82.26} & \textbf{0.328} & \underline{22.68} & \textbf{0.720} & \underline{0.224} \\
\midrule

\multicolumn{9}{c}{\cellcolor[gray]{0.92}\texttt{Wan}2.1 14B ($\texttt{CFG}=5.0, 720\times1280p, \texttt{frames}=80$)} \\
\midrule

Full Prec. & 16/16 & 100\% & 0.182 & 90.79 & 0.000 & - & - & -  \\
\midrule

SmoothQuant & 6/6 & 100\% & 0.178 & 62.25 & 0.363 & 13.06 & 0.404 & 0.656 \\
QuaRot & 6/6 & 100\% & 0.180 & 66.56 & 0.313 & 13.59 & 0.409 & 0.566 \\
ViDiT-Q & 6/6 & 100\% & 0.180 & 71.26 & 0.251 & 15.30 & 0.513 & 0.376 \\
Q-VDiT & 6/6 & 100\% & 0.180 & 89.10 & 0.082 & \underline{18.13} & 0.610 & \underline{0.264} \\
QuaRot+SVG & 6/6 & 25\% & 0.179 & 67.64 & 0.336 & 13.60 & 0.407 & 0.555 \\
Q-VDiT+SVG & 6/6 & 25\% & 0.179 & 88.29 & 0.091 & 16.69 & 0.563 & 0.323 \\
QuaRot+SVG & 6/6 & 15\% & 0.180 & 60.14 & 0.396 & 13.55 & 0.399 & 0.567 \\
Q-VDiT+SVG & 6/6 & 15\% & 0.179 & 85.26 & 0.182 & 15.94 & 0.532 & 0.367 \\
\rowcolor{mycolor!30}\textbf{QuantSparse} & 6/6 & 25\% & \textbf{0.182} & \underline{89.96} & \textbf{0.002} & \textbf{18.67} & \textbf{0.622} & \textbf{0.240} \\
\rowcolor{mycolor!30}\textbf{QuantSparse} & 6/6 & 15\% & \underline{0.181} & \textbf{92.87} & \underline{0.060} & \textbf{18.67} & \underline{0.616} & 0.277 \\

\bottomrule
\end{tabular}

%% file: tables/vbench.tex
\begin{tabular}{lcccccccccc}
\toprule
\textbf{Method} & \textbf{\makecell{\#Bits \\ (W/A)}} & \textbf{Density} & IQ & AQ & MS & DD & BC & SuC & ScC & OC \\
\midrule

\multicolumn{11}{c}{\cellcolor[gray]{0.92}\texttt{Wan}2.1 1.3B ($\texttt{CFG}=6.0, 480\times832p, \texttt{frames}=80$)} \\
\midrule

Full Prec. & 16/16 & 100\% & 64.05 & 57.86 & 97.03 & 87.50 & 94.94 & 93.00 & 16.72 & 23.16 \\
\midrule
QuaRot+SVG & 6/6 & 40\% & 62.53 & 52.16 & 95.48 & 81.94 & 93.65 & 89.20 & 12.43 & 22.42 \\
Q-VDiT+SVG & 6/6 & 40\% & 64.01 & 53.89 & 96.25 & 81.94 & 94.23 & 91.78 & 17.81 & 22.90 \\
\rowcolor{mycolor!30}\textbf{QuantSparse} & 6/6 & 40\% & \textbf{64.96} & \textbf{56.44} & \textbf{96.68} & \textbf{83.33} & \textbf{94.84} & \textbf{92.56} & \textbf{18.46} & \textbf{23.12} \\
\midrule

QuaRot+SVG & 4/8 & 40\% & 54.45 & 43.60 & 96.29 & 73.61 & 94.99 & 87.02 & 8.14 & 18.88 \\
Q-VDiT+SVG & 4/8 & 40\% & 56.08 & 48.12 & 97.27 & 61.11 & 95.86 & 89.72 & 10.32 & 19.89 \\
\rowcolor{mycolor!30}\textbf{QuantSparse} & 4/8 & 40\% & \textbf{64.41} & \textbf{58.00} & \textbf{97.35} & \textbf{87.50} & \textbf{94.99} & \textbf{93.02} & \textbf{18.24} & \textbf{23.31} \\
\midrule

\multicolumn{11}{c}{\cellcolor[gray]{0.92}HunyuanVideo 13B ($\texttt{CFG}=6.0, 512\times768p, \texttt{frames}=60$)} \\
\midrule

Full Prec. & 16/16 & 100\% & 62.30 & 62.49 & 99.00 & 56.94 & 98.08 & 95.30 & 33.36 & 26.85  \\
\midrule

QuaRot+SVG & 6/6 & 25\% & 56.82 & 57.23 & 97.93 & 40.00 & 97.75 & 95.10 & 23.98 & 25.63 \\
Q-VDiT+SVG & 6/6 & 25\% & 59.22 & 58.77 & 97.96 & 40.00 & 97.60 & 95.68 & 26.80 & 25.87 \\
QuaRot+SVG & 6/6 & 15\% & 53.95 & 56.43 & 97.84 & 38.89 & 97.48 & 94.40 & 23.36 & 25.57 \\
Q-VDiT+SVG & 6/6 & 15\% & 57.43 & 58.02 & 97.84 & 38.61 & 97.07 & 95.20 & 24.27 & 25.74  \\
\rowcolor{mycolor!30}\textbf{QuantSparse} & 6/6 & 25\% & \textbf{60.24} & \textbf{59.06} & \textbf{99.01} & \textbf{43.06} & \textbf{98.33} & \textbf{96.06} & \textbf{28.42} & \underline{26.62} \\
\rowcolor{mycolor!30}\textbf{QuantSparse} & 6/6 & 15\% & \underline{59.54} & \underline{58.87} & \underline{98.95} & \underline{40.28} & \underline{98.08} & \underline{95.84} & \underline{27.69} & \textbf{26.63} \\
\midrule

QuaRot+SVG & 4/8 & 25\% & 45.81 & 56.59 & 98.26 & 22.22 & 98.18 & 95.78 & 21.00 & 24.64 \\
Q-VDiT+SVG & 4/8 & 25\% & 44.94 & 56.62 & 98.36 & 23.61 & 97.98 & 96.06 & 18.53 & 24.81 \\
QuaRot+SVG & 4/8 & 15\% & 43.51 & 55.35 & 98.21 & 20.83 & 97.21 & 95.15 & 18.31 & 24.50 \\
Q-VDiT+SVG & 4/8 & 15\% & 42.16 & 55.32 & 98.32 & 20.83 & 97.96 & 95.48 & 16.64 & 24.68 \\
\rowcolor{mycolor!30}\textbf{QuantSparse} & 4/8 & 25\% & \textbf{59.85} & \textbf{59.37} & \textbf{99.08} & \underline{38.89} & \textbf{98.32} & \textbf{96.41} & \underline{29.80} & \textbf{26.92} \\
\rowcolor{mycolor!30}\textbf{QuantSparse} & 4/8 & 15\% & \underline{59.27} & \underline{59.20} & \underline{99.04} & \textbf{40.28} & \underline{98.21} & \underline{96.18} & \textbf{30.31} & \textbf{26.92} \\
\midrule

\multicolumn{11}{c}{\cellcolor[gray]{0.92}\texttt{Wan}2.1 14B ($\texttt{CFG}=5.0, 480\times832p, \texttt{frames}=80$)} \\
\midrule

Full Prec. & 16/16 & 100\% & 63.38 & 59.56 & 96.73 & 86.11 & 96.71 & 90.84 & 28.13 & 25.68  \\
\midrule

QuaRot+SVG & 6/6 & 25\% & 61.77 & 54.13 & 95.89 & 45.83 & 94.78 & 90.20 & 17.59 & 23.37 \\
Q-VDiT+SVG & 6/6 & 25\% & 60.92 & 57.53 & 96.44 & 82.50 & 95.48 & 89.34 & 27.76 & 25.46 \\
QuaRot+SVG & 6/6 & 15\% & 61.42 & 54.09 & 95.78 & 45.83 & 94.70 & 89.95 & 16.50 & 23.08 \\
Q-VDiT+SVG & 6/6 & 15\% & 59.77 & 56.56 & 96.31 & 82.50 & 95.68 & 89.25 & 27.05 & 25.28 \\
\rowcolor{mycolor!30}\textbf{QuantSparse} & 6/6 & 25\% & \textbf{63.89} & \textbf{58.77} & \textbf{96.77} & \underline{84.72} & \textbf{96.48} & \textbf{90.91} & \textbf{29.80} & \underline{25.80} \\
\rowcolor{mycolor!30}\textbf{QuantSparse} & 6/6 & 15\% & \underline{63.87} & \underline{58.32} & \underline{96.69} & \textbf{90.28} & \underline{96.29} & \underline{90.85} & \underline{29.14} & \textbf{26.07} \\
\midrule

QuaRot+SVG & 4/8 & 25\% & 62.53 & 57.24 & 96.52 & 87.50 & 95.40 & 89.77 & 22.69 & 25.11 \\
Q-VDiT+SVG & 4/8 & 25\% & 60.92 & 58.53 & 96.44 & 87.50 & 95.48 & 89.34 & 22.76 & 25.46 \\
QuaRot+SVG & 4/8 & 15\% & 61.16 & 56.71 & 96.32 & 87.50 & 95.33 & 89.67 & 22.14 & 25.05 \\
Q-VDiT+SVG & 4/8 & 15\% & 59.77 & 57.56 & 96.31 & 87.50 & 95.68 & 89.25 & 22.04 & 25.08 \\
\rowcolor{mycolor!30}\textbf{QuantSparse} & 4/8 & 25\% & \underline{63.55} & \textbf{59.59} & \textbf{96.82} & \textbf{87.50} & \textbf{96.69} & \textbf{90.69} & \textbf{27.76} & \underline{25.81} \\
\rowcolor{mycolor!30}\textbf{QuantSparse} & 4/8 & 15\% & \textbf{63.81} & \underline{58.86} & \underline{96.75} & \textbf{87.50} & \underline{96.56} & \underline{90.55} & \underline{26.09} & \textbf{25.93} \\

\bottomrule
\end{tabular}

%% file: tables/abla_local.tex
\begin{tabular}{l|llll}
\toprule
\textbf{Method} & VQA$_{\mathbf{\red{\uparrow}}}$ &  PSNR$_{\mathbf{\red{\uparrow}}}$ & SSIM$_{\mathbf{\red{\uparrow}}}$ & LPIPS$_{\mathbf{\red{\downarrow}}}$ \\
\midrule  

None & 81.92 & 14.35 & 0.486 & 0.425 \\
\textit{Random} & 83.17 & 15.49 & 0.523 & 0.372 \\
\rowcolor{mycolor!30} \textbf{Salient} & \textbf{86.95}$_{\mathbf{\red{+5.03}}}$ & \textbf{16.82}$_{\mathbf{\red{+2.47}}}$ & \textbf{0.561}$_{\mathbf{\red{+0.075}}}$ & \textbf{0.325}$_{\mathbf{\red{-0.100}}}$ \\
\bottomrule
\end{tabular}

%% file: tables/abla_distill.tex
\begin{tabular}{c|llll}
\toprule
\textbf{Value} & VQA$_{\mathbf{\red{\uparrow}}}$ &  PSNR$_{\mathbf{\red{\uparrow}}}$ & SSIM$_{\mathbf{\red{\uparrow}}}$ & LPIPS$_{\mathbf{\red{\downarrow}}}$ \\
\midrule  

\makecell{None} & 81.92 & 14.35 & 0.486 & 0.425 \\

\multicolumn{5}{c}{\cellcolor[gray]{0.92} pooling stride $s$} \\

\rowcolor{mycolor!30} s=64 & 85.19 & 16.05 & 0.543 & 0.348 \\
\rowcolor{mycolor!30} \textbf{s=128} & 85.26 & 16.01 & 0.547 & 0.349 \\
\rowcolor{mycolor!30} s=256 & 85.12 & 15.93 & 0.545 & 0.355 \\

\multicolumn{5}{c}{\cellcolor[gray]{0.92} salient token $k$} \\

\rowcolor{mycolor!30} k=128 & 86.21 & 16.72 & 0.551 & 0.349 \\
\rowcolor{mycolor!30} \textbf{k=256} & 86.95 & 16.82 & 0.561 & 0.325 \\
\rowcolor{mycolor!30} k=512 & 86.93 & 16.95 & 0.561 & 0.324 \\

\bottomrule
\end{tabular}

%% file: tables/abla_svd.tex
\begin{tabular}{l|llll}
\toprule
\textbf{Method} & VQA$_{\mathbf{\red{\uparrow}}}$ &  PSNR$_{\mathbf{\red{\uparrow}}}$ & SSIM$_{\mathbf{\red{\uparrow}}}$ & LPIPS$_{\mathbf{\red{\downarrow}}}$ \\
\midrule  

None & 68.00 & 14.16 & 0.470 & 0.445 \\
\textit{First} & 70.82 & 17.08 & 0.572 & 0.285 \\
\textit{Second} & 89.73 & 18.68 & 0.616 & 0.258 \\
\textit{Third} & 89.71 & 18.70 & 0.620 & 0.263 \\
\rowcolor{mycolor!30} \textit{top-8} & 91.12 & 18.69 & 0.621 & 0.253 \\
\rowcolor{mycolor!30} \textbf{\textit{top-16}} & \textbf{91.98}$_{\mathbf{\red{+23.98}}}$ & \textbf{18.72}$_{\mathbf{\red{+4.56}}}$ & \textbf{0.630}$_{\mathbf{\red{+0.160}}}$ & \textbf{0.240}$_{\mathbf{\red{-0.205}}}$ \\
\rowcolor{mycolor!30} \textit{top-32} & 91.75 & 18.72 & 0.628 & 0.242 \\
\bottomrule
\end{tabular}

%% file: tables/distill_hyper.tex
\begin{tabular}{c|llll}
\toprule
\textbf{$\lambda_{*}$} & VQA$_{\mathbf{\red{\uparrow}}}$ &  PSNR$_{\mathbf{\red{\uparrow}}}$ & SSIM$_{\mathbf{\red{\uparrow}}}$ & LPIPS$_{\mathbf{\red{\downarrow}}}$ \\
\midrule  

None & 81.92 & 14.35 & 0.486 & 0.425 \\

\multicolumn{5}{c}{\cellcolor[gray]{0.92} * = global} \\

\rowcolor{mycolor!30} 5e-3 & 84.76 & 15.79 & 0.518 & 0.362 \\
\rowcolor{mycolor!30} \textbf{1e-4} & 85.26 & \textbf{16.01} & \textbf{0.547} & \textbf{0.349} \\
\rowcolor{mycolor!30} 5e-4 & \textbf{85.33} & 15.72 & 0.540 & 0.351 \\

\multicolumn{5}{c}{\cellcolor[gray]{0.92} * = local} \\

\rowcolor{mycolor!30} 5e-3 & 86.73 & \textbf{16.86} & 0.547 & 0.336 \\
\rowcolor{mycolor!30} \textbf{1e-4} & \textbf{86.95} & 16.82 & 0.561 & \textbf{0.325} \\
\rowcolor{mycolor!30} 5e-4 & 86.54 & 16.72 & \textbf{0.562} & 0.328 \\

\bottomrule
\end{tabular}

%% file: tables/calib.tex
\begin{tabular}{l|llll}
\toprule
\multirow{2}{*}{\textbf{Method}} & \multicolumn{2}{c}{\textbf{Calibration Overload}} & \multicolumn{2}{c}{\textbf{Performance}} \\
\cmidrule(lr){2-3}
\cmidrule(lr){4-5}
& GPU Memory (GB)$_{\mathbf{\red{\downarrow}}}$ & GPU Time (Hours)$_{\mathbf{\red{\downarrow}}}$ & PSNR$_{\mathbf{\red{\uparrow}}}$ & LPIPS$_{\mathbf{\red{\downarrow}}}$ \\
\midrule  

\multicolumn{5}{c}{\cellcolor[gray]{0.92}\texttt{Wan}2.1 1.3B} \\
\midrule

PTQ & 16.21 & 0.62 & 10.57 & 0.587 \\
+\textit{Global} & 16.27$_{\mathbf{\red{+0.4\%}}}$ & 0.63$_{\mathbf{\red{+0.2\%}}}$ & 13.27 & 0.452 \\
+\textit{Local} & 16.28$_{\mathbf{\red{+0.4\%}}}$ & 0.63$_{\mathbf{\red{+0.2\%}}}$ & 13.85 & 0.421 \\
\rowcolor{mycolor!30}\textbf{QuantSparse} & 16.34$_{\mathbf{\red{+0.8\%}}}$ & 0.64$_{\mathbf{\red{+1.6\%}}}$ & \textbf{15.22}$_{\mathbf{\red{+4.65}}}$ & \textbf{0.338}$_{\mathbf{\red{-0.249}}}$ \\
\midrule

\multicolumn{5}{c}{\cellcolor[gray]{0.92}HunyuanVideo 13B} \\
\midrule

PTQ & 39.22 & 5.08 & 16.27 & 0.472 \\
+\textit{Global} & 39.33$_{\mathbf{\red{+0.3\%}}}$ & 5.10$_{\mathbf{\red{+0.4\%}}}$ & 18.42 & 0.357 \\
+\textit{Local} & 39.32$_{\mathbf{\red{+0.3\%}}}$ & 5.11$_{\mathbf{\red{+0.6\%}}}$ & 18.96 & 0.342 \\
\rowcolor{mycolor!30}\textbf{QuantSparse} & 39.41$_{\mathbf{\red{+0.5\%}}}$ & 5.13$_{\mathbf{\red{+1.0\%}}}$ & \textbf{20.86}$_{\mathbf{\red{+4.59}}}$ & \textbf{0.272}$_{\mathbf{\red{-0.200}}}$ \\
\midrule

\multicolumn{5}{c}{\cellcolor[gray]{0.92}\texttt{Wan}2.1 14B} \\
\midrule

PTQ & 47.39 & 2.57 & 14.35 & 0.425 \\
+\textit{Global} & 47.54$_{\mathbf{\red{+0.3\%}}}$ & 2.59$_{\mathbf{\red{+0.8\%}}}$ & 16.01 & 0.349 \\
+\textit{Local} & 47.50$_{\mathbf{\red{+0.2\%}}}$ & 2.58$_{\mathbf{\red{+0.4\%}}}$ & 16.82 & 0.325 \\
\rowcolor{mycolor!30}\textbf{QuantSparse} & 47.65$_{\mathbf{\red{+0.5\%}}}$ & 2.60$_{\mathbf{\red{+1.1\%}}}$ & \textbf{18.72}$_{\mathbf{\red{+4.37}}}$ & \textbf{0.240}$_{\mathbf{\red{-0.185}}}$ \\

\bottomrule
\end{tabular}

%% file: tables/sparse_attn.tex
\begin{tabular}{l|llll}
\toprule
\multirow{2}{*}{\textbf{Method}} & \multicolumn{2}{c}{\textbf{Inference Overload}} & \multicolumn{2}{c}{\textbf{Performance}} \\
\cmidrule(lr){2-3}
\cmidrule(lr){4-5}
& GPU Memory (GB)$_{\mathbf{\red{\downarrow}}}$ & DiT Time (s)$_{\mathbf{\red{\downarrow}}}$ & PSNR$_{\mathbf{\red{\uparrow}}}$ & LPIPS$_{\mathbf{\red{\downarrow}}}$ \\
\midrule  

\multicolumn{5}{c}{\cellcolor[gray]{0.92}\texttt{Wan}2.1 1.3B} \\
\midrule

None & 5.44 & 312 & 10.57 & 0.587 \\
+\textit{First} & 5.84$_{\mathbf{\red{+7\%}}}$ & 313$_{\mathbf{\red{+0.3\%}}}$ & 12.76 & 0.493 \\
+\textit{Second} & 5.93$_{\mathbf{\red{+9\%}}}$ & 313$_{\mathbf{\red{+0.3\%}}}$ & 13.55 & 0.427 \\
\rowcolor{mycolor!30}\textbf{QuantSparse} & 5.93$_{\mathbf{\red{+9\%}}}$ & 313$_{\mathbf{\red{+0.3\%}}}$ & \textbf{15.22}$_{\mathbf{\red{+4.65}}}$ & \textbf{0.338}$_{\mathbf{\red{-0.249}}}$ \\
\midrule

\multicolumn{5}{c}{\cellcolor[gray]{0.92}HunyuanVideo 13B} \\
\midrule

None & 24.34 & 725 & 16.27 & 0.472 \\
+\textit{First} & 26.51$_{\mathbf{\red{+9\%}}}$ & 729$_{\mathbf{\red{+0.6\%}}}$ & 18.25 & 0.381 \\
+\textit{Second} & 27.02$_{\mathbf{\red{+11\%}}}$ & 730$_{\mathbf{\red{+0.7\%}}}$ & 19.03 & 0.317 \\
\rowcolor{mycolor!30}\textbf{QuantSparse} & 27.02$_{\mathbf{\red{+11\%}}}$ & 731$_{\mathbf{\red{+0.8\%}}}$ & \textbf{20.86}$_{\mathbf{\red{+4.59}}}$ & \textbf{0.272}$_{\mathbf{\red{-0.200}}}$ \\
\midrule

\multicolumn{5}{c}{\cellcolor[gray]{0.92}\texttt{Wan}2.1 14B} \\
\midrule

None & 26.04 & 2589 & 14.16 & 0.445 \\
+\textit{First} & 27.86$_{\mathbf{\red{+7\%}}}$ & 2593$_{\mathbf{\red{+0.2\%}}}$ & 17.08 & 0.285 \\
+\textit{Second} & 28.14$_{\mathbf{\red{+8\%}}}$ & 2594$_{\mathbf{\red{+0.2\%}}}$ & 18.68 & 0.258 \\
\rowcolor{mycolor!30}\textbf{QuantSparse} & 28.14$_{\mathbf{\red{+8\%}}}$ & 2594$_{\mathbf{\red{+0.2\%}}}$ & \textbf{18.72}$_{\mathbf{\red{+4.56}}}$ & \textbf{0.240}$_{\mathbf{\red{-0.205}}}$ \\

\bottomrule
\end{tabular}

%% file: tables/efficiency_more.tex
\begin{tabular}{ccc|c|ccccc}
\toprule
\multicolumn{3}{c}{\textbf{Method}} & \multirow{2}{*}{\textbf{Density}$_{\mathbf{\red{\downarrow}}}$} & \multicolumn{3}{c}{\textbf{Quality}} & \multicolumn{2}{c}{\textbf{Latency \& Speed}} \\
\cmidrule(lr){1-3}
\cmidrule(lr){5-7}  
\cmidrule(lr){8-9}
QuantSparse & SageAttention & TeaCache & & CLIPSIM$_{\mathbf{\red{\uparrow}}}$ & VQA$_{\mathbf{\red{\uparrow}}}$ & $\Delta$FSCore$_{\mathbf{\red{\downarrow}}}$ & DiT Time$_{\mathbf{\red{\downarrow}}}$ & Speedup$_{\mathbf{\red{\uparrow}}}$ \\ 
\midrule  

\multicolumn{9}{c}{\cellcolor[gray]{0.92}\texttt{Wan}2.1 14B ($\texttt{CFG}=5.0, 720\times1280p, \texttt{frames}=80$)} \\
\midrule

\multicolumn{3}{c}{Full Prec.} & 100\% & 0.182 & 90.79 & 0.000 & 4031s & 1.00$\times$  \\
\midrule

\cmark &  &   & \multirow{3}{*}{25\%} & 0.183 & 91.98 & 0.056 & 2594s & 1.55$\times$ \\
\cmark  & \cmark &   & \multirow{3}{*}{} & 0.181 & 91.70 & 0.240 & 2480s & 1.63$\times$ \\
\cmark & \cmark & \cmark & \multirow{3}{*}{} & 0.180 & 84.01 & 0.211 & 1802s & 2.24$\times$ \\

\midrule

\cmark &  &   & \multirow{3}{*}{15\%} & 0.182 & 90.73 & 0.042 & 2315s & 1.74$\times$ \\
\cmark & \cmark &   & \multirow{3}{*}{} & 0.180 & 90.58 & 0.046 & 2201s & 1.83$\times$ \\
\cmark & \cmark & \cmark & \multirow{3}{*}{} & 0.179 & 86.24 & 0.249 & 1629s & 2.47$\times$ \\

\bottomrule
\end{tabular}

%% file: main.bbl
\begin{thebibliography}{65}
\providecommand{\natexlab}[1]{#1}
\providecommand{\url}[1]{\texttt{#1}}
\expandafter\ifx\csname urlstyle\endcsname\relax
  \providecommand{\doi}[1]{doi: #1}\else
  \providecommand{\doi}{doi: \begingroup \urlstyle{rm}\Url}\fi

\bibitem[An et~al.(2025)An, Yu, Wang, Zhang, and Song]{an2025TheInnovationInformatics}
Zhulin An, Xinqiang Yu, Chu Wang, Yinlong Zhang, and Chunhe Song.
\newblock Embodied intelligence: Recent advances and future perspectives, 2025.
\newblock ISSN 3105-8515.
\newblock URL \url{https://www.the-innovation.org/informatics/article/id/68da75b1eaedd90f412200cb}.

\bibitem[Ashkboos et~al.(2024)Ashkboos, Mohtashami, Croci, Li, Cameron, Jaggi, Alistarh, Hoefler, and Hensman]{ashkboos2024quarot}
Saleh Ashkboos, Amirkeivan Mohtashami, Maximilian~L Croci, Bo~Li, Pashmina Cameron, Martin Jaggi, Dan Alistarh, Torsten Hoefler, and James Hensman.
\newblock Quarot: Outlier-free 4-bit inference in rotated llms.
\newblock \emph{arXiv preprint arXiv:2404.00456}, 2024.

\bibitem[Chen et~al.(2024)Chen, Meng, Tang, Ma, Jiang, Wang, Wang, and Zhu]{chen2024qdit}
Lei Chen, Yuan Meng, Chen Tang, Xinzhu Ma, Jingyan Jiang, Xin Wang, Zhi Wang, and Wenwu Zhu.
\newblock Q-dit: Accurate post-training quantization for diffusion transformers.
\newblock \emph{arXiv preprint arXiv:2406.17343}, 2024.

\bibitem[Chitty-Venkata et~al.(2023)Chitty-Venkata, Mittal, Emani, Vishwanath, and Somani]{chitty2023transformerquantsurvey}
Krishna~Teja Chitty-Venkata, Sparsh Mittal, Murali Emani, Venkatram Vishwanath, and Arun~K Somani.
\newblock A survey of techniques for optimizing transformer inference.
\newblock \emph{Journal of Systems Architecture}, pp.\  102990, 2023.

\bibitem[Ding et~al.(2024)Ding, Feng, Chen, Guo, and Liu]{ding2024reg}
Yifu Ding, Weilun Feng, Chuyan Chen, Jinyang Guo, and Xianglong Liu.
\newblock Reg-ptq: Regression-specialized post-training quantization for fully quantized object detector.
\newblock In \emph{Proceedings of the IEEE/CVF Conference on Computer Vision and Pattern Recognition}, pp.\  16174--16184, 2024.

\bibitem[Feng et~al.(2024)Feng, Yang, An, Huang, Diao, Wang, and Xu]{feng2024rdd}
Weilun Feng, Chuanguang Yang, Zhulin An, Libo Huang, Boyu Diao, Fei Wang, and Yongjun Xu.
\newblock Relational diffusion distillation for efficient image generation.
\newblock In \emph{Proceedings of the 32nd ACM International Conference on Multimedia}, pp.\  205--213, 2024.

\bibitem[Feng et~al.(2025{\natexlab{a}})Feng, Qin, Yang, An, Huang, Diao, Wang, Tao, Xu, and Magno]{feng2025mpqdm}
Weilun Feng, Haotong Qin, Chuanguang Yang, Zhulin An, Libo Huang, Boyu Diao, Fei Wang, Renshuai Tao, Yongjun Xu, and Michele Magno.
\newblock Mpq-dm: Mixed precision quantization for extremely low bit diffusion models.
\newblock In \emph{Proceedings of the AAAI Conference on Artificial Intelligence}, volume~39, pp.\  16595--16603, 2025{\natexlab{a}}.

\bibitem[Feng et~al.(2025{\natexlab{b}})Feng, Qin, Yang, Li, Yang, Li, An, Huang, Magno, and Xu]{feng2025s2qvdit}
Weilun Feng, Haotong Qin, Chuanguang Yang, Xiangqi Li, Han Yang, Yuqi Li, Zhulin An, Libo Huang, Michele Magno, and Yongjun Xu.
\newblock S$^2$q-vdit: Accurate quantized video diffusion transformer with salient data and sparse token distillation.
\newblock \emph{arXiv preprint arXiv:2508.04016}, 2025{\natexlab{b}}.

\bibitem[Feng et~al.(2025{\natexlab{c}})Feng, Yang, Qin, Li, Wang, An, Huang, Diao, Zhao, Xu, et~al.]{feng2025qvdit}
Weilun Feng, Chuanguang Yang, Haotong Qin, Xiangqi Li, Yu~Wang, Zhulin An, Libo Huang, Boyu Diao, Zixiang Zhao, Yongjun Xu, et~al.
\newblock Q-vdit: Towards accurate quantization and distillation of video-generation diffusion transformers.
\newblock In \emph{International Conference on Machine Learning}, pp.\  16956--16976. PMLR, 2025{\natexlab{c}}.

\bibitem[Feng et~al.(2025{\natexlab{d}})Feng, Yang, Qin, Li, Li, An, Huang, Diao, Zhuang, Magno, et~al.]{feng2025mpqdmv2}
Weilun Feng, Chuanguang Yang, Haotong Qin, Yuqi Li, Xiangqi Li, Zhulin An, Libo Huang, Boyu Diao, Fuzhen Zhuang, Michele Magno, et~al.
\newblock Mpq-dmv2: Flexible residual mixed precision quantization for low-bit diffusion models with temporal distillation.
\newblock \emph{arXiv preprint arXiv:2507.04290}, 2025{\natexlab{d}}.

\bibitem[Fu et~al.()Fu, Huang, Ning, Zhang, Chen, Wu, Wang, Huang, Li, Yan, et~al.]{fu2406moa}
Tianyu Fu, Haofeng Huang, Xuefei Ning, Genghan Zhang, Boju Chen, Tianqi Wu, Hongyi Wang, Zixiao Huang, Shiyao Li, Shengen Yan, et~al.
\newblock Moa: Mixture of sparse attention for automatic large language model compression, 2024.
\newblock \emph{URL https://arxiv. org/abs/2406.14909}.

\bibitem[Gao et~al.(2024)Gao, Zeng, Du, Cao, Zhou, Qi, Lai, So, Cao, Yang, et~al.]{gao2024seerattention}
Yizhao Gao, Zhichen Zeng, Dayou Du, Shijie Cao, Peiyuan Zhou, Jiaxing Qi, Junjie Lai, Hayden Kwok-Hay So, Ting Cao, Fan Yang, et~al.
\newblock Seerattention: Learning intrinsic sparse attention in your llms.
\newblock \emph{arXiv preprint arXiv:2410.13276}, 2024.

\bibitem[Gholami et~al.(2022)Gholami, Kim, Dong, Yao, Mahoney, and Keutzer]{gholami2022quantizationsurvey}
Amir Gholami, Sehoon Kim, Zhen Dong, Zhewei Yao, Michael~W Mahoney, and Kurt Keutzer.
\newblock A survey of quantization methods for efficient neural network inference.
\newblock In \emph{Low-Power Computer Vision}, pp.\  291--326. Chapman and Hall/CRC, 2022.

\bibitem[He et~al.(2023)He, Liu, Wu, Zhou, and Zhuang]{he2023efficientdm}
Yefei He, Jing Liu, Weijia Wu, Hong Zhou, and Bohan Zhuang.
\newblock Efficientdm: Efficient quantization-aware fine-tuning of low-bit diffusion models.
\newblock \emph{arXiv preprint arXiv:2310.03270}, 2023.

\bibitem[He et~al.(2024)He, Liu, Liu, Wu, Zhou, and Zhuang]{he2024ptqd}
Yefei He, Luping Liu, Jing Liu, Weijia Wu, Hong Zhou, and Bohan Zhuang.
\newblock Ptqd: Accurate post-training quantization for diffusion models.
\newblock \emph{Advances in Neural Information Processing Systems}, 36, 2024.

\bibitem[{HPC-AI}(2024)]{open-sora}
{HPC-AI}.
\newblock Open-sora, 2024.
\newblock URL \url{https://github.com/hpcaiitech/Open-Sora}.

\bibitem[Huang et~al.(2024{\natexlab{a}})Huang, Gong, Liu, Chen, and Liu]{huang2024tfmq}
Yushi Huang, Ruihao Gong, Jing Liu, Tianlong Chen, and Xianglong Liu.
\newblock Tfmq-dm: Temporal feature maintenance quantization for diffusion models.
\newblock In \emph{Proceedings of the IEEE/CVF Conference on Computer Vision and Pattern Recognition}, pp.\  7362--7371, 2024{\natexlab{a}}.

\bibitem[Huang et~al.(2024{\natexlab{b}})Huang, He, Yu, Zhang, Si, Jiang, Zhang, Wu, Jin, Chanpaisit, et~al.]{huang2024vbench}
Ziqi Huang, Yinan He, Jiashuo Yu, Fan Zhang, Chenyang Si, Yuming Jiang, Yuanhan Zhang, Tianxing Wu, Qingyang Jin, Nattapol Chanpaisit, et~al.
\newblock Vbench: Comprehensive benchmark suite for video generative models.
\newblock In \emph{Proceedings of the IEEE/CVF Conference on Computer Vision and Pattern Recognition}, pp.\  21807--21818, 2024{\natexlab{b}}.

\bibitem[Jacob et~al.(2018)Jacob, Kligys, Chen, Zhu, Tang, Howard, Adam, and Kalenichenko]{jacob2018quantizationandtrain}
Benoit Jacob, Skirmantas Kligys, Bo~Chen, Menglong Zhu, Matthew Tang, Andrew Howard, Hartwig Adam, and Dmitry Kalenichenko.
\newblock Quantization and training of neural networks for efficient integer-arithmetic-only inference.
\newblock In \emph{Proceedings of the IEEE conference on computer vision and pattern recognition}, pp.\  2704--2713, 2018.

\bibitem[Kong et~al.(2024)Kong, Tian, Zhang, Min, Dai, Zhou, Xiong, Li, Wu, Zhang, et~al.]{kong2024hunyuanvideo}
Weijie Kong, Qi~Tian, Zijian Zhang, Rox Min, Zuozhuo Dai, Jin Zhou, Jiangfeng Xiong, Xin Li, Bo~Wu, Jianwei Zhang, et~al.
\newblock Hunyuanvideo: A systematic framework for large video generative models.
\newblock \emph{arXiv preprint arXiv:2412.03603}, 2024.

\bibitem[Li et~al.(2024{\natexlab{a}})Li, Lin, Zhang, Cai, Li, Guo, Xie, Meng, Zhu, and Han]{li2024svdqunat}
Muyang Li, Yujun Lin, Zhekai Zhang, Tianle Cai, Xiuyu Li, Junxian Guo, Enze Xie, Chenlin Meng, Jun-Yan Zhu, and Song Han.
\newblock Svdqunat: Absorbing outliers by low-rank components for 4-bit diffusion models.
\newblock \emph{arXiv preprint arXiv:2411.05007}, 2024{\natexlab{a}}.

\bibitem[Li et~al.(2024{\natexlab{b}})Li, Xu, Cao, Sun, and Zhang]{li2024qdm}
Yanjing Li, Sheng Xu, Xianbin Cao, Xiao Sun, and Baochang Zhang.
\newblock Q-dm: An efficient low-bit quantized diffusion model.
\newblock \emph{Advances in Neural Information Processing Systems}, 36, 2024{\natexlab{b}}.

\bibitem[Li et~al.(2024{\natexlab{c}})Li, Zhang, Lin, Xiong, Long, Deng, Zhang, Liu, Huang, Xiao, et~al.]{li2024hunyuandit}
Zhimin Li, Jianwei Zhang, Qin Lin, Jiangfeng Xiong, Yanxin Long, Xinchi Deng, Yingfang Zhang, Xingchao Liu, Minbin Huang, Zedong Xiao, et~al.
\newblock Hunyuan-dit: A powerful multi-resolution diffusion transformer with fine-grained chinese understanding.
\newblock \emph{arXiv preprint arXiv:2405.08748}, 2024{\natexlab{c}}.

\bibitem[Liu et~al.(2024{\natexlab{a}})Liu, Zhang, Wang, Wei, Qiu, Zhao, Zhang, Ye, and Wan]{liu2024teacache}
Feng Liu, Shiwei Zhang, Xiaofeng Wang, Yujie Wei, Haonan Qiu, Yuzhong Zhao, Yingya Zhang, Qixiang Ye, and Fang Wan.
\newblock Timestep embedding tells: It's time to cache for video diffusion model.
\newblock \emph{arXiv preprint arXiv:2411.19108}, 2024{\natexlab{a}}.

\bibitem[Liu et~al.(2024{\natexlab{b}})Liu, Cun, Liu, Wang, Zhang, Chen, Liu, Zeng, Chan, and Shan]{liu2024evalcrafter}
Yaofang Liu, Xiaodong Cun, Xuebo Liu, Xintao Wang, Yong Zhang, Haoxin Chen, Yang Liu, Tieyong Zeng, Raymond Chan, and Ying Shan.
\newblock Evalcrafter: Benchmarking and evaluating large video generation models.
\newblock In \emph{Proceedings of the IEEE/CVF Conference on Computer Vision and Pattern Recognition}, pp.\  22139--22149, 2024{\natexlab{b}}.

\bibitem[Liu et~al.(2024{\natexlab{c}})Liu, Zhang, Li, Yan, Gao, Chen, Yuan, Huang, Sun, Gao, et~al.]{liu2024sora}
Yixin Liu, Kai Zhang, Yuan Li, Zhiling Yan, Chujie Gao, Ruoxi Chen, Zhengqing Yuan, Yue Huang, Hanchi Sun, Jianfeng Gao, et~al.
\newblock Sora: A review on background, technology, limitations, and opportunities of large vision models.
\newblock \emph{arXiv preprint arXiv:2402.17177}, 2024{\natexlab{c}}.

\bibitem[Loshchilov \& Hutter(2017)Loshchilov and Hutter]{loshchilov2017adamw}
Ilya Loshchilov and Frank Hutter.
\newblock Decoupled weight decay regularization.
\newblock \emph{arXiv preprint arXiv:1711.05101}, 2017.

\bibitem[Lou et~al.(2024)Lou, Jia, Zheng, and Tu]{lou2024sparser}
Chao Lou, Zixia Jia, Zilong Zheng, and Kewei Tu.
\newblock Sparser is faster and less is more: Efficient sparse attention for long-range transformers.
\newblock \emph{arXiv preprint arXiv:2406.16747}, 2024.

\bibitem[Lu et~al.(2025)Lu, Jiang, Liu, Du, Jiang, Hong, Liu, He, Yuan, Wang, et~al.]{lu2025moba}
Enzhe Lu, Zhejun Jiang, Jingyuan Liu, Yulun Du, Tao Jiang, Chao Hong, Shaowei Liu, Weiran He, Enming Yuan, Yuzhi Wang, et~al.
\newblock Moba: Mixture of block attention for long-context llms.
\newblock \emph{arXiv preprint arXiv:2502.13189}, 2025.

\bibitem[Ma et~al.(2024{\natexlab{a}})Ma, Fang, and Wang]{ma2024deepcache}
Xinyin Ma, Gongfan Fang, and Xinchao Wang.
\newblock Deepcache: Accelerating diffusion models for free.
\newblock In \emph{Proceedings of the IEEE/CVF conference on computer vision and pattern recognition}, pp.\  15762--15772, 2024{\natexlab{a}}.

\bibitem[Ma et~al.(2023{\natexlab{a}})Ma, Jin, Zheng, Wang, Li, Wu, Jiang, Zhang, and Ji]{ma2023ompq}
Yuexiao Ma, Taisong Jin, Xiawu Zheng, Yan Wang, Huixia Li, Yongjian Wu, Guannan Jiang, Wei Zhang, and Rongrong Ji.
\newblock Ompq: Orthogonal mixed precision quantization.
\newblock In \emph{Proceedings of the AAAI conference on artificial intelligence}, volume~37, pp.\  9029--9037, 2023{\natexlab{a}}.

\bibitem[Ma et~al.(2023{\natexlab{b}})Ma, Li, Zheng, Xiao, Wang, Wen, Pan, Chao, and Ji]{ma2023solving}
Yuexiao Ma, Huixia Li, Xiawu Zheng, Xuefeng Xiao, Rui Wang, Shilei Wen, Xin Pan, Fei Chao, and Rongrong Ji.
\newblock Solving oscillation problem in post-training quantization through a theoretical perspective.
\newblock In \emph{Proceedings of the IEEE/CVF Conference on Computer Vision and Pattern Recognition}, pp.\  7950--7959, 2023{\natexlab{b}}.

\bibitem[Ma et~al.(2024{\natexlab{b}})Ma, Li, Zheng, Ling, Xiao, Wang, Wen, Chao, and Ji]{ma2024affinequant}
Yuexiao Ma, Huixia Li, Xiawu Zheng, Feng Ling, Xuefeng Xiao, Rui Wang, Shilei Wen, Fei Chao, and Rongrong Ji.
\newblock Affinequant: Affine transformation quantization for large language models.
\newblock \emph{arXiv preprint arXiv:2403.12544}, 2024{\natexlab{b}}.

\bibitem[Ma et~al.(2024{\natexlab{c}})Ma, Li, Zheng, Ling, Xiao, Wang, Wen, Chao, and Ji]{ma2024outlier}
Yuexiao Ma, Huixia Li, Xiawu Zheng, Feng Ling, Xuefeng Xiao, Rui Wang, Shilei Wen, Fei Chao, and Rongrong Ji.
\newblock Outlier-aware slicing for post-training quantization in vision transformer.
\newblock In \emph{Forty-first International Conference on Machine Learning}, 2024{\natexlab{c}}.

\bibitem[Peebles \& Xie(2023)Peebles and Xie]{peebles2023dit}
William Peebles and Saining Xie.
\newblock Scalable diffusion models with transformers.
\newblock In \emph{Proceedings of the IEEE/CVF International Conference on Computer Vision}, pp.\  4195--4205, 2023.

\bibitem[Pilipovi{\'c} et~al.(2018)Pilipovi{\'c}, Buli{\'c}, and Risojevi{\'c}]{pilipovic2018cnnquantsurvey}
Ratko Pilipovi{\'c}, Patricio Buli{\'c}, and Vladimir Risojevi{\'c}.
\newblock Compression of convolutional neural networks: A short survey.
\newblock In \emph{2018 17th International Symposium INFOTEH-JAHORINA (INFOTEH)}, pp.\  1--6. IEEE, 2018.

\bibitem[Ren et~al.(2025)Ren, Yu, He, Yuille, and Chen]{ren2025grouping}
Sucheng Ren, Qihang Yu, Ju~He, Alan Yuille, and Liang-Chieh Chen.
\newblock Grouping first, attending smartly: Training-free acceleration for diffusion transformers.
\newblock \emph{arXiv preprint arXiv:2505.14687}, 2025.

\bibitem[Saharia et~al.(2022)Saharia, Chan, Saxena, Li, Whang, Denton, Ghasemipour, Gontijo~Lopes, Karagol~Ayan, Salimans, et~al.]{saharia2022drawbench}
Chitwan Saharia, William Chan, Saurabh Saxena, Lala Li, Jay Whang, Emily~L Denton, Kamyar Ghasemipour, Raphael Gontijo~Lopes, Burcu Karagol~Ayan, Tim Salimans, et~al.
\newblock Photorealistic text-to-image diffusion models with deep language understanding.
\newblock \emph{Advances in neural information processing systems}, 35:\penalty0 36479--36494, 2022.

\bibitem[Shang et~al.(2023)Shang, Yuan, Xie, Wu, and Yan]{shang2023ptq4dm}
Yuzhang Shang, Zhihang Yuan, Bin Xie, Bingzhe Wu, and Yan Yan.
\newblock Post-training quantization on diffusion models.
\newblock In \emph{Proceedings of the IEEE/CVF conference on computer vision and pattern recognition}, pp.\  1972--1981, 2023.

\bibitem[Sun et~al.(2024{\natexlab{a}})Sun, Zhang, Shah, Sun, Zhang, Li, Duan, Wei, and Ranjan]{sun2024sorasurvey}
Rui Sun, Yumin Zhang, Tejal Shah, Jiahao Sun, Shuoying Zhang, Wenqi Li, Haoran Duan, Bo~Wei, and Rajiv Ranjan.
\newblock From sora what we can see: A survey of text-to-video generation.
\newblock \emph{arXiv preprint arXiv:2405.10674}, 2024{\natexlab{a}}.

\bibitem[Sun et~al.(2024{\natexlab{b}})Sun, Liu, Bai, Bao, Zhao, Li, Hu, Yu, Hou, Yuan, et~al.]{sun2024flatquant}
Yuxuan Sun, Ruikang Liu, Haoli Bai, Han Bao, Kang Zhao, Yuening Li, Jiaxin Hu, Xianzhi Yu, Lu~Hou, Chun Yuan, et~al.
\newblock Flatquant: Flatness matters for llm quantization.
\newblock \emph{arXiv preprint arXiv:2410.09426}, 2024{\natexlab{b}}.

\bibitem[Thakkar et~al.(2023)]{cutlass}
V.~Thakkar et~al.
\newblock {CUTLASS}, 2023.
\newblock URL \url{https://github.com/NVIDIA/cutlass}.

\bibitem[Wan et~al.(2025)Wan, Wang, Ai, Wen, Mao, Xie, Chen, Yu, Zhao, Yang, et~al.]{wan2025wan}
Team Wan, Ang Wang, Baole Ai, Bin Wen, Chaojie Mao, Chen-Wei Xie, Di~Chen, Feiwu Yu, Haiming Zhao, Jianxiao Yang, et~al.
\newblock Wan: Open and advanced large-scale video generative models.
\newblock \emph{arXiv preprint arXiv:2503.20314}, 2025.

\bibitem[Wei et~al.(2024)Wei, Ma, Yang, and Yao]{wei202ptqsurvey}
Lu~Wei, Zhong Ma, Chaojie Yang, and Qin Yao.
\newblock Advances in the neural network quantization: A comprehensive review.
\newblock \emph{Applied Sciences}, 14\penalty0 (17):\penalty0 7445, 2024.

\bibitem[Wu et~al.(2021)Wu, Huang, Zhang, Li, Ji, Yang, Sapiro, and Duan]{wu2021godiva}
Chenfei Wu, Lun Huang, Qianxi Zhang, Binyang Li, Lei Ji, Fan Yang, Guillermo Sapiro, and Nan Duan.
\newblock Godiva: Generating open-domain videos from natural descriptions.
\newblock \emph{arXiv preprint arXiv:2104.14806}, 2021.

\bibitem[Wu et~al.(2023)Wu, Zhang, Liao, Chen, Hou, Wang, Sun, Yan, and Lin]{wu2023dover}
Haoning Wu, Erli Zhang, Liang Liao, Chaofeng Chen, Jingwen Hou, Annan Wang, Wenxiu Sun, Qiong Yan, and Weisi Lin.
\newblock Exploring video quality assessment on user generated contents from aesthetic and technical perspectives.
\newblock In \emph{Proceedings of the IEEE/CVF International Conference on Computer Vision}, pp.\  20144--20154, 2023.

\bibitem[Wu et~al.(2024)Wu, Wang, Shang, Shah, and Yan]{wu2024ptq4dit}
Junyi Wu, Haoxuan Wang, Yuzhang Shang, Mubarak Shah, and Yan Yan.
\newblock Ptq4dit: Post-training quantization for diffusion transformers.
\newblock \emph{arXiv preprint arXiv:2405.16005}, 2024.

\bibitem[Xi et~al.(2025)Xi, Yang, Zhao, Xu, Li, Li, Lin, Cai, Zhang, Li, et~al.]{xi2025svg}
Haocheng Xi, Shuo Yang, Yilong Zhao, Chenfeng Xu, Muyang Li, Xiuyu Li, Yujun Lin, Han Cai, Jintao Zhang, Dacheng Li, et~al.
\newblock Sparse videogen: Accelerating video diffusion transformers with spatial-temporal sparsity.
\newblock \emph{arXiv preprint arXiv:2502.01776}, 2025.

\bibitem[Xiao et~al.(2024{\natexlab{a}})Xiao, Zhang, Han, Xiao, Lin, Zhang, Liu, and Sun]{xiao2024infllm}
Chaojun Xiao, Pengle Zhang, Xu~Han, Guangxuan Xiao, Yankai Lin, Zhengyan Zhang, Zhiyuan Liu, and Maosong Sun.
\newblock Infllm: Training-free long-context extrapolation for llms with an efficient context memory.
\newblock \emph{Advances in Neural Information Processing Systems}, 37:\penalty0 119638--119661, 2024{\natexlab{a}}.

\bibitem[Xiao et~al.(2023{\natexlab{a}})Xiao, Lin, Seznec, Wu, Demouth, and Han]{xiao2023smoothquant}
Guangxuan Xiao, Ji~Lin, Mickael Seznec, Hao Wu, Julien Demouth, and Song Han.
\newblock Smoothquant: Accurate and efficient post-training quantization for large language models.
\newblock In \emph{International Conference on Machine Learning}, pp.\  38087--38099. PMLR, 2023{\natexlab{a}}.

\bibitem[Xiao et~al.(2023{\natexlab{b}})Xiao, Tian, Chen, Han, and Lewis]{xiao2023efficient}
Guangxuan Xiao, Yuandong Tian, Beidi Chen, Song Han, and Mike Lewis.
\newblock Efficient streaming language models with attention sinks.
\newblock \emph{arXiv preprint arXiv:2309.17453}, 2023{\natexlab{b}}.

\bibitem[Xiao et~al.(2024{\natexlab{b}})Xiao, Tang, Zuo, Guo, Yang, Tang, Fu, and Han]{xiao2024duoattention}
Guangxuan Xiao, Jiaming Tang, Jingwei Zuo, Junxian Guo, Shang Yang, Haotian Tang, Yao Fu, and Song Han.
\newblock Duoattention: Efficient long-context llm inference with retrieval and streaming heads.
\newblock \emph{arXiv preprint arXiv:2410.10819}, 2024{\natexlab{b}}.

\bibitem[Yuan et~al.(2025)Yuan, Gao, Dai, Luo, Zhao, Zhang, Xie, Wei, Wang, Xiao, et~al.]{yuan2025native}
Jingyang Yuan, Huazuo Gao, Damai Dai, Junyu Luo, Liang Zhao, Zhengyan Zhang, Zhenda Xie, YX~Wei, Lean Wang, Zhiping Xiao, et~al.
\newblock Native sparse attention: Hardware-aligned and natively trainable sparse attention.
\newblock \emph{arXiv preprint arXiv:2502.11089}, 2025.

\bibitem[Yuan et~al.(2024)Yuan, Zhang, Pu, Ning, Zhang, Zhao, Yan, Dai, and Wang]{yuan2024ditfastattn}
Zhihang Yuan, Hanling Zhang, Lu~Pu, Xuefei Ning, Linfeng Zhang, Tianchen Zhao, Shengen Yan, Guohao Dai, and Yu~Wang.
\newblock Ditfastattn: Attention compression for diffusion transformer models.
\newblock \emph{Advances in Neural Information Processing Systems}, 37:\penalty0 1196--1219, 2024.

\bibitem[Zhang et~al.(2024{\natexlab{a}})Zhang, Huang, Zhang, Wei, Zhu, and Chen]{zhang2024sageattention2}
Jintao Zhang, Haofeng Huang, Pengle Zhang, Jia Wei, Jun Zhu, and Jianfei Chen.
\newblock Sageattention2: Efficient attention with thorough outlier smoothing and per-thread int4 quantization.
\newblock \emph{arXiv preprint arXiv:2411.10958}, 2024{\natexlab{a}}.

\bibitem[Zhang et~al.(2024{\natexlab{b}})Zhang, Wei, Huang, Zhang, Zhu, and Chen]{zhang2024sageattention}
Jintao Zhang, Jia Wei, Haofeng Huang, Pengle Zhang, Jun Zhu, and Jianfei Chen.
\newblock Sageattention: Accurate 8-bit attention for plug-and-play inference acceleration.
\newblock \emph{arXiv preprint arXiv:2410.02367}, 2024{\natexlab{b}}.

\bibitem[Zhang et~al.(2025{\natexlab{a}})Zhang, Wei, Zhang, Xu, Huang, Wang, Jiang, Zhu, and Chen]{zhang2025sageattention3}
Jintao Zhang, Jia Wei, Pengle Zhang, Xiaoming Xu, Haofeng Huang, Haoxu Wang, Kai Jiang, Jun Zhu, and Jianfei Chen.
\newblock Sageattention3: Microscaling fp4 attention for inference and an exploration of 8-bit training.
\newblock \emph{arXiv preprint arXiv:2505.11594}, 2025{\natexlab{a}}.

\bibitem[Zhang et~al.(2025{\natexlab{b}})Zhang, Xiang, Huang, Wei, Xi, Zhu, and Chen]{zhang2025spargeattn}
Jintao Zhang, Chendong Xiang, Haofeng Huang, Jia Wei, Haocheng Xi, Jun Zhu, and Jianfei Chen.
\newblock Spargeattn: Accurate sparse attention accelerating any model inference.
\newblock \emph{arXiv preprint arXiv:2502.18137}, 2025{\natexlab{b}}.

\bibitem[Zhang et~al.(2025{\natexlab{c}})Zhang, Chen, Su, Ding, Stoica, Liu, and Zhang]{zhang2025fasttile}
Peiyuan Zhang, Yongqi Chen, Runlong Su, Hangliang Ding, Ion Stoica, Zhenghong Liu, and Hao Zhang.
\newblock Fast video generation with sliding tile attention.
\newblock \emph{arXiv preprint arXiv:2502.04507}, 2025{\natexlab{c}}.

\bibitem[Zhang et~al.(2018)Zhang, Isola, Efros, Shechtman, and Wang]{zhang2018lpips}
Richard Zhang, Phillip Isola, Alexei~A Efros, Eli Shechtman, and Oliver Wang.
\newblock The unreasonable effectiveness of deep features as a perceptual metric.
\newblock In \emph{Proceedings of the IEEE conference on computer vision and pattern recognition}, pp.\  586--595, 2018.

\bibitem[Zhang et~al.(2025{\natexlab{d}})Zhang, Xing, Xia, Liu, Peng, Tao, Wan, Lo, and Jia]{zhang2025jenga}
Yuechen Zhang, Jinbo Xing, Bin Xia, Shaoteng Liu, Bohao Peng, Xin Tao, Pengfei Wan, Eric Lo, and Jiaya Jia.
\newblock Training-free efficient video generation via dynamic token carving.
\newblock \emph{arXiv preprint arXiv:2505.16864}, 2025{\natexlab{d}}.

\bibitem[Zhang et~al.(2023)Zhang, Sheng, Zhou, Chen, Zheng, Cai, Song, Tian, R{\'e}, Barrett, et~al.]{zhang2023h2o}
Zhenyu Zhang, Ying Sheng, Tianyi Zhou, Tianlong Chen, Lianmin Zheng, Ruisi Cai, Zhao Song, Yuandong Tian, Christopher R{\'e}, Clark Barrett, et~al.
\newblock H2o: Heavy-hitter oracle for efficient generative inference of large language models.
\newblock \emph{Advances in Neural Information Processing Systems}, 36:\penalty0 34661--34710, 2023.

\bibitem[Zhao et~al.(2024)Zhao, Fang, Liu, Rui, Soedarmadji, Li, Lin, Dai, Yan, Yang, et~al.]{zhao2024vidit}
Tianchen Zhao, Tongcheng Fang, Enshu Liu, Wan Rui, Widyadewi Soedarmadji, Shiyao Li, Zinan Lin, Guohao Dai, Shengen Yan, Huazhong Yang, et~al.
\newblock Vidit-q: Efficient and accurate quantization of diffusion transformers for image and video generation.
\newblock \emph{arXiv preprint arXiv:2406.02540}, 2024.

\bibitem[Zheng et~al.(2024{\natexlab{a}})Zheng, Liu, Bian, Ma, Zhang, Wang, Guo, and Qin]{zheng2024bidm}
Xingyu Zheng, Xianglong Liu, Yichen Bian, Xudong Ma, Yulun Zhang, Jiakai Wang, Jinyang Guo, and Haotong Qin.
\newblock Bidm: Pushing the limit of quantization for diffusion models.
\newblock \emph{arXiv preprint arXiv:2412.05926}, 2024{\natexlab{a}}.

\bibitem[Zheng et~al.(2024{\natexlab{b}})Zheng, Qin, Ma, Zhang, Hao, Wang, Zhao, Guo, and Liu]{zheng2024binarydm}
Xingyu Zheng, Haotong Qin, Xudong Ma, Mingyuan Zhang, Haojie Hao, Jiakai Wang, Zixiang Zhao, Jinyang Guo, and Xianglong Liu.
\newblock Binarydm: Towards accurate binarization of diffusion model.
\newblock \emph{arXiv preprint arXiv:2404.05662}, 2024{\natexlab{b}}.

\end{thebibliography}
